\documentclass{article}
\usepackage{bbm}
\usepackage{dsfont} %
\usepackage{amsthm, url}

\usepackage[toc,page,header]{appendix}
\usepackage{minitoc}

\usepackage{tcolorbox}
\tcbuselibrary{listingsutf8}
\usepackage{amsmath}
\usepackage{amssymb}
\usepackage{graphicx}

\newcommand\blfootnote[1]{%
  \begin{NoHyper}%
  \renewcommand\thefootnote{}\footnote{#1}%
  \addtocounter{footnote}{-1}%
  \end{NoHyper}%
}

\newtheorem{theorem}{Theorem}

\doparttoc %
\faketableofcontents %

\newcommand{\antonios}[1]{\textcolor{blue}{[Antonios: #1]}}

\usepackage[table,xcdraw]{xcolor}
\usepackage{adjustbox}
\usepackage{tikz}
\usetikzlibrary{trees,positioning,shapes,shadows,arrows.meta}

\definecolor{light_red}{rgb}{1, 0.5, 0.5}
\definecolor{light_blue}{rgb}{0.5, 0.5, 1}
\definecolor{light_orange}{rgb}{1, 0.75, 0.5}
\definecolor{dark_green}{rgb}{0, 0.42, 0.3}

\usepackage{pdflscape}
\usepackage{multirow}
\usepackage{multicol}
\usepackage{makecell}
\usepackage{booktabs}
\usepackage{array}
\usepackage{rotating}
\usepackage{bbding}
\usepackage[edges]{forest}
\usepackage{makecell}

\definecolor{light_red}{rgb}{1, 0.5, 0.5}
\definecolor{light_blue}{rgb}{0.5, 0.5, 1}
\definecolor{light_orange}{rgb}{1, 0.75, 0.5}
\definecolor{dark_green}{rgb}{0, 0.42, 0.3}

\usepackage{multirow}

\usepackage{amsmath,amsfonts,stmaryrd,amssymb,mathtools} %
\usepackage{cancel}

\usepackage{booktabs}

\usepackage{enumerate} %

\usepackage[framemethod=tikz]{mdframed} %

\usepackage{hyperref}

\usepackage{algorithm}
\usepackage{algorithmicx}
\usepackage{algpseudocode}

\usepackage{pdfpages}
\usepackage{cleveref}

\usepackage{pifont}%

\usepackage{tikz}

\usepackage{rotating}
\usepackage{lscape}
\usepackage{wrapfig}

\usetikzlibrary{mindmap,shadows}

\usepackage[main, final]{neurips_2025}

\usepackage[utf8]{inputenc} %
\usepackage[T1]{fontenc}    %
\usepackage{hyperref}       %
\usepackage{url}            %
\usepackage{booktabs}       %
\usepackage{amsfonts}       %
\usepackage{nicefrac}       %
\usepackage{microtype}      %
\usepackage{enumitem}

\newcommand{\eqdef}{\,{\buildrel \text{def} \over =}\,}

\title{C3PO: Optimized Large Language Model Cascades with Probabilistic Cost Constraints for Reasoning}

\author{%
Antonios Valkanas$^{*123}$ \quad Soumyasundar Pal$^{4}$ \quad Pavel Rumiantsev$^{123}$\\ \quad \textbf{Yingxue Zhang}$^4$ \quad \textbf{Mark Coates}$^{123}$ \\
$^1$McGill University \quad $^2$ Mila - Quebec AI Institute \quad $^3$ Int. Lab. Learning Systems \quad $^4$ Huawei\\
Montr\'eal, Canada\\
\texttt{\{antonios.valkanas, pavel.rumiantsev\}@mail.mcgill.ca}\\
\texttt{\{soumyasundar.pal3, yingxue.zhang\}@huawei.com}\\
\texttt{mark.coates@mcgill.ca}
}

\begin{document}

\maketitle

\begin{abstract}
Large language models (LLMs) have achieved impressive results on complex reasoning tasks, but their high inference cost remains a major barrier to real-world deployment. A promising solution is to use cascaded inference, where small, cheap models handle easy queries, and only the hardest examples are escalated to more powerful models. However, existing cascade methods typically rely on supervised training with labeled data, offer no theoretical generalization guarantees, and provide limited control over test-time computational cost.
We introduce \textbf{C3PO} (\textit{Cost Controlled Cascaded Prediction Optimization}), a self-supervised framework for optimizing LLM cascades under probabilistic cost constraints. By focusing on minimizing regret with respect to the most powerful model (MPM), C3PO avoids the need for labeled data by constructing a cascade using only unlabeled model outputs. It leverages conformal prediction to bound the probability that inference cost exceeds a user-specified budget.
We provide theoretical guarantees on both cost control and generalization error, and show that our optimization procedure is effective even with small calibration sets. Empirically, C3PO achieves state-of-the-art performance across a diverse set of reasoning benchmarks including GSM8K, MATH-500, BigBench-Hard and AIME, outperforming strong LLM cascading baselines in both accuracy and cost-efficiency. Our results demonstrate that principled, label-free cascade optimization can enable scalable LLM deployment.
\end{abstract}

\section{Introduction}

\blfootnote{$^*$The  main contributor's contact address is: \texttt{antonios.valkanas@mail.mcgill.ca}.}Large language models (LLMs) have demonstrated remarkable capabilities in a wide range of reasoning tasks, from arithmetic problem solving to common sense and formal logical reasoning. Techniques such as chain‐of‐thought prompting have further unlocked multi‐step reasoning by encouraging LLMs to provide explicit intermediate rationale steps before arriving at an answer \citep{wei2022cot}. However, these gains come at a steep cost as each additional token or sample incurs inference time and monetary expenditure, imposing a barrier for real‐world deployment of LLMs.

A natural remedy is to employ a \emph{cascade} of models of increasing size and accuracy. This way, cheap and small models handle easy queries, deferring only the hard cases to a powerful LLM. From early works in classifier cascades~\citep{viola2001rapid} to more recent efforts in LLM cascading, it has been conclusively shown that cascades can yield dramatic cost savings with minimal accuracy loss. Yet existing LLM cascade approaches share two major limitations. First, they rely on large labeled datasets for meta‐model training under budget constraints, which is expensive to collect and slow to adapt to new domains~\citep{zhang2024treacle}. Second, they offer few theoretical guarantees on either generalization error or worst‐case cost overruns, leaving practitioners to tune budgets and thresholds by hand~\citep{yue2024large}. While model weight tuning is not a major hindrance in general cascade settings, the high training costs of LLMs make finetuning costly. Furthermore, the need for test time adaptation using few examples, motivates development of specialized LLM cascade methods that do not require ground truth labels.

In this work, we introduce \emph{Cost Controlled Cascaded Prediction Optimization} (\textbf{C3PO}), a self‐supervised cascaded inference framework that overcomes these challenges. Rather than depending on labels or costly meta‐training, C3PO learns to exit early by exploiting \emph{consistency signals} between a cheap model and the \emph{most powerful model} (MPM) in the cascade. 
Intuitively, when the confidence of a small model is high, we can safely stop the inference, saving MPM's inference cost while maintaining its accuracy. In such cases, it is highly likely that both the cheap model and the MPM would be correct. Additionally, when both models are wrong and we still exit early, we accept a controlled loss in accuracy in exchange for a significant cost reduction. Only ambiguous cases where the cheap model is expected to disagree with MPM should be forwarded to the next model in the cascade.
C3PO makes three key technical contributions:
\begin{itemize}[leftmargin=*,align=left]
\item \textbf{Data‐efficient cascade learning.} We show that cascade decision rules can be learned using only a small “self‐supervision” pool of unlabeled prompts—fewer than 1\% of the examples used by \emph{state‐of‐the‐art} methods such as TREACLE \citep{zhang2024treacle}. This self‐supervised approach requires no ground‐truth labels, enabling easy adaptation to new domains and task distributions.
\item \textbf{Rigorous theoretical guarantees.} We derive conformal cost bounds that with arbitrary probability control the cascade’s inference cost under a user-specified budget. In parallel, we establish PAC‐Bayesian generalization bounds certifying that the learned cascade decision rules generalize.
\item \textbf{\emph{State‐of‐the‐art} empirical performance.} C3PO consistently outperforms across a diverse suite of reasoning benchmarks including arithmetic (GSM8K, SVAMP), formal mathematical (MATH-500, AIME), and logical reasoning tasks (CommonSenseQA, BIG‐Bench-Hard).
\end{itemize}
\begin{figure}[t]
    \centering
    \includegraphics[trim={0 1cm 0 0},clip, width=0.35\linewidth ]{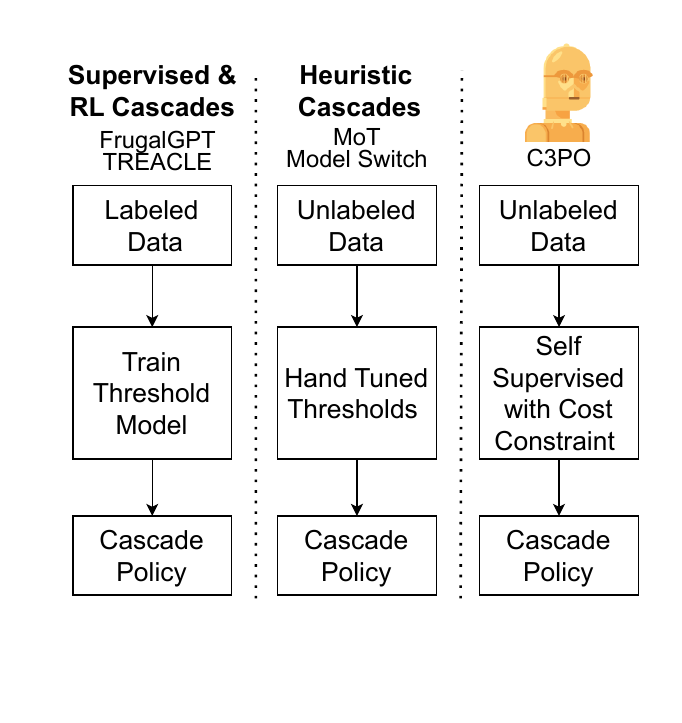}
\includegraphics[trim={0 1.15cm 0 0},clip, width=0.64\linewidth]{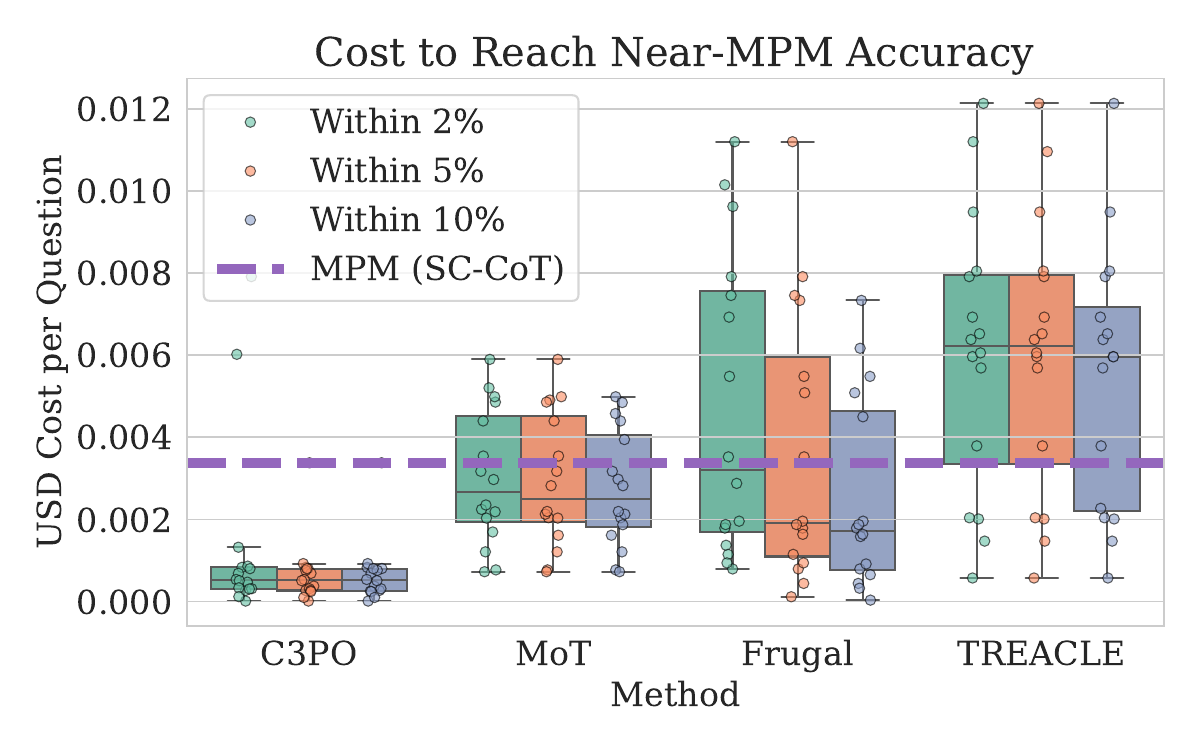}
\vspace{-0.5cm}
    \caption{Left: Existing LLM cascading strategies include supervised learning, reinforcement learning, and learning-free heuristics. In contrast, C3PO proposes a novel, cost-constrained, self supervised (label-free) cascade paradigm.
    Right: Surpassing existing cascade approaches tremendously, C3PO offers markedly superior cost-effectiveness across 16 benchmarks, requiring \textbf{less than 20\% of the cost of the most powerful model (MPM)} (cost shown in purple) for an accuracy gap of at most 2, 5, and 10\% using a LLAMA cascade. In this boxplot, each dot represents a dataset and the whiskers extend to 90\% coverage.}
    \label{fig:boxplot}
\end{figure}

\section{Related Work}

\textbf{LLM Cascading} is a structured inference paradigm in which language models are queried in a sequential manner based on their complexity and cost. The goal is to minimize inference costs while maintaining high answer quality. 
FrugalGPT~\citep{chen2024frugalgpt} first demonstrated the efficacy of cascading in reducing LLM inference costs by predicting the likelihood that an LLM output is correct using a BERT-like meta-model. If this prediction exceeds a user-defined threshold, the model's output is returned as the final answer; otherwise, the next, more capable model is invoked. \citet{zhang2023ecoassistant} extend this idea by incorporating LLM-based self-verification before escalating to a more powerful model, though this introduces additional delays due to sequential dependencies. AutoMix~\citep{aggarwal2024automix} leverages self-consistency sampling to decide whether to defer to stronger LLMs, learning thresholds over sampled predictions to optimize for quality and cost. Similarly, Mixture-of-Thoughts (MoT)~\citep{yue2024large} integrates external programmatic solvers and majority voting to assess the adequacy of weak model responses before escalation.
More recent approaches such as ModelSwitch~\citep{chen2025we} employ self-consistency as a decision criterion to escalate queries. If samples from the current LLM are inconsistent, a larger model is queried, and a final decision is made via voting over all collected samples. Our method differs crucially from other self consistency based approaches by learning to map self consistency scores to agreement probabilities with the MPM rather than with the ground truth. This eliminates the need for dataset labels and simplifies the cascade learning greatly as we do not need many samples to understand LLM output correlations.
TREACLE~\citep{zhang2024treacle} and Online Cascade~\citep{nie2024cascade} approach the cascade learning problem using reinforcement and imitation learning, respectively. TREACLE trains a deep Q-network to learn cascade decisions, while Online Cascade distills knowledge from stronger models into weaker ones. However, both methods rely on large supervised training sets and are less practical in label-scarce environments. Additionally, Online Cascade requires fine tuning of model parameters, which is very costly.

\textbf{LLM routing approaches} are another prominent class of multi-LLM inference, but they fall outside the scope of this work. Routers select the best LLM \emph{prior to inference}, often relying on learned or heuristic models to choose among available options based solely on input characteristics. In contrast, cascades defer this decision until {\em after} an LLM has generated a response, leveraging internal confidence signals to determine whether to exit or escalate. Routing methods like RouteLLM~\citep{ong2025routellm}, SelectLLM~\citep{maurya2024selectllm}, and Symbolic-MoE~\citep{chen2025symbolic} require task-specific training data and often struggle to generalize beyond their training distribution. Cascades, by contrast, offer greater flexibility by basing decisions on the output behavior of models rather than input features alone, at the price of multiple LLM calls per query.

\section{Problem Statement}

We consider a system comprised of $m$ large language models (LLMs), denoted $\mathcal{M} = \{\mathcal{M}_1, \mathcal{M}_2, \dots, \mathcal{M}_m\}$, each parameterized by a neural conditional generative model $f_j : \mathcal{X}\mapsto \mathcal{A}$ which maps from the prompt space $\mathcal{X}$ to the answer space $\mathcal{A}$.
Additionally, we assume that the $j$-th model also returns a confidence score $s_j \in [0, 1]$ along with its answer $\hat{y}_j$. 
We impose a mild correlation assumption on $s_j$ that it is stochastically increasing as the probability of correctness of $\hat{y}_j$ increases.
The models are arranged in the order of increasing inference cost. That is, $\mathcal{M}_1$ is the cheapest, while $\mathcal{M}_m$ (MPM) is the most expensive, and $\mathrm{cost}(\mathcal{M}_i) \leq \mathrm{cost}(\mathcal{M}_j)$ for $i<j$. For an input $x$ drawn i.i.d. from a distribution $\mathcal{X}$, model $\mathcal{M}_j$ produces a prediction $\hat{y}_j$, which is evaluated using the 0-1 regret with respect to the most powerful model's prediction $\hat{y}_m$:  
\begin{align}
    \ell\big(\hat{y}_j, \hat{y}_m\big) =
    \begin{cases}
        0, & \text{if } \hat{y}_j = \hat{y}_m, \\
        1, & \text{otherwise}.
    \end{cases}
    \label{eq:regret}
\end{align}
Here, we are operating in an unsupervised setting without access to the true labels, so the regret is formulated with respect to the most powerful LLM's answer. That is, we never have access to the ground truth $y$, but only to the predictions of the $m$ models $\{\hat{y}_j\}_{j=1}^{m}$.

To reduce inference cost while maintaining high accuracy, we employ a cascade decision rule. Given an input $x$, the cascade begins by querying the first model, $\mathcal{M}_1$. After obtaining the output from $\mathcal{M}_j$, the system must decide whether to exit early and return $\hat{y}_j$ if it is highly likely to match $\hat{y}_m$, or to escalate the query to $\mathcal{M}_{j+1}$ if the current output is deemed unreliable. This process continues until an early exit occurs or the query $x$ reaches $\mathcal{M}_m$, whose answer is always used if no earlier model qualifies. Following empirical results from~\citet{yue2024large}, we do not include outputs from model $\mathcal{M}_{j}$ when prompting model $\mathcal{M}_{j+1}$, as it can potentially increase the error rate and prompting costs.
Each model $\mathcal{M}_j$ incurs an inference cost $c_j$, satisfying  $c_1 \leq c_2 \leq \cdots \leq c_m$. The cost associated with querying a model reflects monetary API fees, inference latency, or a combination thereof. We use a standard cost model from the literature that mimics real-world API cost structures~\citep{chen2024frugalgpt}. 
The costs $\{c_j\}_{j=1}^m$ can be assumed to be fixed, known a priori, and query-independent for our analysis, since we know the per-token costs for different LLMs, the typical prompts and response lengths, and the latency.
If the cascade stops at model $\mathcal{M}_{z}$, the total inference cost incurred is $    \mathrm{Cost} = \sum_{k=1}^{z} c_k$.

The objective is to design a cascade selection strategy that minimizes the total expected loss, measured as the regret relative to the best model $\mathcal{M}_m$ (Eq.~\eqref{eq:regret}), while ensuring that the expected inference cost exceeds some prespecified budget $C^*$ with a probability less than or equal to a fixed, small number $\alpha$. Here $\alpha \in (0, 1)$ denotes the maximally allowed cost-constraint violation rate.

We model the binary decision to exit the cascade after querying   $\mathcal{M}_j$ by a learnable thresholding rule applied on $s_j$. Let $\boldsymbol{\tau}\eqdef[\tau_1, \dots, \tau_{m-1}, 0]$ be a vector of learnable thresholds of confidence scores $\mathbf{S}\eqdef[s_1, \dots, s_{m-1}, 1]$ such that each model $\mathcal{M}_j$ has its own learnable confidence score threshold $\tau_j \in [0, 1]$. When $s_j$ takes a value above the critical threshold $\tau_j$ this leads to an exit decision at $\mathcal{M}_j$. Note that for the last model in the cascade, $\tau_m=0$ and $s_m=1$. Thus the model always exits, as it is impossible to defer. We define the exit point of the cascade as the first model index for which the model confidence is greater than the threshold: $z(\mathbf{S}, \boldsymbol{\tau}) = \min_j \{j : s_j\geq \tau_j \}.$

To learn the optimal thresholds, we are given $N$ questions sampled from $\mathcal{X}$ for which we do not require ground truth labels. In addition, we have access to sampled responses and their confidence scores from each model for each of these $N$ questions. We call these sample questions and LLM outputs the dataset $\mathcal{D} 
\eqdef \{x^i, (\hat{y}_1^i, s_1^i ), \dots, (\hat{y}_m^i, s_m^i ) \}_{i=1}^N$.
At test time, the cascade output is evaluated by the percent agreement with the ground truth labels of a held-out dataset (accuracy). Thus, the primary learning objective is to solve the following optimization problem:
\begin{equation}\label{eq:obj}
\min_{\boldsymbol{\tau}} L(\boldsymbol{\tau}), \text{where } L(\boldsymbol{\tau}) \eqdef \mathbb{E}_{X} \left[ \mathbb{E}_{(\hat{Y}_1, S_1), \dots, (\hat{Y}_m, S_m) \mid X} \, \ell\Big(\hat{Y}_{z(\mathbf{S}, \boldsymbol{\tau})}, \hat{Y}_m\Big) \right]\,,
\end{equation}
subject to a cost constraint. This constraint is soft, in the sense that the realized cost of answering a random query $X$ violating the allowable budget $C^*$ can be tolerated provided that such occurrences can only happen with a probability not exceeding $\alpha$. This constraint is formally specified as:
\begin{equation}\label{eq:cost_obj}
\mathbb{E}_{X} \left[ \mathbb{E}_{\mathbf{S}|X} \bigg[\mathds{1}\left\{\sum_{k=1}^{z(\mathbf{S}, \boldsymbol{\tau} )} c_k > C^* \right\} \bigg]\right] \leqslant \alpha.
\end{equation}

\section{Methodology}

We propose a principled methodology for optimizing exit thresholds in a LLM cascade, subject to a soft constraint on inference costs. The approach consists of three key components: (i) optimizing over thresholds to minimize empirical regret; (ii) conformal prediction-based adjustment for probabilistic cost control; and (iii) establishing generalization guarantees using PAC-Bayesian analysis.

To ensure our optimization and conformal guarantees are valid, we partition the dataset $\mathcal{D}$ into the self-supervision data, $\mathcal{D}_{\mathrm{SS}}$, which are used to learn the thresholds, and calibration data, $\mathcal{D}_{\mathrm{Cal}}$, which are used for evaluating whether a particular threshold vector satisfies the conformal cost guarantee. Thus, we have two non-overlapping splits of  the dataset $\mathcal{D} = \mathcal{D}_{\mathrm{SS}} \cup \mathcal{D}_{\mathrm{Cal}}\eqdef \{x^i, (\hat{y}_1^i, s_1^i ), \dots, (\hat{y}_m^i, s_m^i ) \}_{i=1}^{N_{\mathrm{SS}} + N_{\mathrm{Cal}}}$ with $N = N_{\mathrm{SS}} + N_{\mathrm{Cal}}$ and $\mathcal{D}_{\mathrm{SS}} \cap \mathcal{D}_{\mathrm{Cal}}=\varnothing$. Furthermore, we define an empirical approximation of~\Cref{eq:obj} using $\mathcal{D}_{\mathrm{SS}}$ as: 
\begin{align}\label{eq:empirical}
    \widehat{L}(\boldsymbol{\tau}) = \frac{1}{N_{\mathrm{SS}}}\sum_{i=1}^{N_{\mathrm{SS}}}\ell\left(\hat{y}_{z(\mathbf{S}^i, \boldsymbol{\tau})}, \hat{y}_m\right).
\end{align}

\subsection{Discretization}
The empirical 0--1 regret $\widehat L(\boldsymbol{\tau})$ and cumulative cost are piecewise-constant functions of $\boldsymbol{\tau}$, with discontinuities occurring only when a threshold $\tau_j$ crosses a confidence score $s_j$ in the finite dataset $\mathcal{D}_{\mathrm{SS}}$. This structure arises because, while $s_j$ may theoretically take continuous values, the dataset $\mathcal{D}_{\mathrm{SS}}$ contains at most $N_{\mathrm{SS}}$ distinct confidence scores per model. Sorting these scores for $\mathcal{M}_j$ yields $N_{\mathrm{SS}}+1$ non-overlapping intervals, and varying  $\tau_j$ within each such interval, while keeping all other thresholds fixed, does not alter the set of examples that exit at $\mathcal{M}_j$. As a result, $\widehat L(\boldsymbol{\tau})$ and the cost remain constant within each interval; changes occur only at the empirical confidence values $\{s_j^{(i)}\}_{i=1}^{N_{\mathrm{SS}}}$.  
Therefore, the loss surface is flat almost everywhere (with the exception of sampled $s_j$ values) rendering gradients zero and gradient-based optimization ineffective. 

\paragraph{Grid Setup} For each model $\mathcal{M}_j$ (except for the MPM), we construct a candidate grid of thresholds defined as:
$\mathcal{T}_j = \{ \frac{k}{K-2} \;|\; k \in \{0, 1, \dots, K-1\} \}$ with integer $K$ defining the grid resolution. Note that this grid contains the edge points 0 and $(K-1)/(K-2)$, potentially allow the cascade system to learn to always exit at a particular model, or to always skip it.
This is useful for handling edge cases. For example, when the reasoning problems are trivial, we would wish to always exit at the first model. Alternatively, when a collection of weaker models produce effectively useless predictions, the cascade would perform better by discarding those models entirely.

\paragraph{Optimizing the cascade thresholds} Due to the non-differentiable 0--1 loss, the optimization is performed by an exhaustive or heuristic search over all possible combinations of thresholds $\boldsymbol{\tau}$ selected from discrete grids. Note that this is usually not computationally unreasonable, since we are not considering cascades of dozens of LLMs. For each candidate threshold configuration, the cascade is evaluated on $\mathcal{D}_{\mathrm{SS}}$ to compute both the average 0--1 loss and the average cumulative cost. The configuration that empirically minimizes $\mathcal{L}(\boldsymbol{\tau})$, while satisfying the cost constraint, is selected as optimal. 
Pseudocode is provided in Algorithm~\ref{alg:brute-force-quantile}.

\subsection{Efficient Conformal Search Strategy}
From a practical point of view, a brute-force search can be a viable approach in a setting with fewer than 5 models in the cascade.
For example, in our experiments, searching for the optimal threshold using a brute-force search for a cascade of 4 LLMs with 10 discrete threshold levels for each of the first 3 of them, takes approximately 0.01 seconds on a M3 CPU for GSM8K using 50 questions in $\mathcal{D}_{\mathrm{SS}}$. Compared to the inference time of LLMs response generation, this is a small amount of time.
Cascades are unlikely to be considerably larger than this in practice.

\paragraph{Selecting Grid Resolution under Statistical Indistinguishability}

The finite calibration size $N_{\mathrm{SS}}$ induces a statistical resolution limit. Given $N_{\mathrm{SS}}$ samples, differences in empirical regret smaller than $O(1/\sqrt{N_{\mathrm{SS}}})$ cannot be distinguished using an appropriate statistical test (proof in App.~\ref{app:grid}).
Thus, over‑refinement of the grid does not offer any empirical benefit, yet inflates the size of hypothesis class and degrades the theoretical generalization guarantee. 
We empirically observed that using <10 grid points for each $\tau_j$ works in our experiments, and using more points is not beneficial.

\paragraph{Grid-Search with Quantile Check}
To enforce the probabilistic cost constraint $\Pr(C_{\mathrm{test}} \leqslant C^*) \geqslant (1{-}\alpha)$ during threshold optimization, we embed a quantile filter within the grid search.  For each candidate configuration $\boldsymbol{\tau}$, we evaluate the cascade on $\mathcal{D}_{\mathrm{Cal}}$ to compute the cumulative cost $C^i$ for each $i=1,\dots,N_{\mathrm{Cal}}$.
We then sort the costs in the ascending order to obtain order statistics $C_{(1)}\leqslant\cdots\leqslant C_{(N)}$. We accept $\boldsymbol{\tau}$ as a valid candidate to be considered as a cost-constrained minimizer of the regret over $\mathcal{D}_{\mathrm{SS}}$, if and only if the empirical $(1{-}\alpha)$ cost-quantile $q_{1{-}\alpha}\le C^*$. 
Among all accepted configurations, we select the $\boldsymbol{\tau}$  with the minimal empirical regret $\widehat L$ on $\mathcal{D}_{\mathrm{SS}}$.  As shown in~\Cref{thm:conformal}, this ensures that the probabilistic cost-constraint is satisfied for a test query.  

\begin{algorithm}[t]
\caption{Grid Search with Quantile Check}\label{alg:brute-force-quantile}
\begin{algorithmic}[1]
\Require Grids $\prod_{j=1}^{m-1}\mathcal{T}_j$, self‐supervision data $\mathcal{D}_{\mathrm{SS}}$, calibration set $\mathcal{D}_{\mathrm{Cal}}$, budget $C^*$, risk level $\alpha$
\State bestRegret $\leftarrow \infty$, bestTau $\leftarrow$ $(0,0,\dots,0)$
\ForAll{$(\tau_1,\tau_2,\dots,\tau_{m-1}) \in \mathcal{T}_1 \times \cdots \times \mathcal{T}_{m-1}$}
  \State Assemble threshold vector $\boldsymbol{\tau}$
  \State $\widehat L \leftarrow $Evaluate ~\cref{eq:empirical} on $\mathcal{D}_{\mathrm{SS}}$
  \State Evaluate all instances in $\mathcal{D}_{\mathrm{Cal}}$ and store the cascade costs $\{C_j\}$.
  \State Sort $\{C_j\}$ to $C_{(1)} \le \cdots \le C_{(N)}$
  \State $k \leftarrow \lceil (N+1)(1-\alpha)\rceil$
  \State $q_{1-\alpha}\leftarrow C_{(k)}$
  \If{$q_{1-\alpha}\le C^*$ and $\widehat L<\text{bestRegret}$}
    \State bestRegret $\leftarrow \widehat L$
    \State bestTau $\leftarrow \boldsymbol{\tau}$
  \EndIf
\EndFor
\State \Return bestTau
\end{algorithmic}
\end{algorithm}
\subsection{Theoretical Results}  
In this section, we provide important theoretical guarantees for our algorithm. First, in~\Cref{thm:conformal}, we show that our search algorithm only returns solutions that are provably under the budget (if any such solutions exist). Second, in~\Cref{thm:generalization}, we examine the generalization error of our algorithm using PAC-Bayes theory. 

\begin{theorem}[Conformal Cost Guarantee]\label{thm:conformal}
Let $\boldsymbol\tau$ be the thresholds and $\mathcal{D}_{\mathrm{Cal}}$ denote a calibration set containing $N_{\mathrm{Cal}}$ questions and the cascade answers, obtained using 
 thresholds $\boldsymbol\tau$. Sort the costs of the cascade on the calibration dataset and define the rank of the budget $C^*$ as $k\eqdef \min\{p: C_{(p-1)} \leqslant  C^* \leqslant C_{(p)}\}$. If $k \geqslant \lceil (N_{Cal}+1)(1 - \alpha) \rceil$ is satisfied, then the inference cost $C_{\mathrm{test}}$ for a new query can be bounded under exchangeability of calibration and test data with 
$\Pr(C_{\mathrm{test}} > C^*) \leqslant \alpha$.
\end{theorem}

\begin{proof}
See App.~\ref{app:conformal}.
\end{proof}

\Cref{thm:conformal} demonstrates that, assuming exchangeability of samples from $\mathcal{D}_{\mathrm{Cal}}$  and the test set, any new test query can only exceed the budget constraint with probability at most $\alpha$. In App.~\ref{sec:stochastic_cost} we show how the results of Theorem 1 can be extended when the costs are not assumed fixed and known \textit{a priori}.

In~\Cref{thm:generalization}, we provide PAC-Bayes analysis that bounds the excess test regret due to the evaluation on the learned thresholds from $\mathcal{D}_{\mathrm{SS}}$ in terms of the number of models, grid-size, and the number of samples used to minimize the empirical regret. 
\begin{theorem}[Generalization Bound]\label{thm:generalization}
Let $\mathcal{H}= \prod_{j=1}^{m-1}\mathcal{T}_j$ denote the hypothesis class of all threshold combinations with $|\mathcal{H}| = K^{(m-1)}$, where $K$ denotes the grid-size and $m$ is the number of LLMs in the cascade.
Let $\mathcal{H}_c \subset \mathcal{H}$ be the subset of thresholds for which the cost-constraint in Eq.~\eqref{eq:cost_obj} is satisfied and $\tau^*$ is the learned threshold vector from C3PO using $\mathcal{D_{\mathrm{SS}}}$ and $\mathcal{D_{\mathrm{Cal}}}$.  Then, for any $\delta \in (0, 1)$, with probability at least $(1-\delta)$, we have 
\begin{equation}
L(\boldsymbol{\tau}^*) \leqslant \min_{\boldsymbol{\tau} \in \mathcal{H}_c} L(\boldsymbol{\tau}) + 2\sqrt{\dfrac{(m-1) \log K - \log \delta}{2N_{\mathrm{SS}}}}.
\end{equation}
\end{theorem}
\begin{proof}
See App.~\ref{app:generalization}.
\end{proof}
\paragraph{Discussion} As a concrete example, take $m=3$, $K=10$, $N_{\mathrm{SS}}=150$, and $\delta=0.05$. Then
$
\log|\mathcal{H}| = 2\log 10 \approx 4.61, $ and $\log(1/\delta) \approx 2.99,
$
so the upper bound on excess regret evaluates to
\begin{equation}
\epsilon = \sqrt{\frac{4.61 + 2.99}{300}} \approx 0.159.
\end{equation}
Thus, the test regret w.r.t. to the answer of the MPM in the cascade can only exceed the minimum regret one could possibly attain while satisfying the cost criterion on the test set, by at most $2\times0.159$ with more than 95\% certainty. In other words, the percentage of test questions for which C3PO's answer does not agree with the MPM, but an agreement with MPM can be obtained by solving the optimization problem we propose directly on the test-set under the same budget constraint, cannot exceed 32\% with probability more than 0.95.
Note that, solving the optimization problem directly on the test-set is impractical, since we would need to query all LLMs for each test question to perform the grid search, defeating the purpose of cascade learning using cost-constrained optimization. 

The generalization bound degrades logarithmically with $K$ and linearly in $m{-}1$, and improves at the rate $O(1/\sqrt{N_{\mathrm{SS}}})$. This shows a clear tradeoff. Increasing the grid resolution $K$ or the number of cascade stages $m$ improves fitting of the training data during optimization but may adversely affect generalization guarantees if $N_{\mathrm{SS}}$ remains constant. 
Finally, note that the guarantee requires the calibration set to be i.i.d. from the data distribution and assumes that the hypothesis class $\mathcal{H}$ is fixed prior to observing the calibration data. Under these conditions, we obtain a remarkable result: the empirical risk minimizer over a \textit{combinatorially large configuration space} enjoys \textit{high-probability generalization guarantees with exponential concentration}.

\section{Experiments}

\subsection{Datasets}
We evaluate our method across a diverse suite of 16 datasets spanning three distinct reasoning categories: logical reasoning, arithmetic reasoning, and mathematical reasoning. The \textbf{logical reasoning} group includes tasks designed to assess abstract and commonsense inference capabilities, such as \textit{CommonSenseQA}~\citep{talmor2018commonsenseqa}, and \textit{BIG-Bench-Hard}~\citep{suzgun2023bbh} tasks, including \textit{causal judgement, date understanding, disambiguationQA, formal fallacies, geometric shapes, movie recommendation, penguins, ruin names, snarks, sports, and temporal sequences}. These benchmarks emphasize multi-step deduction, ambiguity resolution, and contextual judgment. In contrast, the \textbf{arithmetic reasoning} group contains datasets like \textit{GSM8K}~\citep{cobbe2021gsm8k}, \textit{SVAMP}~\citep{patel2021}, and \textit{AQuA}~\citep{ling2017}, which require basic numerical and word-problem-solving skills, often through step-by-step calculation. Finally, the \textbf{mathematical reasoning} category targets advanced problem-solving ability, represented by the \textit{MATH-500}~\citep{hendrycksmath2021} and \textit{AIME}~\citep{aime2024} datasets, which consist of high-school level and competition-style math questions that demand rigorous algebraic and geometric manipulation. These datasets enable a nuanced assessment of reasoning capabilities across domains. All of these benchmarks are available under open-source licenses~\footnote{Apache 2.0 [AQuA], MIT [CommonSenseQA; SVAMP; GSM8K; MATH; BIG-Bench Hard], CC0 [AIME].}. We make our code available~\footnote{\url{https://github.com/AntonValk/C3PO-LLM}}.

\subsection{Models}\label{ssec: models}
We conduct experiments on both open and closed source LLMs. We use the following model families:\\
\textbf{LLAMA}: The open weights Meta models comprise Llama 3.2-1B-Instruct, Llama 3.2-3B-Instruct, Llama 3.3-70B-Instruct, and Llama-3.1-405B-Instruct. These models have been fine-tuned for instruction following and exhibit competitive performance on reasoning tasks.\\ 
\textbf{QWEN}: The QWEN family includes Qwen2.5-1.5B-Instruct, Qwen2.5-32B-Instruct, and Qwen2.5-72B-Instruct. These open source models balance efficiency with advanced reasoning capabilities.\\
\textbf{GPT}: The OpenAI model family includes GPT 3.5 Turbo, GPT-4o-mini, and o3-mini. These models are renowned for their strong few-shot learning capabilities and robust chain-of-thought reasoning. In our experiments, they serve as a benchmark for high-performance proprietary LLMs.

\subsection{Baselines}

We compare C3PO with several existing cascade LLM approaches from the literature. 
\textbf{FrugalGPT}~\citep{chen2024frugalgpt} learns the probability of correctness of a solution using a lightweight, supervised training procedure from DistilBert embedding of the question and output CoT. Subsequently, it employs a threshold decision rule for cascading several LLMs. 
\textbf{Mixture of Thoughts (MoT)}~\citep{yue2024large} is an unsupervised cascading framework involving a weak and a strong LLM, which rely on an LLM's confidence in the most frequent answer when a single prompt is provided. MoT proposes a family of similar cascade architectures. We select the specific method that matches our setting: CoT-1D-Vote.
\textbf{TREACLE}~\citep{zhang2024treacle} learns a reinforcement learning (RL)-based decision rule tailored for LLM reasoning, which requires true labels for a subset of questions for Deep Q-Network (DQN) training. The original TREACLE architecture includes a prompt adaptation component; we remove this in our implementation to adapt to our setting where all models are evaluated on fixed prompts.
\textbf{Model Switch}~\citep{chen2025we} is an unsupervised early exit module with the option to defer to a more elaborate model. If no model is sufficiently confident in its prediction, the model outputs a weighted ensemble prediction.

Note that all our baselines are applicable to the cascade setting, where we assume that routing decisions are made \emph{after} generating a candidate response. 
This is a different setting from the routing methods that must make routing decisions before seeing candidate outputs. 

\begin{figure}[t]
    \centering
\includegraphics[width=\linewidth]{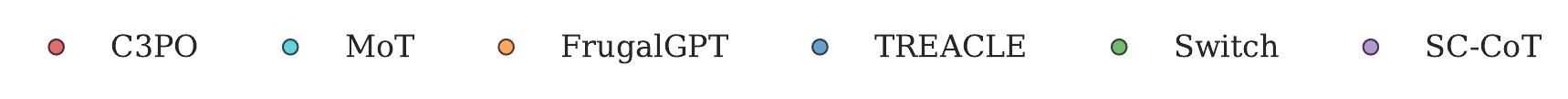}
\includegraphics[width=0.3\linewidth]{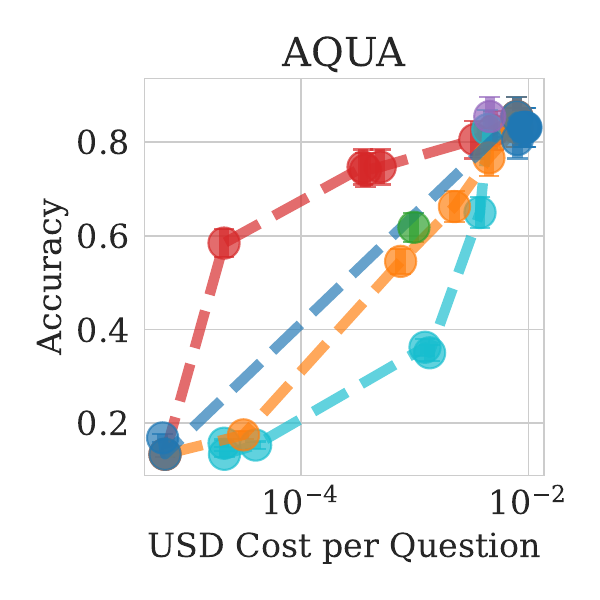}
\includegraphics[width=0.3\linewidth]{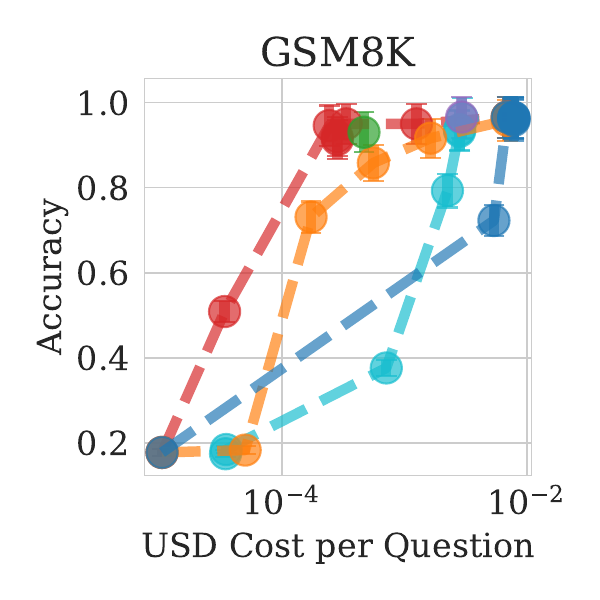}
\includegraphics[width=0.3\linewidth]{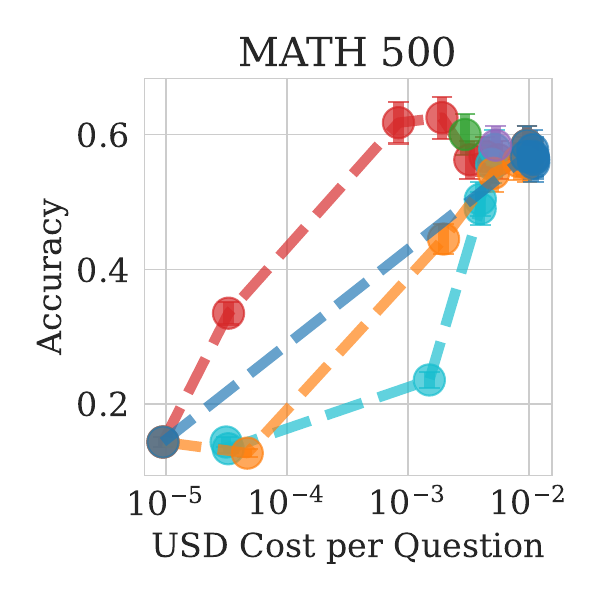}
\includegraphics[width=0.3\linewidth]{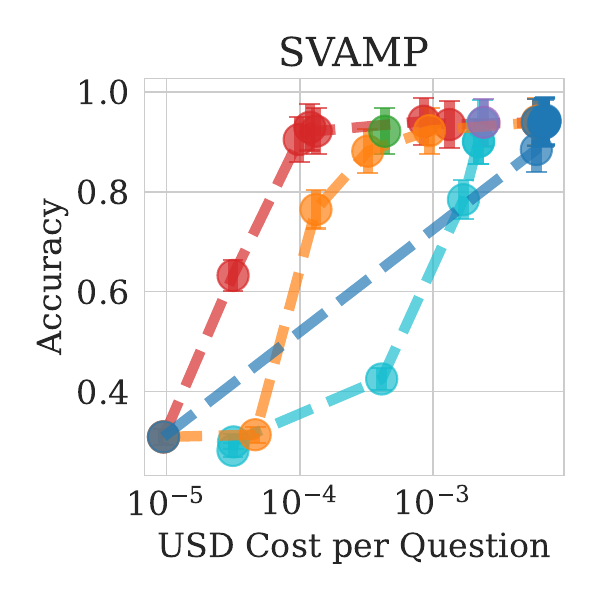}
\includegraphics[width=0.3\linewidth]{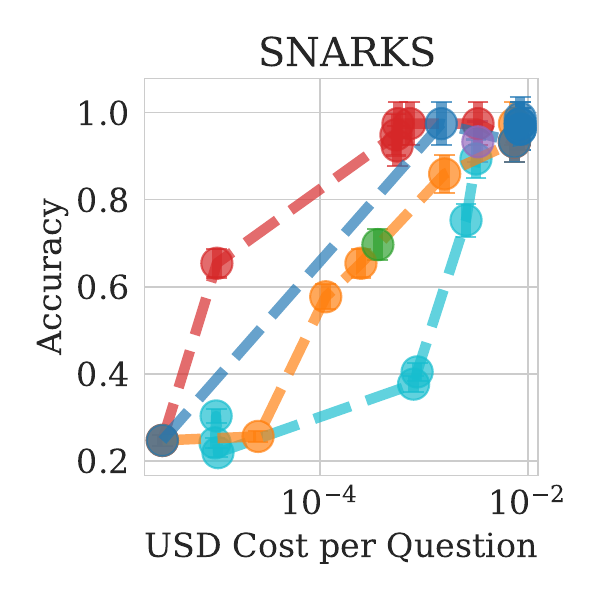}
\includegraphics[width=0.3\linewidth]{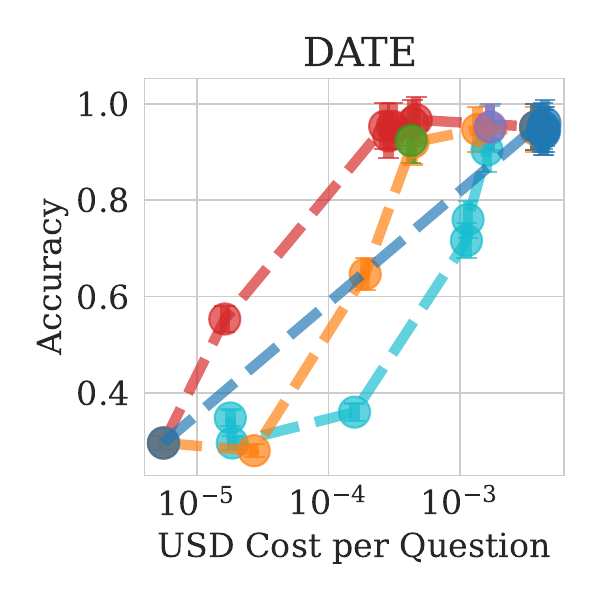}
    \caption{C3PO achieves the best performance for a wide range of cost budgets on the LLAMA cascade. Error bars denote 90\% confidence interval. GPT and QWEN cascade results in App.~\ref{app:full_results}.}
    \label{fig:date}
\end{figure}

 \subsection{Results}
\paragraph{Can C3PO outperform existing cascades at different budgets?}

We conduct experiments by organizing the models in the three cascades listed in~\Cref{ssec: models}: the LLAMA cascade which consists of four LLAMA-3 models of 1B, 3B, 70B and 405B variants, the QWEN-2.5 cascade 1.5B, 32B, 72B models  and the GPT cascade consisting of GPT 3.5 Turbo, GPT-4o-mini and GPT-4. All models are prompted with 8-shot examples and during evaluation output 5 CoT samples. For all cascades we conduct the same experiment independently. Each time, for a given dataset, we deploy C3PO and the baselines with a training set of $100$ reasoning problems. Unsupervised methods such as C3PO, Mixture-of-Thoughts, and ModelSwitch receive only the questions and 5 CoT outputs from each LLM per question. Supervised methods such as FrugalGPT and TREACLE also receive the ground truth labels to the reasoning problems. The seeds are fixed such that all methods receive the same questions and sample identical CoTs during training and testing. Additionally, the few shot examples in the prompts are identical for all methods. To evaluate the methods we compare them across a range of given budgets. The range extends from trivially using the cheapest model of the cascade with 5 CoT samples, all the way up to the cost equivalent to going through the entire cascade for all questions with 5 CoT samples per model. We note that since TREACLE can choose to early exit with an arbitrary number of CoT samples it sometimes chooses to exit with fewer than 5 CoTs. It can thus choose to use even lower than the minimal budget in some easy cases.

\begin{figure}[ht!]
    \centering
    \includegraphics[width=0.9\linewidth, trim={0cm 0 5cm 0},clip]{figs/method_legend.pdf}
    \includegraphics[width=0.3\linewidth]{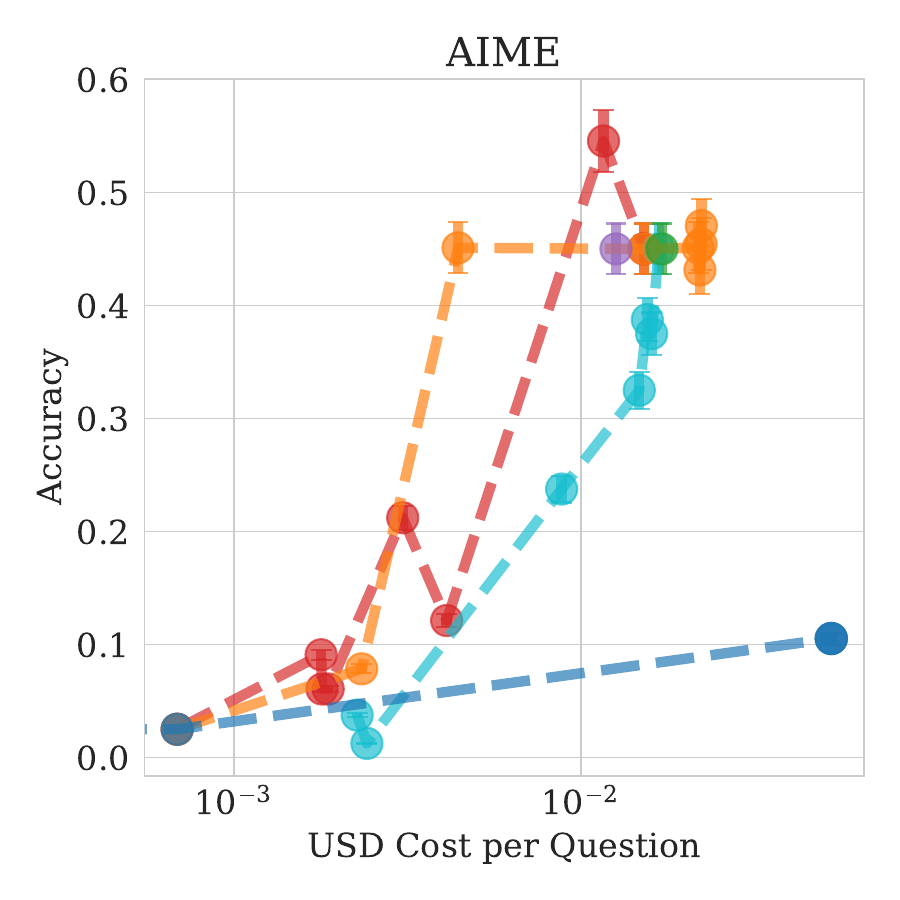}
    \includegraphics[width=0.5\linewidth]{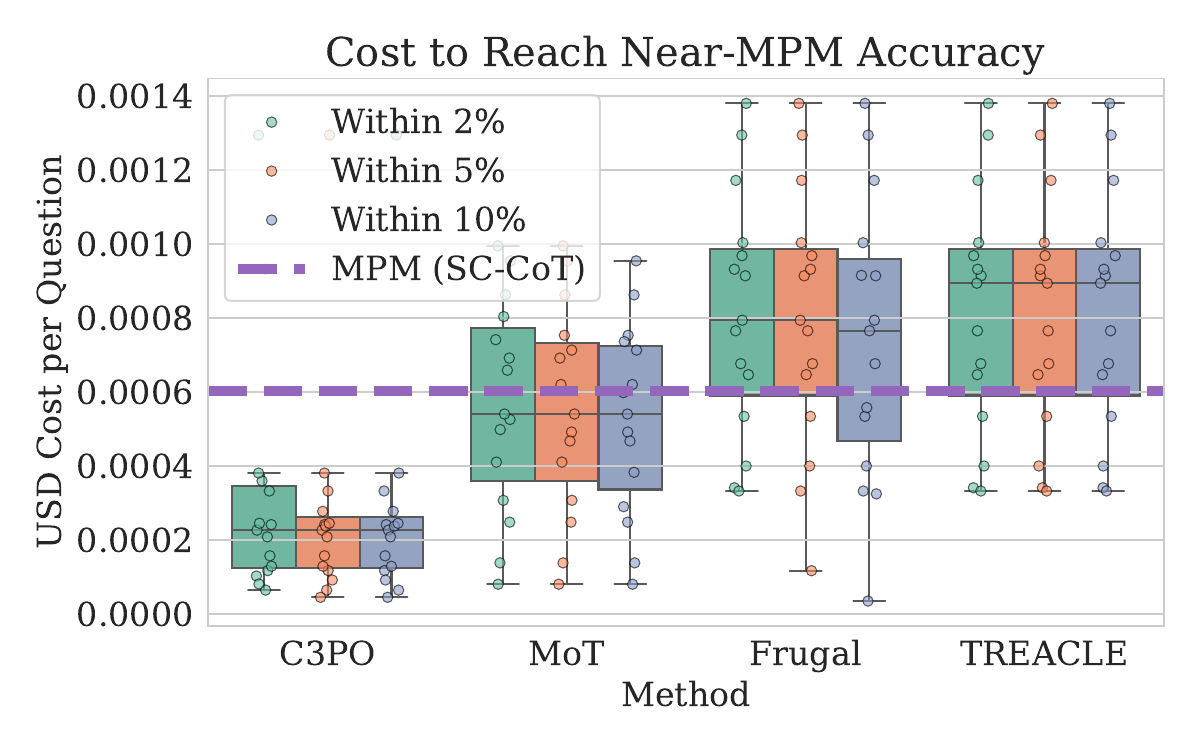}
    \caption{Left: Results of GPT cascade on AIME. Right: Here we see that a mixed model family cascade comprised of LLaMA 3.2 1B-Instruct, Qwen 2.5 32B-Instruct and GPT 4o-mini operates with roughly the same performance as cascades of models comprised within the same family. Therefore, we conclude that the method is robust to LLM family selection.}
    \label{fig:mixed_model_family}
\end{figure}

The results for the LLAMA cascade for AQuA, GSM8K, MATH-500, SVAMP and a subset of BigBenchHard (SNARKS, DATE) are depicted in Fig.~\ref{fig:date}. AIME results are on Fig.~\ref{fig:mixed_model_family} (left). We observe that in all cases our model outperforms for most cost configurations. \textbf{Notably, for large datasets such as SVAMP and MATH-500 we obtain the top accuracy performance at 10 times lower the cost compared to other cascade baselines.} The performance for other datasets and the QWEN and GPT cascades is qualitatively similar and reported in App.~\ref{app:qwen} and App.~\ref{app:gpt} respectively.
\begin{figure}[t]
    \centering
    \includegraphics[width=0.9\linewidth, trim={0cm 0 5cm 0},clip]{figs/method_legend.pdf}
\includegraphics[width=0.8\linewidth]{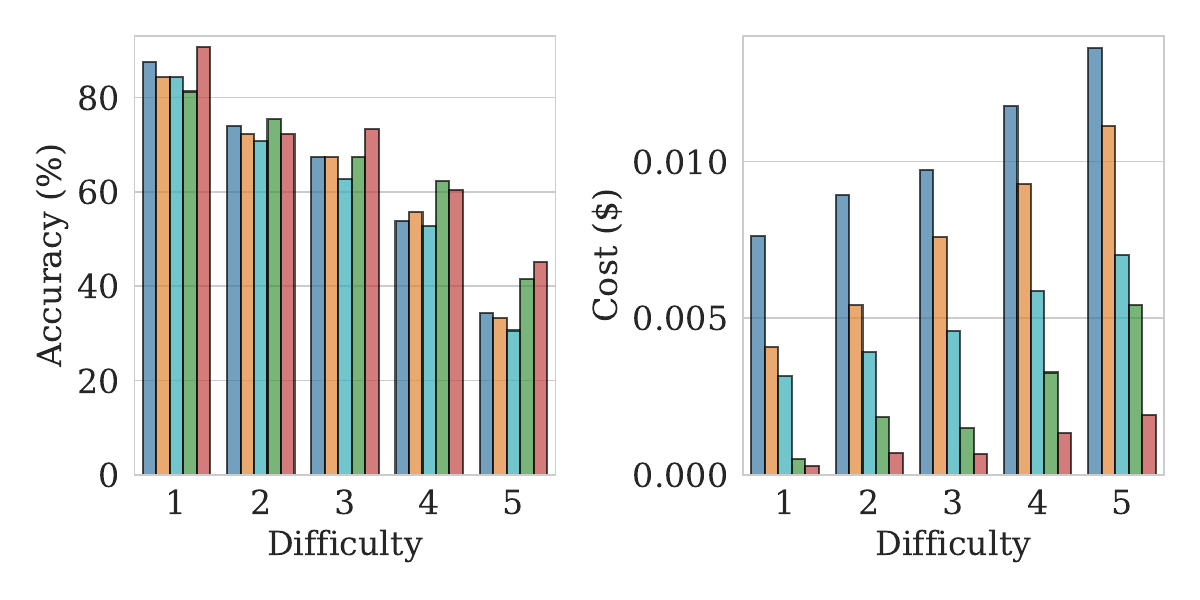}
    \caption{Average accuracies and dollar costs of different algorithms for five different difficulty levels of the MATH-500 dataset using the LLAMA cascade.}
    \label{fig:difficulty_MATH 500_llama}
\end{figure}
\paragraph{How does C3PO compare to Self-Consistency for a comparable maximum allowable budget?} 
From Fig.~\ref{fig:date}, we observe that in all cases, C3PO achieves an accuracy, which is either comparable to or higher than that of self-consistency using the MPM at a fraction of the incurred cost. For example,  from Fig.~\ref{fig:date}, we observe that \textbf{on MATH-500, C-3PO achieves 62.5\% accuracy with cost of 0.0019 USD/question, whereas SC using MPM manages to obtain 57\% accuracy at a cost of 0.0053 USD/question.}
This comparison validates the core philosophy of employing a LLM-cascade as an effective cost-saving alternative to performing self-consistency using a strong LLM.

\paragraph{Does C3PO actually satisfy the theoretical conformal guarantee for cost and the generalization bound?}

To evaluate whether C3PO satisfies its theoretical guarantee on inference cost,
we consider a collection of 15 datasets, 2 different cascade configurations (using LLAMA and Qwen models), 5 different target budget ($C^*$) levels for each cascade, and two different levels of $\alpha$ (0.05 and 0.1).
For each combination of dataset, cascade, budget level, and $\alpha$, we calibrate C3PO using a held-out calibration set and then evaluate it on a disjoint test set. The conformal guarantee implies that the proportion of test examples for which the inference cost exceeds $C^*$ (referred to empirical violation rate) should be at most $\alpha$.
A violation occurs when this is not satisfied.
We observe only a single violation in 300 ($15 \times 2 \times 5 \times 2$) runs of C3PO, which provides strong empirical evidence in favor of the validity of the cost guarantee. In addition, in each of these runs of C3PO, the generalization bound specified in Theorem~\ref{thm:generalization} is empirically verified to be true, and in fact the train-test gap in regrets is significantly lower. 

\paragraph{In comparison to baselines, how does C3PO perform and allocate budget for questions with varying difficulties?} 
Although the main results demonstrate the beneficial cost-accuracy trade-off offered by C3PO on multiple benchmarks, they do not provide any details of the allocation of the computational budget across different questions with different difficulties.
For the MATH-500 dataset, we have access to the `true' difficulty of each question as part of the metadata, based on human annotation. 
Therefore, we conduct a detailed analysis to compare the average accuracy and incurred cost of all baselines  for each difficulty level of MATH-500. For this analysis, we evaluate each method using the hyperparameters which resulted in their maximum accuracy in Figure~\ref{fig:date}.
From  Figure~\ref{fig:difficulty_MATH 500_llama}, we observe that for all algorithms, the accuracy decreases and inference cost increases, as the questions become more difficult. 
C3PO incurs significantly lower cost for each difficulty level, while having the highest overall accuracy.

\paragraph{How sensitive is C3PO performance when using cascades comprised of models from different families?}
We investigate the effectiveness of cross family models by creating a cascade composed of LLaMA 3.2 1B-Instruct, Qwen 2.5 32B-Instruct and GPT 4o-mini.
After running experiments under the same datasets our results shown in Fig.~\ref{fig:mixed_model_family} mirror right side of Fig.~\ref{fig:boxplot} without significant differences. Our model clearly reduces costs and achieves near parity with MPM at one third of the cost. Therefore, we conclude that the method is robust to LLM family selection.

\paragraph{How how does the model perform under distribution shift?}
\begin{wrapfigure}[20]{r}{0.5\textwidth}
    \centering
    \vspace{-0.5cm}
    \includegraphics[width=\linewidth]{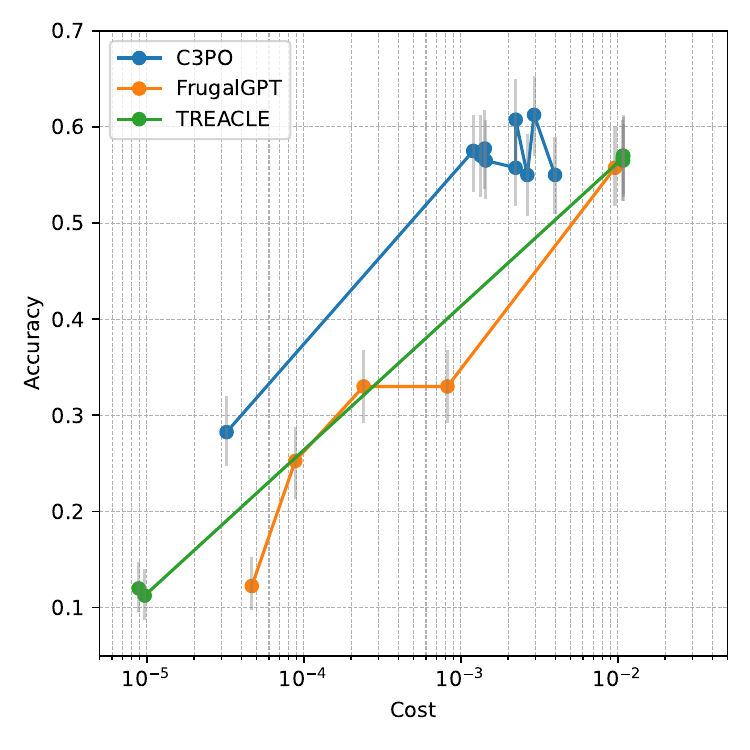}
    \vspace{-0.5cm}
    \caption{Distribution shift experiment. Training on GSM8K, testing on MATH500 shows that C3PO has the best domain adaptation performance.
    }
    \label{fig:dist_shift_1}
\end{wrapfigure}
We train C3PO, FrugalGPT, and TREACLE independently on SVAMP and GSM8K. and evaluat the learned exit policy on MATH-500 test set. This simulates a realistic setting where cascade policies are optimized using data from different supervision domains than the deployment task.    
The results demonstrate excellent distribution shift robustness for C3PO in comparison to the baselines (Fig.~\ref{fig:dist_shift_1} and App. Fig~\ref{fig:dist_shift_2}).

\section{Conclusion}

In this work, we introduced C3PO, a novel cascade framework that dynamically decides whether to accept an intermediate model’s prediction or defer to the more expensive next model in the cascade based on learned thresholds and conformal cost calibration. Our methodology combines discrete grid search over thresholds with rigorous conformal guarantees on expected cost and PAC-Bayes generalization bounds, ensuring both theoretical soundness and practical reliability. Beyond these empirical gains, C3PO’s design is highly extensible: one can incorporate richer confidence surrogates, employ more advanced search heuristics, or integrate powerful verification models without altering the core cascade logic. \textbf{Limitations:} A limitation of our approach is that the conformal cost guarantee can lead to overly conservative cascade decisions leading the model to under-utilize the cost budget sometimes.
\textbf{Broader impact:} By demonstrating that principled early‐exit strategies can deliver both high accuracy and cost efficiency, our work paves the way for more sustainable and scalable deployment of large language models in real‐world reasoning applications. 

\newpage

\begin{ack}
This research was funded by the Natural Sciences and Engineering Research Council of Canada (NSERC), [reference number 260250].
Cette recherche a été financée par le Conseil de recherches en sciences naturelles et en génie du Canada (CRSNG), [numéro de référence 260250]. 
During this research project Antonios Valkanas was supported through NSERC Postgraduate Scholarships – Doctoral (PGS-D) program, the Stavros S. Niarchos Foundation Fellowship and the Vadasz Doctoral Fellowship. Ce projet de recherche n\textsuperscript{o}~324302 est rendu possible grâce au financement du Fonds de recherche du Québec.
\end{ack}

\bibliographystyle{unsrtnat}
\bibliography{refs}

\begin{thebibliography}{54}
\providecommand{\natexlab}[1]{#1}
\providecommand{\url}[1]{\texttt{#1}}
\expandafter\ifx\csname urlstyle\endcsname\relax
  \providecommand{\doi}[1]{doi: #1}\else
  \providecommand{\doi}{doi: \begingroup \urlstyle{rm}\Url}\fi

\bibitem[Wei et~al.(2022{\natexlab{a}})Wei, Wang, Schuurmans, Bosma, ichter, Xia, Chi, Le, and Zhou]{wei2022cot}
Jason Wei, Xuezhi Wang, Dale Schuurmans, Maarten Bosma, brian ichter, Fei Xia, Ed~Chi, Quoc~V Le, and Denny Zhou.
\newblock Chain-of-thought prompting elicits reasoning in large language models.
\newblock In S.~Koyejo, S.~Mohamed, A.~Agarwal, D.~Belgrave, K.~Cho, and A.~Oh, editors, \emph{Proc. Adv. Neural Info. Proc. Sys. (NeurIPS)}, pages 24824--24837, Dec. 2022{\natexlab{a}}.

\bibitem[Viola and Jones(2001)]{viola2001rapid}
Paul Viola and Michael Jones.
\newblock Rapid object detection using a boosted cascade of simple features.
\newblock In \emph{Proc. IEEE Conf. Comp. Vision and Pattern Recognition (CVPR)}, pages 510--518, Kauai, HI, USA, Dec. 2001.

\bibitem[Zhang et~al.(2024)Zhang, Huang, Taga, Joe-Wong, Oymak, and Chen]{zhang2024treacle}
Xuechen Zhang, Zijian Huang, Ege~Onur Taga, Carlee Joe-Wong, Samet Oymak, and Jiasi Chen.
\newblock Efficient contextual {LLM} cascades through budget-constrained policy learning.
\newblock In \emph{Proc. Conf. Neural Info. Proces. Syst. (NeurIPS)}, pages 91691--91722, Vancouver, Canada, Dec. 2024.

\bibitem[Yue et~al.(2024)Yue, Zhao, Zhang, Du, and Yao]{yue2024large}
Murong Yue, Jie Zhao, Min Zhang, Liang Du, and Ziyu Yao.
\newblock Large language model cascades with mixture of thought representations for cost-efficient reasoning.
\newblock In \emph{Proc. Int. Conf. Learning Representations (ICLR)}, Vienna, Austria, May 2024.

\bibitem[Chen et~al.(2024)Chen, Zaharia, and Zou]{chen2024frugalgpt}
Lingjiao Chen, Matei Zaharia, and James Zou.
\newblock Frugalgpt: How to use large language models while reducing cost and improving performance.
\newblock \emph{Trans. Machine Learning Research (TMLR)}, 2024.

\bibitem[Zhang et~al.(2023)Zhang, Krishna, Awadallah, and Wang]{zhang2023ecoassistant}
Jieyu Zhang, Ranjay Krishna, Ahmed~H Awadallah, and Chi Wang.
\newblock Ecoassistant: Using {LLM} assistant more affordably and accurately, 2023.

\bibitem[Aggarwal et~al.(2024)Aggarwal, Madaan, Anand, Potharaju, Mishra, Zhou, Gupta, Rajagopal, Kappaganthu, Yang, Upadhyay, Faruqui, and .]{aggarwal2024automix}
Pranjal Aggarwal, Aman Madaan, Ankit Anand, Srividya~Pranavi Potharaju, Swaroop Mishra, Pei Zhou, Aditya Gupta, Dheeraj Rajagopal, Karthik Kappaganthu, Yiming Yang, Shyam Upadhyay, Manaal Faruqui, and Mausam .
\newblock Automix: Automatically mixing language models.
\newblock In \emph{Proc. Conf. Neural Info. Proces. Syst. (NeurIPS)}, pages 131000--131034, Vancouver, Canada, Dec. 2024.

\bibitem[Chen et~al.(2025{\natexlab{a}})Chen, Xun, Zhou, Qi, Zhang, Chen, Hu, Qu, Ouyang, and Hu]{chen2025we}
Jianhao Chen, Zishuo Xun, Bocheng Zhou, Han Qi, Qiaosheng Zhang, Yang Chen, Wei Hu, Yuzhong Qu, Wanli Ouyang, and Shuyue Hu.
\newblock Do we truly need so many samples? {M}ulti-{LLM} repeated sampling efficiently scale test-time compute, 2025{\natexlab{a}}.

\bibitem[Nie et~al.(2024)Nie, Ding, Hu, Jermaine, and Chaudhuri]{nie2024cascade}
Lunyiu Nie, Zhimin Ding, Erdong Hu, Christopher Jermaine, and Swarat Chaudhuri.
\newblock Online cascade learning for efficient inference over streams.
\newblock In \emph{Proc. Int. Conf. Machine Learning (ICML)}, page 38071–38090, Vienna, Austria, Jul. 2024.

\bibitem[Ong et~al.(2025)Ong, Almahairi, Wu, Chiang, Wu, Gonzalez, Kadous, and Stoica]{ong2025routellm}
Isaac Ong, Amjad Almahairi, Vincent Wu, Wei-Lin Chiang, Tianhao Wu, Joseph~E. Gonzalez, M~Waleed Kadous, and Ion Stoica.
\newblock Route{LLM}: Learning to route {LLM}s from preference data.
\newblock In \emph{Int. Conf. Learning Representations (ICLR)}, 2025.

\bibitem[Maurya et~al.(2024)Maurya, Srivatsa, and Kochmar]{maurya2024selectllm}
Kaushal~Kumar Maurya, KV~Srivatsa, and Ekaterina Kochmar.
\newblock {SelectLLM}: Query-aware efficient selection algorithm for large language models, 2024.

\bibitem[Chen et~al.(2025{\natexlab{b}})Chen, Yun, Stengel-Eskin, Chen, and Bansal]{chen2025symbolic}
Justin Chih-Yao Chen, Sukwon Yun, Elias Stengel-Eskin, Tianlong Chen, and Mohit Bansal.
\newblock {Symbolic Mixture-of-Experts}: Adaptive skill-based routing for heterogeneous reasoning, 2025{\natexlab{b}}.

\bibitem[Talmor et~al.(2019)Talmor, Herzig, Lourie, and Berant]{talmor2018commonsenseqa}
Alon Talmor, Jonathan Herzig, Nicholas Lourie, and Jonathan Berant.
\newblock {C}ommonsense{QA}: {A} question answering challenge targeting commonsense knowledge.
\newblock In \emph{Proc. Conf. N. {A}merican Chapter Assoc. Comput. Linguistics}, pages 4149--4158, Minneapolis, MN, USA, Jun. 2019.

\bibitem[Suzgun et~al.(2023)Suzgun, Scales, Sch{\"a}rli, Gehrmann, Tay, Chung, Chowdhery, Le, Chi, Zhou, and Wei]{suzgun2023bbh}
Mirac Suzgun, Nathan Scales, Nathanael Sch{\"a}rli, Sebastian Gehrmann, Yi~Tay, Hyung~Won Chung, Aakanksha Chowdhery, Quoc Le, Ed~Chi, Denny Zhou, and Jason Wei.
\newblock Challenging {BIG}-bench tasks and whether chain-of-thought can solve them.
\newblock In \emph{Proc. Assoc. Comput. Linguistics (ACL)}, pages 13003--13051, Toronto, Canada, Jul. 2023.

\bibitem[Cobbe et~al.(2021)Cobbe, Kosaraju, Bavarian, Chen, Jun, Kaiser, Plappert, Tworek, Hilton, Nakano, Hesse, and Schulman]{cobbe2021gsm8k}
Karl Cobbe, Vineet Kosaraju, Mohammad Bavarian, Mark Chen, Heewoo Jun, Lukasz Kaiser, Matthias Plappert, Jerry Tworek, Jacob Hilton, Reiichiro Nakano, Christopher Hesse, and John Schulman.
\newblock Training verifiers to solve math word problems, 2021.

\bibitem[Patel et~al.(2021)Patel, Bhattamishra, and Goyal]{patel2021}
Arkil Patel, Satwik Bhattamishra, and Navin Goyal.
\newblock Are {NLP} models really able to solve simple math word problems?
\newblock In \emph{Proc. Conf. North Amer. Chapter of the Associ. Comput. Linguistics (ACL)}, pages 2080--2094, Jun. 2021.

\bibitem[Ling et~al.(2017)Ling, Yogatama, Dyer, and Blunsom]{ling2017}
Wang Ling, Dani Yogatama, Chris Dyer, and Phil Blunsom.
\newblock Program induction by rationale generation: Learning to solve and explain algebraic word problems.
\newblock In \emph{Proc. Conf. Empirical Methods in Natural Language Process. (EMNLP)}, Copenhagen, Denmark, Sep. 2017.

\bibitem[Hendrycks et~al.(2021)Hendrycks, Burns, Kadavath, Arora, Basart, Tang, Song, and Steinhardt]{hendrycksmath2021}
Dan Hendrycks, Collin Burns, Saurav Kadavath, Akul Arora, Steven Basart, Eric Tang, Dawn Song, and Jacob Steinhardt.
\newblock Measuring mathematical problem solving with the math dataset.
\newblock In \emph{Proc. Adv. Neural Inf. Process. Syst. (NeurIPS)}, pages 1--8, Dec. 2021.

\bibitem[{Mathematical Association of America}(2024)]{aime2024}
{Mathematical Association of America}.
\newblock Aime competition problems and solutions (1983–present).
\newblock \url{https://artofproblemsolving.com/wiki/index.php/AIME_Problems_and_Solutions}, 2024.
\newblock Accessed: 2025-10-06.

\bibitem[Wei et~al.(2022{\natexlab{b}})Wei, Wang, Schuurmans, Bosma, Xia, Chi, Le, Zhou, et~al.]{wei2022chain}
Jason Wei, Xuezhi Wang, Dale Schuurmans, Maarten Bosma, Fei Xia, Ed~Chi, Quoc~V Le, Denny Zhou, et~al.
\newblock Chain-of-thought prompting elicits reasoning in large language models.
\newblock \emph{NeurIPS}, pages 24824--24837, 2022{\natexlab{b}}.

\bibitem[Aggarwal et~al.(2023)Aggarwal, Madaan, Yang, and Mausam]{aggarwal2023adapativesc}
Pranjal Aggarwal, Aman Madaan, Yiming Yang, and Mausam.
\newblock Let`s sample step by step: Adaptive-consistency for efficient reasoning and coding with {LLM}s.
\newblock In Houda Bouamor, Juan Pino, and Kalika Bali, editors, \emph{Proc. Conf. Empirical Methods in Natural Language Proces. (ACL)}, pages 12375--12396, Singapore, Dec. 2023.

\bibitem[Li et~al.(2024{\natexlab{a}})Li, Yuan, Feng, Pan, Sun, Wang, Wang, and Li]{li2024adaptivesc}
Yiwei Li, Peiwen Yuan, Shaoxiong Feng, Boyuan Pan, Bin Sun, Xinglin Wang, Heda Wang, and Kan Li.
\newblock Turning dust into gold: distilling complex reasoning capabilities from llms by leveraging negative data.
\newblock In \emph{Proc. AAAI Conf. Artificial Intell.}, 2024{\natexlab{a}}.

\bibitem[{Zhu} et~al.(2024){Zhu}, {Shen}, {Zhao}, and {Zou}]{zhu2024path_consistency}
Jiace {Zhu}, Yingtao {Shen}, Jie {Zhao}, and An~{Zou}.
\newblock {Path-Consistency: Prefix Enhancement for Efficient Inference in LLM}.
\newblock \emph{arXiv e-prints}, art. arXiv:2409.01281, August 2024.
\newblock \doi{10.48550/arXiv.2409.01281}.

\bibitem[Su et~al.(2025)Su, Sukhbaatar, Rabbat, Tian, and Zheng]{su2025dualformerControllableFast}
DiJia Su, Sainbayar Sukhbaatar, Michael Rabbat, Yuandong Tian, and Qinqing Zheng.
\newblock Dualformer: Controllable fast and slow thinking by learning with randomized reasoning traces.
\newblock In \emph{The Thirteenth International Conference on Learning Representations}, 2025.

\bibitem[DeepSeek-AI et~al.(2025)DeepSeek-AI, Guo, Yang, Zhang, Song, Zhang, Xu, Zhu, Ma, Wang, Bi, Zhang, Yu, Wu, Wu, Gou, Shao, Li, Gao, Liu, Xue, Wang, Wu, Feng, Lu, Zhao, Deng, Zhang, Ruan, Dai, Chen, Ji, Li, Lin, Dai, Luo, Hao, Chen, Li, Zhang, Bao, Xu, Wang, Ding, Xin, Gao, Qu, Li, Guo, Li, Wang, Chen, Yuan, Qiu, Li, Cai, Ni, Liang, Chen, Dong, Hu, Gao, Guan, Huang, Yu, Wang, Zhang, Zhao, Wang, Zhang, Xu, Xia, Zhang, Zhang, Tang, Li, Wang, Li, Tian, Huang, Zhang, Wang, Chen, Du, Ge, Zhang, Pan, Wang, Chen, Jin, Chen, Lu, Zhou, Chen, Ye, Wang, Yu, Zhou, Pan, Li, Zhou, Wu, Ye, Yun, Pei, Sun, Wang, Zeng, Zhao, Liu, Liang, Gao, Yu, Zhang, Xiao, An, Liu, Wang, Chen, Nie, Cheng, Liu, Xie, Liu, Yang, Li, Su, Lin, Li, Jin, Shen, Chen, Sun, Wang, Song, Zhou, Wang, Shan, Li, Wang, Wei, Zhang, Xu, Li, Zhao, Sun, Wang, Yu, Zhang, Shi, Xiong, He, Piao, Wang, Tan, Ma, Liu, Guo, Ou, Wang, Gong, Zou, He, Xiong, Luo, You, Liu, Zhou, Zhu, Xu, Huang, Li, Zheng, Zhu, Ma, Tang, Zha, Yan, Ren, Ren, Sha, Fu, Xu, Xie, Zhang,
  Hao, Ma, Yan, Wu, Gu, Zhu, Liu, Li, Xie, Song, Pan, Huang, Xu, Zhang, and Zhang]{deepseekai2025deepseekr1}
DeepSeek-AI, Daya Guo, Dejian Yang, Haowei Zhang, Junxiao Song, Ruoyu Zhang, Runxin Xu, Qihao Zhu, Shirong Ma, Peiyi Wang, Xiao Bi, Xiaokang Zhang, Xingkai Yu, Yu~Wu, Z.~F. Wu, Zhibin Gou, Zhihong Shao, Zhuoshu Li, Ziyi Gao, Aixin Liu, Bing Xue, Bingxuan Wang, Bochao Wu, Bei Feng, Chengda Lu, Chenggang Zhao, Chengqi Deng, Chenyu Zhang, Chong Ruan, Damai Dai, Deli Chen, Dongjie Ji, Erhang Li, Fangyun Lin, Fucong Dai, Fuli Luo, Guangbo Hao, Guanting Chen, Guowei Li, H.~Zhang, Han Bao, Hanwei Xu, Haocheng Wang, Honghui Ding, Huajian Xin, Huazuo Gao, Hui Qu, Hui Li, Jianzhong Guo, Jiashi Li, Jiawei Wang, Jingchang Chen, Jingyang Yuan, Junjie Qiu, Junlong Li, J.~L. Cai, Jiaqi Ni, Jian Liang, Jin Chen, Kai Dong, Kai Hu, Kaige Gao, Kang Guan, Kexin Huang, Kuai Yu, Lean Wang, Lecong Zhang, Liang Zhao, Litong Wang, Liyue Zhang, Lei Xu, Leyi Xia, Mingchuan Zhang, Minghua Zhang, Minghui Tang, Meng Li, Miaojun Wang, Mingming Li, Ning Tian, Panpan Huang, Peng Zhang, Qiancheng Wang, Qinyu Chen, Qiushi Du, Ruiqi Ge, Ruisong
  Zhang, Ruizhe Pan, Runji Wang, R.~J. Chen, R.~L. Jin, Ruyi Chen, Shanghao Lu, Shangyan Zhou, Shanhuang Chen, Shengfeng Ye, Shiyu Wang, Shuiping Yu, Shunfeng Zhou, Shuting Pan, S.~S. Li, Shuang Zhou, Shaoqing Wu, Shengfeng Ye, Tao Yun, Tian Pei, Tianyu Sun, T.~Wang, Wangding Zeng, Wanjia Zhao, Wen Liu, Wenfeng Liang, Wenjun Gao, Wenqin Yu, Wentao Zhang, W.~L. Xiao, Wei An, Xiaodong Liu, Xiaohan Wang, Xiaokang Chen, Xiaotao Nie, Xin Cheng, Xin Liu, Xin Xie, Xingchao Liu, Xinyu Yang, Xinyuan Li, Xuecheng Su, Xuheng Lin, X.~Q. Li, Xiangyue Jin, Xiaojin Shen, Xiaosha Chen, Xiaowen Sun, Xiaoxiang Wang, Xinnan Song, Xinyi Zhou, Xianzu Wang, Xinxia Shan, Y.~K. Li, Y.~Q. Wang, Y.~X. Wei, Yang Zhang, Yanhong Xu, Yao Li, Yao Zhao, Yaofeng Sun, Yaohui Wang, Yi~Yu, Yichao Zhang, Yifan Shi, Yiliang Xiong, Ying He, Yishi Piao, Yisong Wang, Yixuan Tan, Yiyang Ma, Yiyuan Liu, Yongqiang Guo, Yuan Ou, Yuduan Wang, Yue Gong, Yuheng Zou, Yujia He, Yunfan Xiong, Yuxiang Luo, Yuxiang You, Yuxuan Liu, Yuyang Zhou, Y.~X. Zhu,
  Yanhong Xu, Yanping Huang, Yaohui Li, Yi~Zheng, Yuchen Zhu, Yunxian Ma, Ying Tang, Yukun Zha, Yuting Yan, Z.~Z. Ren, Zehui Ren, Zhangli Sha, Zhe Fu, Zhean Xu, Zhenda Xie, Zhengyan Zhang, Zhewen Hao, Zhicheng Ma, Zhigang Yan, Zhiyu Wu, Zihui Gu, Zijia Zhu, Zijun Liu, Zilin Li, Ziwei Xie, Ziyang Song, Zizheng Pan, Zhen Huang, Zhipeng Xu, Zhongyu Zhang, and Zhen Zhang.
\newblock Deepseek-r1: Incentivizing reasoning capability in llms via reinforcement learning, 2025.

\bibitem[Yu et~al.(2025)Yu, Xu, Jin, Sankararaman, He, Zhou, Zeng, Helenowski, Zhu, Wang, et~al.]{yu2025think}
Zishun Yu, Tengyu Xu, Di~Jin, Karthik~Abinav Sankararaman, Yun He, Wenxuan Zhou, Zhouhao Zeng, Eryk Helenowski, Chen Zhu, Sinong Wang, et~al.
\newblock Think smarter not harder: Adaptive reasoning with inference aware optimization.
\newblock \emph{arXiv preprint arXiv:2501.17974}, 2025.

\bibitem[Wang et~al.(2024{\natexlab{a}})Wang, Wang, Athiwaratkun, Zhang, and Zou]{wang2024mixture}
Junlin Wang, Jue Wang, Ben Athiwaratkun, Ce~Zhang, and James Zou.
\newblock {Mixture-of-Agents} enhances large language model capabilities.
\newblock \emph{arXiv preprint arXiv:2406.04692}, 2024{\natexlab{a}}.

\bibitem[Jiang et~al.(2023)Jiang, Ren, and Lin]{jiangblender2023}
Dongfu Jiang, Xiang Ren, and Bill~Yuchen Lin.
\newblock Llm-blender: Ensembling large language models with pairwise comparison and generative fusion.
\newblock In \emph{Proc. Assoc. for Computat. Linguistics (ACL)}, 2023.

\bibitem[Mavromatis et~al.(2024)Mavromatis, Karypis, and Karypis]{mavromatis2024pack}
Costas Mavromatis, Petros Karypis, and George Karypis.
\newblock Pack of {LLM}s: Model fusion at test-time via perplexity optimization.
\newblock In \emph{First Conference on Language Modeling (COLM)}, 2024.

\bibitem[Nguyen et~al.(2024)Nguyen, Hoang, Decugis, Manchanda, Chawla, and Doan]{nguyen2024metallm}
Quang~H Nguyen, Duy~C Hoang, Juliette Decugis, Saurav Manchanda, Nitesh~V Chawla, and Khoa~D Doan.
\newblock Metallm: A high-performant and cost-efficient dynamic framework for wrapping llms.
\newblock \emph{arXiv preprint arXiv:2407.10834}, 2024.

\bibitem[Jin et~al.(2024)Jin, Yu, Shu, Zhao, Hua, Meng, Zhang, and Du]{jin2024impactreasoningstep}
Mingyu Jin, Qinkai Yu, Dong Shu, Haiyan Zhao, Wenyue Hua, Yanda Meng, Yongfeng Zhang, and Mengnan Du.
\newblock The impact of reasoning step length on large language models.
\newblock In \emph{Findings of the Association for Computational Linguistics: ACL 2024}. Association for Computational Linguistics, 2024.

\bibitem[Li et~al.(2025)Li, Lv, Shao, Ma, Li, Zheng, Qiu, and Guo]{li2025fastmctssimples}
Peiji Li, Kai Lv, Yunfan Shao, Yichuan Ma, Linyang Li, Xiaoqing Zheng, Xipeng Qiu, and Qipeng Guo.
\newblock Fastmcts: A simple sampling strategy for data synthesis, 2025.

\bibitem[Ding et~al.(2024)Ding, Mallick, Wang, Sim, Mukherjee, R{\"u}hle, Lakshmanan, and Awadallah]{ding2024hybrid}
Dujian Ding, Ankur Mallick, Chi Wang, Robert Sim, Subhabrata Mukherjee, Victor R{\"u}hle, Laks V.~S. Lakshmanan, and Ahmed~Hassan Awadallah.
\newblock Hybrid {LLM}: Cost-efficient and quality-aware query routing.
\newblock In \emph{Int. Conf. Learning Representations (ICLR)}, 2024.

\bibitem[Mohammadshahi et~al.(2024)Mohammadshahi, Shaikh, and Yazdani]{mohammadshahi2024routoolearningroutelarge}
Alireza Mohammadshahi, Arshad~Rafiq Shaikh, and Majid Yazdani.
\newblock Routoo: Learning to route to large language models effectively, 2024.
\newblock URL \url{https://arxiv.org/abs/2401.13979}.

\bibitem[Lu et~al.(2024)Lu, Yuan, Lin, Lin, Yuan, Zhou, and Zhou]{lu2024routing}
Keming Lu, Hongyi Yuan, Runji Lin, Junyang Lin, Zheng Yuan, Chang Zhou, and Jingren Zhou.
\newblock Routing to the expert: Efficient reward-guided ensemble of large language models.
\newblock In Kevin Duh, Helena Gomez, and Steven Bethard, editors, \emph{Proc. Conf. N.A. Chapter of the Assoc. Computat. Linguistics: Human Language Tech.}, pages 1964--1974, Mexico City, Mexico, Jun. 2024.

\bibitem[Zhao et~al.(2024)Zhao, Jin, and Mao]{zhao2024eagle}
Zesen Zhao, Shuowei Jin, and Z.~Morley Mao.
\newblock Eagle: Efficient training-free router for multi-{LLM}inference, 2024.
\newblock URL \url{https://arxiv.org/abs/2409.15518}.

\bibitem[Li et~al.(2024{\natexlab{b}})Li, Yuan, Feng, Pan, Wang, Sun, Wang, and Li]{li2024escape}
Yiwei Li, Peiwen Yuan, Shaoxiong Feng, Boyuan Pan, Xinglin Wang, Bin Sun, Heda Wang, and Kan Li.
\newblock Escape sky-high cost: Early-stopping self-consistency for multi-step reasoning.
\newblock In \emph{Int. Conf. Learning Representations (ICRL)}, 2024{\natexlab{b}}.

\bibitem[Wang et~al.(2024{\natexlab{b}})Wang, Feng, Li, Yuan, Zhang, Tan, Pan, Hu, and Li]{wang2024make}
Xinglin Wang, Shaoxiong Feng, Yiwei Li, Peiwen Yuan, Yueqi Zhang, Chuyi Tan, Boyuan Pan, Yao Hu, and Kan Li.
\newblock Make every penny count: Difficulty-adaptive self-consistency for cost-efficient reasoning.
\newblock \emph{arXiv preprint arXiv:2408.13457}, 2024{\natexlab{b}}.

\bibitem[{Pan} et~al.(2024){Pan}, {Zhang}, {Zhang}, {Liu}, {Wang}, and {Li}]{pan2024dynathink}
Jiabao {Pan}, Yan {Zhang}, Chen {Zhang}, Zuozhu {Liu}, Hongwei {Wang}, and Haizhou {Li}.
\newblock {DynaThink: Fast or Slow? A Dynamic Decision-Making Framework for Large Language Models}.
\newblock \emph{arXiv e-prints}, art. arXiv:2407.01009, July 2024.
\newblock \doi{10.48550/arXiv.2407.01009}.

\bibitem[Ding et~al.(2025)Ding, Jiang, Liu, Jing, Guo, Wang, Zhang, Wang, Liu, Du, Liu, and Tao]{ding2025dynamicparalleltree}
Yifu Ding, Wentao Jiang, Shunyu Liu, Yongcheng Jing, Jinyang Guo, Yingjie Wang, Jing Zhang, Zengmao Wang, Ziwei Liu, Bo~Du, Xianglong Liu, and Dacheng Tao.
\newblock Dynamic parallel tree search for efficient llm reasoning, 2025.

\bibitem[Kang et~al.(2024)Kang, Sun, Chen, and Zou]{kang2024c3otgeneratingshorterchain}
Yu~Kang, Xianghui Sun, Liangyu Chen, and Wei Zou.
\newblock C3ot: Generating shorter chain-of-thought without compromising effectiveness, 2024.

\bibitem[Munkhbat et~al.(2025)Munkhbat, Ho, Kim, Yang, Kim, and Yun]{munkhbat2025selftrainingelicits}
Tergel Munkhbat, Namgyu Ho, Seo~Hyun Kim, Yongjin Yang, Yujin Kim, and Se-Young Yun.
\newblock Self-training elicits concise reasoning in large language models, 2025.

\bibitem[Hao et~al.(2024)Hao, Sukhbaatar, Su, Li, Hu, Weston, and Tian]{hao2024traininglargelanguage}
Shibo Hao, Sainbayar Sukhbaatar, DiJia Su, Xian Li, Zhiting Hu, Jason Weston, and Yuandong Tian.
\newblock Training large language models to reason in a continuous latent space, 2024.
\newblock URL \url{https://arxiv.org/abs/2412.06769}.

\bibitem[Shen et~al.(2025{\natexlab{a}})Shen, Yan, Zhang, Hu, Du, and He]{shen2025codicompressingchain}
Zhenyi Shen, Hanqi Yan, Linhai Zhang, Zhanghao Hu, Yali Du, and Yulan He.
\newblock Codi: Compressing chain-of-thought into continuous space via self-distillation, 2025{\natexlab{a}}.

\bibitem[Shen et~al.(2025{\natexlab{b}})Shen, Wang, Shi, Wang, Zhao, and Gu]{shen2025efficientreasoninghidden}
Xuan Shen, Yizhou Wang, Xiangxi Shi, Yanzhi Wang, Pu~Zhao, and Jiuxiang Gu.
\newblock Efficient reasoning with hidden thinking, 2025{\natexlab{b}}.

\bibitem[Luo et~al.(2025)Luo, Shen, He, Wang, Liu, Li, Tan, Cao, and Tao]{luo2025o1}
Haotian Luo, Li~Shen, Haiying He, Yibo Wang, Shiwei Liu, Wei Li, Naiqiang Tan, Xiaochun Cao, and Dacheng Tao.
\newblock O1-pruner: Length-harmonizing fine-tuning for o1-like reasoning pruning.
\newblock \emph{arXiv preprint arXiv:2501.12570}, 2025.

\bibitem[Shen et~al.(2025{\natexlab{c}})Shen, Zhang, Huang, Shi, Zhang, Yan, Wang, Wang, and Lian]{shen2025dast}
Yi~Shen, Jian Zhang, Jieyun Huang, Shuming Shi, Wenjing Zhang, Jiangze Yan, Ning Wang, Kai Wang, and Shiguo Lian.
\newblock Dast: Difficulty-adaptive slow-thinking for large reasoning models.
\newblock \emph{arXiv preprint arXiv:2503.04472}, 2025{\natexlab{c}}.

\bibitem[Qu et~al.(2025)Qu, Yang, Setlur, Tunstall, Beeching, Salakhutdinov, and Kumar]{qu2025optimizing}
Yuxiao Qu, Matthew~YR Yang, Amrith Setlur, Lewis Tunstall, Edward~Emanuel Beeching, Ruslan Salakhutdinov, and Aviral Kumar.
\newblock Optimizing test-time compute via meta reinforcement fine-tuning.
\newblock \emph{arXiv preprint arXiv:2503.07572}, 2025.

\bibitem[Varshney and Baral(2022)]{varshney2022model}
Neeraj Varshney and Chitta Baral.
\newblock Model cascading: Towards jointly improving efficiency and accuracy of {NLP} systems.
\newblock In Yoav Goldberg, Zornitsa Kozareva, and Yue Zhang, editors, \emph{Proc. Conf. Empirical Methods in Natural Language Proces. (EMNLP)}, pages 11007--11021, Abu Dhabi, United Arab Emirates, Dec. 2022.

\bibitem[Ram{\'i}rez et~al.(2024)Ram{\'i}rez, Lindemann, Birch, and Titov]{ramirez2023cache}
Guillem Ram{\'i}rez, Matthias Lindemann, Alexandra Birch, and Ivan Titov.
\newblock Cache {\&} distil: Optimising {API} calls to large language models.
\newblock In Lun-Wei Ku, Andre Martins, and Vivek Srikumar, editors, \emph{Findings of Assoc. Comput. Linguistics (ACL)}, pages 11838--11853, Bangkok, Thailand, Aug. 2024.

\bibitem[Dekoninck et~al.(2024)Dekoninck, Baader, and Vechev]{dekoninck2024unified}
Jasper Dekoninck, Maximilian Baader, and Martin Vechev.
\newblock A unified approach to routing and cascading for llms.
\newblock \emph{arXiv preprint arXiv:2410.10347}, 2024.

\bibitem[Hu et~al.(2024)Hu, Wang, Zhang, Zhou, Chen, Hu, Xiao, and Tan]{hu2024dynamic}
Jinwu Hu, Yufeng Wang, Shuhai Zhang, Kai Zhou, Guohao Chen, Yu~Hu, Bin Xiao, and Mingkui Tan.
\newblock Dynamic ensemble reasoning for {LLM} experts.
\newblock \emph{arXiv preprint arXiv:2412.07448}, 2024.

\bibitem[Gupta et~al.(2024)Gupta, Narasimhan, Jitkrittum, Rawat, Menon, and Kumar]{gupta2024cascade}
Neha Gupta, Harikrishna Narasimhan, Wittawat Jitkrittum, Ankit~Singh Rawat, Aditya~Krishna Menon, and Sanjiv Kumar.
\newblock Language model cascades: Token-level uncertainty and beyond.
\newblock In \emph{Int. Conf. Learning Representations {(ICLR)}}, Vienna, Austria, May 2024.

\bibitem[Wang et~al.(2023)Wang, Wei, Schuurmans, Le, Chi, Narang, Chowdhery, and Zhou]{wang2023selfconsistency}
Xuezhi Wang, Jason Wei, Dale Schuurmans, Quoc~V Le, Ed~H. Chi, Sharan Narang, Aakanksha Chowdhery, and Denny Zhou.
\newblock Self-consistency improves chain of thought reasoning in language models.
\newblock In \emph{Int. Conf. Learning Representations (ICRL)}, 2023.

\end{thebibliography}
\newpage
\appendix

\addcontentsline{toc}{section}{Appendix} %
\part{\centerline{C3PO:
Optimized LLM Cascades with Probabilistic}
\centerline{Cost Constraints for Reasoning}
\bigskip{}
\bigskip{}
\bigskip{}
\centerline{Appendix}} %
\parttoc %
\newpage

\section{Broader Impact, Limitations and Code}

\setcounter{theorem}{0}

The development of \textbf{C3PO} (Cost Controlled Cascaded Prediction Optimization) has the potential to significantly improve the accessibility, sustainability, and scalability of large language model (LLM) deployments, especially in real-world reasoning tasks. As LLMs continue to expand in size and capability, their computational and monetary costs pose a critical bottleneck that limits their application in low-resource settings, as well as their environmental sustainability. C3PO addresses this issue by offering a label-free, conformally-controlled framework for optimizing cascaded inference, enabling accurate and efficient LLM use while providing formal guarantees on inference cost.

\paragraph{Social and Environmental Impact.} By substantially reducing reliance on powerful LLMs for every query, C3PO can lower the environmental footprint associated with LLM inference. For institutions operating under strict compute or budget constraints such as nonprofits, schools, or healthcare systems, C3PO offers a practical path to deploying high-quality language systems without incurring prohibitive costs. Its ability to generalize from unlabeled data further reduces barriers to adoption in under-resourced domains where labeled data is scarce or unavailable.

\paragraph{Equity and Access.} C3PO’s cost-efficiency and domain adaptability can democratize access to high-quality language technologies.  For example, resource-constrained platforms (e.g., small online businesses) could adopt C3PO to offer reliable results at minimal cost, expanding reach without compromising performance.

\paragraph{Limitations, Risks and Misuse.} As with any method that enables wider LLM deployment, C3PO may inadvertently facilitate the use of language models in settings lacking oversight or appropriate safeguards. Cost-reduction techniques might be applied indiscriminately to critical decision-making domains (e.g., legal or healthcare) without verifying the alignment of model behavior with task requirements. To mitigate these risks, it is important to emphasize that C3PO offers probabilistic control over cost but does not itself guarantee correctness or fairness of model outputs. Proper validation, human-in-the-loop design, and task-specific evaluation must accompany its use in high-stakes applications.

\paragraph{Explainability and Transparency.} C3PO’s reliance on interpretable thresholds and its theoretical guarantees on both generalization and cost provide a degree of transparency that many black-box routing or reinforcement-based systems lack. These properties support responsible use and make C3PO amenable to audit and monitoring, especially in safety-critical deployments. Moreover, its minimal data requirements lower the need for large-scale data collection, which may help reduce privacy concerns associated with supervised training pipelines.

\paragraph{Conclusion.} C3PO provides a scalable and principled framework for efficient LLM inference that is compatible with privacy-sensitive, low-resource, and budget-constrained environments. Its broader societal benefit lies in making LLM-powered reasoning more affordable and sustainable, while its limitations call for responsible deployment practices that pair efficiency with rigorous domain-specific evaluation. \textbf{We make our code available at \url{https://github.com/AntonValk/C3PO-LLM}.}

\newpage

\section{Extended Literature Review}

\subsection{Overview of Efficient LLM Test Time Compute Allocation Literature}
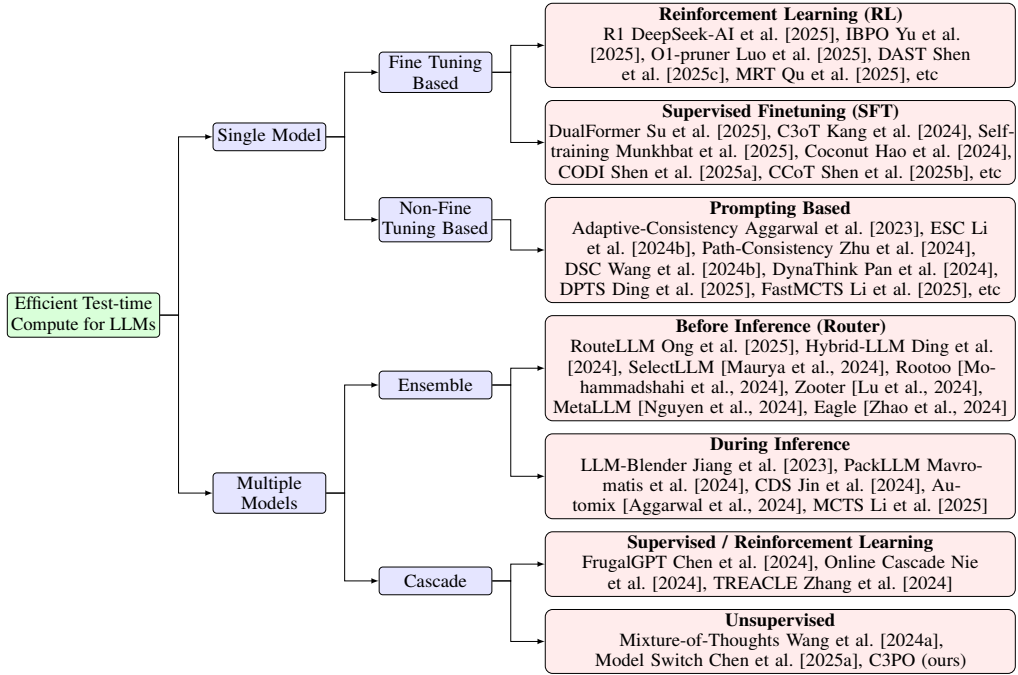
\begin{figure}
    \centering

\tikzset{
    basic/.style  = {rectangle,
    draw,
    rounded corners,
    text opacity=1,
    minimum height=1.5em,
    minimum width=5em,
    inner sep=2pt,
    align=center,
    fill opacity=.5,},
    root/.style   = {basic, rounded corners=2pt, thin, align=center, fill=green!30,child anchor=north, parent anchor=south, anchor=center},
    onode/.style  = {basic, thin, rounded corners=2pt, align=center, fill=green!60, text width=3cm},
    tnode/.style  = {basic, thin, align=center, fill=pink!60, text width=15em},
    xnode/.style  = {basic, thin, rounded corners=2pt, align=center, fill=blue!20, text width=2cm},
    edge from parent/.style={draw=black, edge from parent fork right}
}
\scalebox{0.7}{
\begin{forest}
  forked edges,
    for tree={
      grow=east,
       growth parent anchor=west,
       parent anchor=east,
       child anchor=west,
       minimum width=6em,
    edge path={
      \noexpand\path[\forestoption{edge},->,>=latex]
        (!u.parent anchor) -- +(10pt,0pt) |- (.child anchor)
        \forestoption{edge label};
    }
    },
  [Efficient Test-time\\Compute for LLMs, root, l sep=10mm
    [Multiple Models, xnode, l sep=10mm
      [Cascade, xnode, l sep=10mm
        [\textbf{Unsupervised} \\ 
        Mixture-of-Thoughts~\citet{wang2024mixture}{,}
        Model Switch~\citet{chen2025we}{,} C3PO (ours) , tnode, text width=25em]
        [\textbf{Supervised / Reinforcement Learning} \\ FrugalGPT~\citet{chen2024frugalgpt}{,} Online Cascade~\citet{nie2024cascade}{,} TREACLE~\citet{zhang2024treacle}, tnode, text width=25em]
      ]
      [Ensemble, xnode, l sep=10mm
        [\textbf{During Inference} \\ 
        LLM-Blender~\citet{jiangblender2023}\text{,} PackLLM~\citet{mavromatis2024pack}\text{,} CDS~\citet{jin2024impactreasoningstep}\text{,} Automix~\citep{aggarwal2024automix}\text{,} MCTS~\citet{li2025fastmctssimples}, tnode, text width=25em]
        [\textbf{Before Inference (Router)} \\ RouteLLM~\citet{ong2025routellm}\text{,}
        Hybrid-LLM~\citet{ding2024hybrid}\text{,} SelectLLM~\citep{maurya2024selectllm}\text{,} Rootoo~\citep{mohammadshahi2024routoolearningroutelarge}\text{,} Zooter~\citep{lu2024routing}\text{,} MetaLLM~\citep{nguyen2024metallm}\text{,} Eagle~\citep{zhao2024eagle},
        tnode, text width=25em]
      ]
    ]
    [Single Model, xnode, l sep=10mm
      [Non-Fine Tuning Based, xnode, l sep=10mm
        [\textbf{Prompting Based} \\ Adaptive-Consistency~\citet{aggarwal2023adapativesc}{,} ESC~\citet{li2024escape}{,} Path-Consistency~\citet{zhu2024path_consistency}{,} DSC~\citet{wang2024make}{,} DynaThink~\citet{pan2024dynathink}{,} DPTS~\citet{ding2025dynamicparalleltree}{,} FastMCTS~\citet{li2025fastmctssimples}{,} etc, tnode, text width=25em]
      ]
      [Fine Tuning Based, xnode, l sep=10mm
        [\textbf{Supervised Finetuning (SFT)} \\ DualFormer~\citet{su2025dualformerControllableFast}\text{,} C3oT~\citet{kang2024c3otgeneratingshorterchain}\text{,} Self-training~\citet{munkhbat2025selftrainingelicits}\text{,} Coconut~\citet{hao2024traininglargelanguage}{,} CODI~\citet{shen2025codicompressingchain}{,} CCoT~\citet{shen2025efficientreasoninghidden}{,} etc, tnode, text width=25em]
        [\textbf{Reinforcement Learning (RL)} \\ R1~\citet{deepseekai2025deepseekr1}{,} IBPO~\citet{yu2025think}{,} O1-pruner~\citet{luo2025o1}{,} DAST~\citet{shen2025dast}{,} MRT~\citet{qu2025optimizing}{,}  etc, tnode, text width=25em]
      ]
    ]
  ]
\end{forest}}

    \caption{Taxonomy of recent research on efficient test-time compute for LLMs.}
    \label{fig:lit_surv}
\end{figure}

Fig.~\ref{fig:lit_surv} presents a comprehensive taxonomy of current approaches to reducing test‐time computational cost for large language models (LLMs). At the highest level, the diagram bifurcates into two overarching strategies: those that employ multiple models in a coordinated fashion, and those that operate on a single model but optimize its internal inference process. 

On the top branch of Fig.~\ref{fig:lit_surv}, single‐model approaches achieve efficiency by modifying the inference process of one LLM. Non‐fine‐tuning methods adapt the prompt or sampling strategy such as Chain-of-Thought~\citep{wei2022chain} at test time. Some examples include adaptive self‐consistency~\citep{aggarwal2023adapativesc}, early‐stopping tests for chain‐of‐thought samples~\citep{li2024adaptivesc}, and dynamic path‐consistency pruning~\citep{zhu2024path_consistency} to reduce redundant reasoning steps without additional training. Fine‐tuning based approaches, by contrast, incur an offline training cost to produce a model that internally decides when to exit or which reasoning branches to pursue. Supervised fine‐tuning variants embed early‐exit logic into the model weights~\citep{su2025dualformerControllableFast}, while reinforcement learning variants use reward signals to guide policy learning for token or chain pruning~\citep{deepseekai2025deepseekr1, yu2025think}. \citet{deepseekai2025deepseekr1}) has shown that one can prune CoT generation \emph{inside} a single LLM via an internal reward model. We note that our framework naturally subsumes such approaches as models inside the cascade. As such, any model equipped with an internal CoT stopping rule can participate in the cascade, using our self‐supervised thresholding and conformal‐cost machinery to decide whether to exit. 

On the lower branch of Fig.~\ref{fig:lit_surv}, multiple‐model techniques are organized into cascaded pipelines and ensemble methods. Cascaded pipelines invoke a sequence of LLMs ordered by increasing computational expense: an entry‐level model attempts the task first, and only if its output fails a decision criterion does the system escalate to a more powerful model. Within this category, unsupervised cascades require no labeled data, relying instead on internal agreement or consistency signals among model outputs. Notable instances include methods that sample multiple reasoning chains and halt when they converge (e.g., Mixture‐of‐Thoughts~\citep{wang2024mixture}), or that compare a weaker model’s output against a stronger model for self‐supervised thresholding (as in C3PO). In contrast, supervised and reinforcement‐learning cascades use labeled examples or reward signals to train a meta‐model or policy that decides when to stop or escalate—for example, FrugalGPT~\citep{chen2024frugalgpt} uses a fine‐tuned classifier to predict correctness, while TREACLE~\citep{zhang2024treacle} applies deep Q‐learning to balance cost against accuracy. 

Also under the multiple‐model umbrella, ensemble methods combine responses from several LLMs either during inference (by aggregating or voting on outputs from parallel queries) or before inference (by routing each prompt to a single selected model based on input features). During‐inference ensembles such as LLM‐Blender~\citep{jiangblender2023} and PackLLM~\citep{mavromatis2024pack} query all candidate models and merge their outputs in real time, often improving robustness at the expense of higher instantaneous compute. Pre‐inference routing methods like RouteLLM~\citep{ong2025routellm} and MetaLLM~\citep{nguyen2024metallm} train lightweight selectors to choose the most suitable model for each query, thereby avoiding sequential calls but requiring supervised training on a massive training set or online adaptation to maintain accuracy across domains. We note the special case of Automix~\citep{aggarwal2024automix}. While Automix is a sequential method in the sense that it calls each candidate LLM one at a time, it also uses an external verifier LLM. Thus, due to this presence of at least 2 LLM models per output we classify it as an ensemble method. 

By tracing paths from the root through these subcategories, we observe that C3PO occupies the multiple‐model, cascade, unsupervised quadrant. Unlike supervised cascades, it requires no labeled data; unlike heuristic confidence‐threshold cascades, it leverages self‐supervised agreement metrics and conformal prediction to enforce probabilistic cost bounds; and unlike pure routing or ensemble methods, it dynamically defers per input based on realized outputs. This unified view highlights both the diversity of existing techniques and the unique position of C3PO in providing label-free, theoretically grounded cost control for cascaded LLM inference.

\subsection{Efficient Multi‐Model Inference for LLMs}

Inference cost is a critical barrier to deploying large language models (LLMs) at scale.  A prominent line of research reduces average compute by \emph{cascading} or \emph{routing} queries through multiple LLMs of increasing size and cost, so that cheaper models handle \emph{easy} inputs and only \emph{hard} cases escalate to more expensive ones.  Broadly, existing methods differ in (1) their supervision requirements; (2) decision criteria for stopping or escalating; and (3) whether they provide formal guarantees on cost or accuracy.  Below, we review representative approaches, grouping them by supervision regime and decision strategy, and highlight how our method, C3PO, addresses outstanding gaps.

\paragraph{Unsupervised and Heuristic Cascades}

Early cascade strategies adapted classical classifier‐cascade ideas to LLMs without requiring any labeled data.  \citet{varshney2022model} leverage a model’s softmax‐based confidence: if a small LLM’s maximum token probability exceeds a heuristic threshold, decoding halts; otherwise, the next model is queried.  
While simple and label‐free, this approach relies on manually tuned thresholds and offers no probabilistic cost control. 
\citet{yue2024large} propose \emph{Mixture‐of‐Thoughts (MoT)}, which samples multiple reasoning chains from a weak model and halts when self‐consistency among samples is high.  MoT improves decision robustness via ensemble‐style voting, but incurs extra sampling cost and still lacks formal guarantees on budget overruns.

Neural caching techniques \citep{ramirez2023cache} introduce online distillation of a strong model into a weak one: when the weak model’s confidence is low, the strong model answers and its output augments a cache that refines the weak model over time.  This iterative distillation reduces repeated expensive calls, but the cache requires streaming data and provides no worst‐case cost bounds.  Another line, exemplified by \citet{dekoninck2024unified}, learns a lightweight \emph{cascade and router hybrid model} using unlabeled internal features (e.g., token entropy) to predict whether to exit or escalate.  Under mild distributional assumptions, this router can enforce a user‐specified cost bound with high probability, yet it often needs some held‐out calibration and does not directly optimize for regret relative to the oracle.

\paragraph{Supervised Meta‐Model and Reinforcement Learning Approaches}

To improve predictiveness of the exit decision, several works resort to supervised learning on labeled meta‐data.  \citet{chen2024frugalgpt} train a BERT‐style meta‐model to estimate each LLM’s correctness probability and stop when confidence exceeds a learned threshold.  FrugalGPT delivers strong empirical speedups, but depends on large annotated datasets and lacks formal cost‐violation guarantees.  Reinforcement‐based methods such as \emph{AutoMix} \citep{aggarwal2024automix} and \emph{DER} \citep{hu2024dynamic} frame cascade control as a Markov decision process, learning a policy to decide both whether to stop and which successor model to call.  These methods achieve fine‐grained cost–accuracy trade‐offs but incur high meta‐training overhead and offer only empirical, rather than provable, guarantees. Additionally, these approaches can be called hybrid router-cascade approaches as they either use more than one LLM during inference (AutoMix) or they are free to traverse the cascade in any order they choose (DER) as opposed to fixed cheap-to-expensive route allowed in classic cascade methods. Thus by being allowed to jump to an arbitrary model DER resembles more of a sequential router rather than a cascade.

Quantile‐feature cascades \citep{gupta2024cascade} represent a middle ground. \citet{gupta2024cascade} extract quantile statistics from token‐level confidences to select whether to exit early or to escalate to a larger model. An issue with this formulation is that it requires access to model internals (embeddings), and it is hence not applicable to closed source models.

\paragraph{Discussion and Positioning of C3PO}

Table~\ref{tab:comparison} summarizes key attributes of these methods.  Most unsupervised cascades use confidence or consistency heuristics but lack formal cost guarantees; supervised and RL methods offer strong empirical performance at the expense of labeled data and black‐box policies; routing methods trade multiple calls for a single, pre‐selection step but cannot adapt dynamically post‐response.  None combine (a) \emph{self‐supervised} training from unlabeled agreement signals, (b) \emph{conformal} calibration to bound the probability of exceeding a user budget, and (c) \emph{PAC‐Bayesian} generalization guarantees on accuracy regret.

In contrast, C3PO is entirely label‐free, using only \emph{self‐supervision} from off‐the‐shelf model outputs; it employs \emph{conformal prediction} to guarantee that inference cost exceeds a user‐specified threshold with bounded probability; and it derives \emph{PAC‐Bayesian} bounds on accuracy regret.  This unique combination makes C3PO the first method to blend unsupervised cascade learning with rigorous cost and generalization guarantees, addressing key limitations of prior work.

\begin{table}[h]
  \centering
  \small
  \begin{tabular}{lccc}
    \toprule
    Method & Supervision & Signal & Cost Bounds \\
    \midrule
    Self‐Consistency \citep{wang2023selfconsistency}  
      & None  & Sample consensus        & None            \\
    \rowcolor{gray!15}
    Mixture‐of‐Thoughts (MoT) \citep{yue2024large}  
      & None  & Self‐consistency voting & None            \\
    ESC / Adaptive SC \citep{li2024escape,li2024adaptivesc}  
      & None  & Stopping tests          & None            \\
    Neural Caching \citep{ramirez2023cache}  
      & None  & Conf. + distill         & None            \\
    Cascade Routing \citep{dekoninck2024unified}  
      & Weak  & Entropy/router          & Prob.\ (w/o labels) \\
    \rowcolor{gray!15}
    FrugalGPT \citep{chen2024frugalgpt}  
      & Full  & Meta‐model score        & Empirical       \\
    DER \citep{hu2024dynamic}  
      & Full  & RL policy               & Empirical       \\
    AutoMix \citep{aggarwal2024automix}  
      & Full  & RL policy               & Empirical       \\
    \rowcolor{gray!15}
    TREACLE \citep{zhang2024treacle}  
      & Full  & RL policy               & Empirical       \\
    Online Cascade \citep{nie2024cascade}  
      & Full  & RL policy               & Empirical       \\
    Quantile Cascades \citep{gupta2024cascade}  
      & Full  & Quantile features       & Heuristic       \\
    \rowcolor{gray!15}
    ModelSwitch \citep{chen2025we}  
      & None  & Vote disagreement       & None            \\
    RouteLLM \citep{ong2025routellm}  
      & Full  & Input features          & None            \\
    \rowcolor{gray!15}
    \textbf{C3PO}  
      & None  & Agreement signal        & Conformal       \\
    \bottomrule
  \end{tabular}
  \caption{Comparison of multi‐LLM inference methods: supervision level, decision signal, and cost guarantees. Highlighted rows denote our primary baselines and proposed method. \textbf{Our baselines and proposed method are the only methods that (i) are pure cascade methods; (ii) do not fine tune the LLMs; and (iii) can be applied to both open and closed source LLMs.}}
  \label{tab:comparison}
\end{table}

\newpage

\section{Conformal Cost Guarantee}\label{app:conformal}

\begin{theorem}[Conformal Cost Guarantee]
Let $\boldsymbol\tau$ be the thresholds and $\mathcal{D}_{\mathrm{Cal}}$ denote a calibration set containing $N_{\mathrm{Cal}}$ questions and the cascade answers, obtained using 
 thresholds $\boldsymbol\tau$. Sort the costs of the cascade on the calibration dataset and define the rank of the budget $C^*$ as $k\eqdef \min\{p: C_{(p-1)} \leqslant  C^* \leqslant C_{(p)}\}$. If $k \geqslant \lceil (N_{Cal}+1)(1 - \alpha) \rceil$ is satisfied, then the inference cost $C_{\mathrm{test}}$ for a new query can be bounded under exchangeability of calibration and test data with 
$\Pr(C_{\mathrm{test}} > C^*) \leqslant \alpha$.
\end{theorem}

\begin{proof}
Let $R$ denote the rank of $C_{\mathrm{test}}$ among $\{C_i\}_{i=1}^{N_{\mathrm{Cal}}} \cup \{C_{\mathrm{test}}\}$. By exchangeability, we have,
\begin{equation}
\Pr(R = r) = \frac{1}{N_{\mathrm{Cal}}+1} \quad \text{for all } r \in \{1, \ldots, N_{\mathrm{Cal}}+1\}.
\end{equation}
Therefore,
\begin{equation}
\Pr(C_{\mathrm{test}} > C^*) = 
\Pr(C_{\mathrm{test}} > C_{(k)}) = \Pr(R > k) = \frac{N_{\mathrm{Cal}}+1 - k}{N_{\mathrm{Cal}}+1} \leqslant \alpha,
\end{equation}
since $k \geqslant (N_{\mathrm{Cal}}+1)(1 - \alpha)$ by construction. 
\end{proof}

\subsection{Extension to stochastic cost setting}\label{sec:stochastic_cost}

We show that the conformal cost-control guarantee in Theorem 1 can be extended to the stochastic cost setting, where model cost varies per query (e.g., due to variable output lengths in chain-of-thought reasoning). While some statistical efficiency is lost, the form of the guarantee remains intact.

Recall the setting with fixed costs. Each model $M_j$ has a fixed, known cost $c_j$. The total cascade cost for a query $x$ with exit policy $\tau$ is
\begin{align}
C(x; \tau) = \sum_{k=1}^{z(x, \tau)} c_k,
\end{align}
which is deterministic. Sorting $C(x_i; \tau)$ over the calibration set and selecting the $(1-\alpha)$-quantile $C^*$ gives the guarantee
\begin{align}
\Pr_{x \sim \text{test}}[C(x; \tau) > C^*] \le \alpha,
\end{align}
under the standard assumption that calibration and test samples are exchangeable.

Now, consider stochastic costs.
Now assume each model $M_j$ on input $x$ has a random cost $C_j(x)$ (e.g., proportional to output length), so that the total cost becomes
\begin{align}
C(x; \tau) = \sum_{k=1}^{z(x, \tau)} C_k(x),
\end{align}
which is itself a random variable. If the joint distribution of $\{(x_i, C_1(x_i), \dots, C_m(x_i))\}_{i=1}^{N_\text{cal}}$ and a test sample $(x_\text{test}, C_1(x_\text{test}), \dots, C_m(x_\text{test}))$ is exchangeable, then applying conformal prediction to the realized total costs still yields
\begin{align}
\Pr[C(x_\text{test}; \tau) > C^*] \le \alpha,
\end{align}
where $C^*$ is the empirical $(1 - \alpha)$-quantile of the realized $C(x_i; \tau)$ values on the calibration set. Thus, the guarantee is unchanged in form.

We note the removing the fixed cost assumption can introduce practical complexities. If $C(x)$ has high variance or heavy tails (e.g., due to long CoT outputs for a tiny fraction of questions), more calibration samples may be required for a reliable quantile estimate. Additionally, model costs can be bounded deterministically by $\bar{c}_j$, e.g., by truncating CoT length during generation, one can recover a worst-case cost upper bound $\sum_j \bar{c}_j$, at some loss in predictive power. Note that most APIs allow users to set a limit on maximum generation length (in terms of tokens), so, it is practically easy to control the tail.

Thus, the conformal cost-control guarantee in Theorem 1 extends naturally to the stochastic cost setting. The assumption of fixed per-model costs can be relaxed to per-query random costs, as long as exchangeability is preserved. The primary cost is a potential loss in efficiency, which can be addressed through additional calibration data or bounded model outputs. We will add this discussion in the camera ready version of the paper.

\newpage

\section{Generalization Guarantees}\label{app:generalization}

\begin{theorem}[Generalization Bounds]
Let $\mathcal{H}=\prod_{j=1}^{m-1}\mathcal{T}_j$ denote the hypothesis class of all threshold combinations with $|\mathcal{H}| = K^{(m-1)}$, where $K$ denotes the grid-size and $m$ is the number of LLMs in the cascade.
Let $\mathcal{H}_c \subset \mathcal{H}$ be the subset of thresholds for which the cost-constraint in Eq.~\eqref{eq:cost_obj} is satisfied and $\tau^*$ is the learned threshold vector from C3PO using $\mathcal{D_{\mathrm{SS}}}$ and $\mathcal{D_{\mathrm{Cal}}}$.  Then, for any $\delta \in (0, 1)$, with probability at least $(1-\delta)$, we have 
\begin{align}
L(\boldsymbol{\tau}^*) &\leqslant \widehat{L}(\boldsymbol{\tau}^*) + \sqrt{\dfrac{(m-1) \log K - \log \delta}{2N_{\mathrm{SS}}}}\,,\label{eq:train-test}\\
\text{and, }L(\boldsymbol{\tau}^*) &\leqslant \min_{\boldsymbol{\tau} \in \mathcal{H}_c} L(\boldsymbol{\tau}) + 2\sqrt{\dfrac{(m-1) \log K - \log \delta}{2N_{\mathrm{SS}}}}.\label{eq:test-test}
\end{align}
\end{theorem}

\begin{proof}
For any threshold configuration $\boldsymbol{\tau} \in \mathcal{H}$, we know that $0 \leq L(\boldsymbol{\tau}), \widehat{L}(\boldsymbol{\tau}) \leq 1$. The empirical regret, $\widehat{L}(\boldsymbol{\tau})$ is an unbiased estimator of the `true' regret $L(\boldsymbol{\tau}) = \mathbb{E}\big[\widehat{L}(\boldsymbol{\tau})\big]$ on i.i.d. samples from $\mathcal{D}_{\mathrm{SS}}$.
Applying Hoeffding's inequality, we have, for any $\boldsymbol{\tau}$ and $\epsilon>0$:
\begin{align}
\Pr(L(\boldsymbol{\tau}) > \widehat{L}(\boldsymbol{\tau}) + \epsilon) &\leqslant e^{-2N_{\mathrm{SS}}\epsilon^2}\,,\label{eq:hoeffding1}\\
\text{and, }
\Pr(\widehat{L}(\boldsymbol{\tau}) > L(\boldsymbol{\tau}) + \epsilon) &\leqslant e^{-2N_{\mathrm{SS}}\epsilon^2}\,.\label{eq:hoeffding2}
\end{align}
Applying an union bound over all $\boldsymbol{\tau} \in \mathcal{H}$, we obtain:
\begin{align}
\Pr(\exists \boldsymbol{\tau} \in \mathcal{H} :L(\boldsymbol{\tau}) > \widehat{L}(\boldsymbol{\tau}) + \epsilon) &\leqslant \sum_{\boldsymbol{\tau} \in \mathcal{H}} \Pr(L(\boldsymbol{\tau}) > \widehat{L}(\boldsymbol{\tau}) + \epsilon) \leqslant |\mathcal{H}| e^{-2N_{\mathrm{SS}}\epsilon^2}\,,
\\
\text{and, }
\Pr(\exists \boldsymbol{\tau} \in \mathcal{H} : \widehat{L}(\boldsymbol{\tau}) > L(\boldsymbol{\tau}) + \epsilon) &\leqslant\sum_{\boldsymbol{\tau} \in \mathcal{H}}\Pr(\widehat{L}(\boldsymbol{\tau}) > L(\boldsymbol{\tau}) + \epsilon)    \leqslant |\mathcal{H}| e^{-2N_{\mathrm{SS}}\epsilon^2}\,.
\end{align}

Letting the right-hand side be equal to $\delta$, we solve for $\epsilon$:
\begin{equation}
\delta = |\mathcal{H}| e^{-2N_{\mathrm{SS}}\epsilon^2} \quad \Rightarrow \quad \epsilon = \sqrt{\frac{\log(|\mathcal{H}|/\delta)}{2N_{\mathrm{SS}}}} = \sqrt{\dfrac{(m-1) \log K - \log \delta}{2N_{\mathrm{SS}}}}\,.
\end{equation}
Therefore, with probability at least $(1 - \delta)$, we have the uniform bounds
\begin{align}
\forall \boldsymbol{\tau} \in \mathcal{H}, L(\boldsymbol{\tau}) &\leqslant \widehat{L}(\boldsymbol{\tau}) + \sqrt{\dfrac{(m-1) \log K - \log \delta}{2N_{\mathrm{SS}}}}\,,\label{eq:union_bound1}\\
\text{and}, \widehat{L}(\boldsymbol{\tau}) &\leqslant L(\boldsymbol{\tau}) + \sqrt{\dfrac{(m-1) \log K - \log \delta}{2N_{\mathrm{SS}}}}\,,\label{eq:union_bound2}
\end{align}

Since the bound in eq.~\eqref{eq:union_bound1} holds for all $\boldsymbol{\tau} \in \mathcal{H}$, it holds for the learned threshold vector from C3PO, $\boldsymbol{\tau}^*$. This proves the claim stated in eq.~\eqref{eq:train-test}. Essentially, it shows that the `true' test regret at C3PO's learned threshold cannot exceed the empirical regret on $\mathcal{D}_{\mathrm{SS}}$ by more than $\epsilon$ with a high probability, greater than $(1-\delta)$. 

Now, let us denote by $\widetilde{\boldsymbol{\tau}}$ the minimizer of true risk $L(\boldsymbol{\tau})$ over $\mathcal{H}_c \subset \mathcal{H}$, where $\mathcal{H}_c$ is the subset of thresholds for which the cost-constraint in Eq.~\eqref{eq:cost_obj} is satisfied.

Now, w.p. at least $(1-\delta)$, we have,
\begin{align}
L(\boldsymbol{\tau}^*) &\leqslant \widehat{L}(\boldsymbol{\tau}^*) + \sqrt{\dfrac{(m-1) \log K - \log \delta}{2N_{\mathrm{SS}}}} \,, \text{ from eq.~\eqref{eq:train-test}, since, } \boldsymbol{\tau}^* \in \mathcal{H}_c\\
&\leqslant \widehat{L}(\widetilde{\boldsymbol{\tau}}) + \sqrt{\dfrac{(m-1) \log K - \log \delta}{2N_{\mathrm{SS}}}} \,, \text{ since, } \boldsymbol{\tau}^* \text{ is the minimizer of } \widehat{L}(\boldsymbol{\tau}) \text{ over } \mathcal{H}_c\\
&\leqslant L(\widetilde{\boldsymbol{\tau}}) + \sqrt{\dfrac{(m-1) \log K - \log \delta}{2N_{\mathrm{SS}}}} + \sqrt{\dfrac{(m-1) \log K - \log \delta}{2N_{\mathrm{SS}}}} \,, \text{ using the bound in eq.~\eqref{eq:union_bound2} }\\
&= \min_{\boldsymbol{\tau} \in \mathcal{H}_c} L(\boldsymbol{\tau}) + 2\sqrt{\dfrac{(m-1) \log K - \log \delta}{2N_{\mathrm{SS}}}}\,, \text{ since, } \widetilde{\boldsymbol{\tau}} \text{ is the minimizer of } L(\boldsymbol{\tau}) \text{ over } \mathcal{H}_c,
\end{align}
which proves the claim in eq~\eqref{eq:test-test}. This shows that the `true' regret at C3PO's learned threshold cannot exceed the minimum attainable test regret by more that $2\epsilon$ with a high probability, greater than $(1-\delta)$. 

\end{proof}

\newpage

\section{Upper Bound on the Minimal Detectable Change in Empirical Regret}\label{app:grid}

\begin{theorem}[Upper Bound on the MDC in Empirical Regret]
\label{thm:grid}
Let $\widehat L(\boldsymbol{\tau})$ be the empirical 0--1 regret over $N_{\mathrm{SS}}$ samples at threshold $\boldsymbol{\tau}$. Denote its standard error by $\widehat \sigma_{\widehat L(\boldsymbol{\tau})} = \sqrt{\widehat L(\boldsymbol{\tau})\big(1-\widehat L(\boldsymbol{\tau})\big)/N_{\mathrm{SS}}}$. Let $z_{(1-\alpha/2)}$ denote the $(1-\alpha/2)$ quantile of the standard normal distribution, $\mathcal{N}(0,1)$. Absolute differences between independently estimated empirical regrets at any two thresholds $\boldsymbol{\tau}$ and $\boldsymbol{\tau}'$ (i.e., $|\widehat L(\boldsymbol{\tau})- \widehat L(\boldsymbol{\tau}')|$), which is  smaller than $\Delta_{\mathrm{min}} = z_{(1-\alpha/2)} \sqrt{\widehat \sigma_{\widehat L(\boldsymbol{\tau})}^2 + \widehat \sigma_{\widehat L(\boldsymbol{\tau}')}^2}$, are statistically indistinguishable at the confidence level $(1-\alpha)$ for any $\alpha \in (0, 1)$. In addition, $\Delta_{\mathrm{min}} \leq z_{(1-\alpha/2)}\sqrt{\frac{1}{2N_{\mathrm{SS}}}}$.
\end{theorem}

\begin{proof}
The empirical regret $\widehat L(\boldsymbol{\tau}) \in [0, 1]$ is a binomial proportion with variance:
\begin{align}
\mathrm{Var}\big(\widehat L(\boldsymbol{\tau})\big) &= \frac{L(\boldsymbol{\tau})\big(1 - L(\boldsymbol{\tau})\big)}{N_{\mathrm{SS}}}\,. 
\end{align}
where $L(\boldsymbol{\tau}) = \mathbb{E}[\widehat L (\boldsymbol{\tau})]$ is the true regret. Since we do not know $L(\boldsymbol{\tau})$, we cannot compute $\mathrm{Var}\big(\widehat L(\boldsymbol{\tau})\big)$.
Instead, we define the standard error of $\widehat{L}(\boldsymbol{\tau})$ by $\widehat \sigma_{\widehat{L}(\boldsymbol{\tau})}$, which is the square root of the estimated variance, therefore, $\sigma_{\widehat L (\boldsymbol{\tau})} =\sqrt{\frac{\widehat L(\boldsymbol{\tau})\big(1 - \widehat L(\boldsymbol{\tau})\big)}{N_{\mathrm{SS}}}}$. Note that, $\sigma_{\widehat L (\boldsymbol{\tau})}  \leq \sqrt{\frac{1}{4N_{\mathrm{SS}}}}$, where the equality is attained at $\widehat L (\boldsymbol{\tau}) = \frac{1}{2}$.

We define a \textbf{hypothesis test for distinguishability}. 
For thresholds $\boldsymbol{\tau}$ and $\boldsymbol{\tau}'$, define the $Z$-statistic
\begin{align}
Z = \frac{\widehat L(\boldsymbol{\tau}) - \widehat L(\boldsymbol{\tau}')}{\sqrt{\widehat \sigma_{\widehat L(\boldsymbol{\tau})}^2 + \widehat \sigma_{\widehat L(\boldsymbol{\tau}')}^2}}.
\end{align}
Under $H_0: L(\boldsymbol{\tau}) = L(\boldsymbol{\tau}')$, and $Z \sim \mathcal{N}(0,1)$. $H_0$ can be rejected at level $\alpha$ if $|Z| > z_{(1-\alpha/2)}$.

The critical value of $|\widehat L(\boldsymbol{\tau})- \widehat L(\boldsymbol{\tau}')|$, induced by this hypothesis test is termed Minimal Detectable Change (MDC) in empirical regret and is denoted by $\Delta_{\mathrm{min}}$.

\begin{align}
\Delta_{\mathrm{min}} &= z_{(1-\alpha/2)} \sqrt{\widehat \sigma_{\widehat L(\boldsymbol{\tau})}^2 + \widehat \sigma_{\widehat L(\boldsymbol{\tau}')}^2} \\
&\leqslant z_{(1-\alpha/2)} \sqrt{\frac{1}{2N_{\mathrm{SS}}}}.
\end{align}
As an example, for $\alpha=0.05$ and worst-case $\widehat L=0.5$, $\Delta_{\mathrm{min}} \leq 1.38/\sqrt{N_{\mathrm{SS}}}$.
  
\end{proof}

\newpage

\section{Inference Cost Model}
In our experiments, we used various LLAMA, QWEN, and GPT models.
All three GPT models are closed-source, but can be publicly accessed using the OpenAI API\footnote{\url{https://openai.com/api}}.
The LLAMA and QWEN models are open-source, although in practice, we used a commercial API service\footnote{\url{https://docs.nebius.com/studio/inference/api}} to query these models
due to hardware constraints.
The pricing information for LLAMA and QWEN can be found at \url{https://nebius.com/prices-ai-studio} and the GPT token cost details are available at \url{https://openai.com/api/pricing/}. The detailed per token costs for different LLM families are available in Tables~\ref{tab:llama_costs},~\ref{tab:qwen_costs},~\ref{tab:gpt_costs}.

For any single query, the total cost of sampling multiple CoTs is computed by adding the dollar costs for prompting, i.e., the cost of processing the system prompt and input prompt, and the total cost of generating multiple CoTs.
We conducted all experiments during April and May 2025 and used the pricing information available at that time for dollar cost accounting throughout the paper.

Although token costs change over time (typically at least once a year), we argue that this does not invalidate any conclusions from this work for the following reason. When pricing changes (usually becoming cheaper) the API providers ensure that
the ratio of token costs between different LLMs within the same family remains consistent before and after the pricing change as the main driver of costs is the model memory which is constant.

This guarantees that the dollar cost of a single query using a fixed prompt scales predictably, and that cost calculations for all baseline algorithms and C3PO are uniformly affected. For example, a 20\% reduction in token pricing corresponds to a 20\% reduction in cost across all methods considered in this paper. Therefore, any comparison between two LLM cascades yields the same conclusion regardless of pricing changes.

\begin{table}[h]
\centering
\caption{Cost per million tokens for LLaMA models. Price source: \url{https://nebius.com/prices-ai-studio}}
\label{tab:llama_costs}
\begin{tabular}{lcc}
\toprule
\textbf{LLaMA Model} & \textbf{Input Cost (\$/M tokens)} & \textbf{Output Cost (\$/M tokens)} \\
\midrule
LLaMA 3.2 1B-Instruct   & 0.005 & 0.01  \\
LLaMA 3.2 3B-Instruct   & 0.01  & 0.02  \\
LLaMA 3.3 70B-Instruct  & 0.13  & 0.40  \\
LLaMA 3.1 405B-Instruct & 1.00  & 3.00  \\
\bottomrule
\end{tabular}
\end{table}

\begin{table}[h]
\centering
\caption{Cost per million tokens for QWEN models. Price source: \url{https://nebius.com/prices-ai-studio}}
\label{tab:qwen_costs}
\begin{tabular}{lcc}
\toprule
\textbf{Qwen Model} & \textbf{Input Cost (\$/M tokens)} & \textbf{Output Cost (\$/M tokens)} \\
\midrule
Qwen 2.5 1B-Instruct  & 0.02 & 0.06 \\
Qwen 2.5 32B-Instruct  & 0.06 & 0.20 \\
Qwen 2.5 72B-Instruct  & 0.13 & 0.40 \\
\bottomrule
\end{tabular}
\end{table}

\begin{table}[h]
\centering
\caption{Cost per million tokens for GPT models. Price source: \url{https://platform.openai.com/docs/pricing}}
\label{tab:gpt_costs}
\begin{tabular}{lcc}
\toprule
\textbf{GPT Model} & \textbf{Input Cost (\$/M tokens)} & \textbf{Output Cost (\$/M tokens)} \\
\midrule
GPT 3.5-Turbo & 0.50 & 1.50 \\
GPT 4o-mini    & 0.15 & 0.60 \\
OpenAI o3-mini     & 1.10 & 4.40 \\
\bottomrule
\end{tabular}
\end{table}

\newpage
\section{Complete Experimental Results}\label{app:full_results}
In Figure~\ref{fig:date} of the main paper, we compared the performance of all baselines and the proposed C3PO on 6 datasets using the LLAMA cascade. Here, we present the complete experimental results on 16 datasets and 3 LLM cascades. Accuracy vs. cost plots using LLAMA, QWEN, and GPT cascades on 16 datasets are shown in Figures~\ref{fig:main_llama},~\ref{fig:main_qwen}, and~\ref{fig:main_gpt} respectively. In addition, for each of the 3 LLM cascades, we include summary boxplots to compare the distributions of required inference costs to reach near-MPM accuracy across 16 datasets, to compare C3PO with the baseline algorithms. These boxplots for the LLAMA, QWEN and GPT cascades are shown in Figures~\ref{fig:boxplotllama},~\ref{fig:boxplotqwen}, and~\ref{fig:boxplotgpt} respectively.

\subsection{LLAMA Results}

\begin{figure}[H]
    \centering
\includegraphics[width=\linewidth]{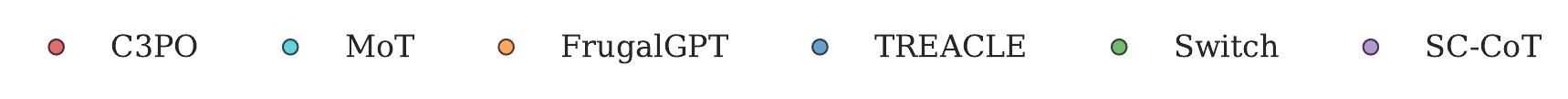}
\includegraphics[width=0.23\textwidth]{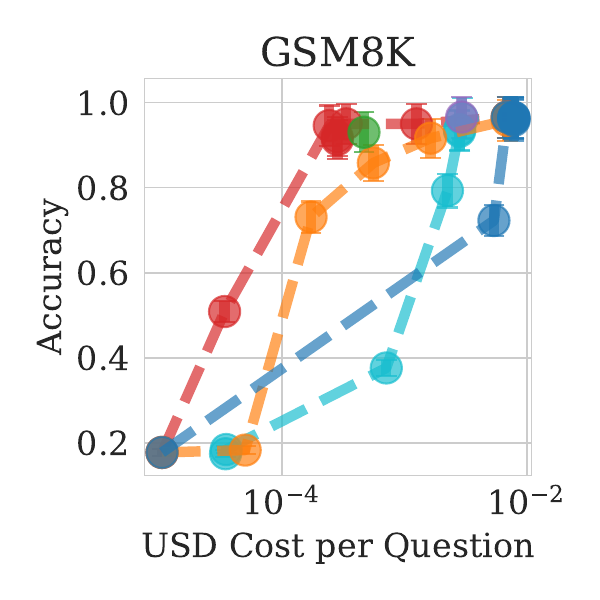}
\includegraphics[width=0.23\textwidth]{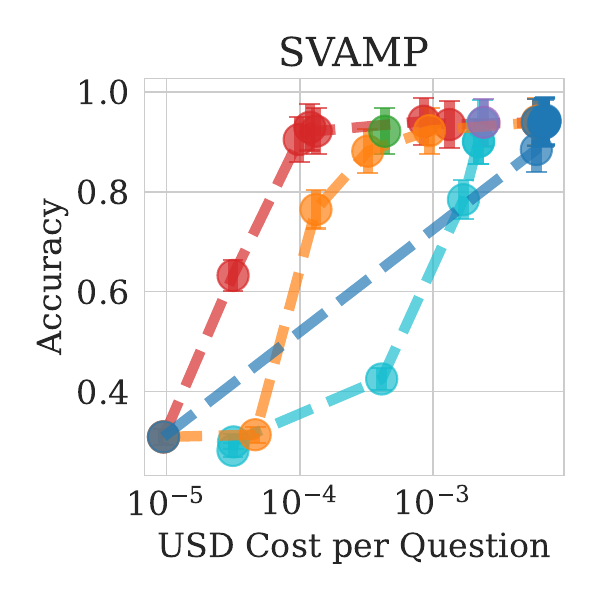}
\includegraphics[width=0.23\textwidth]{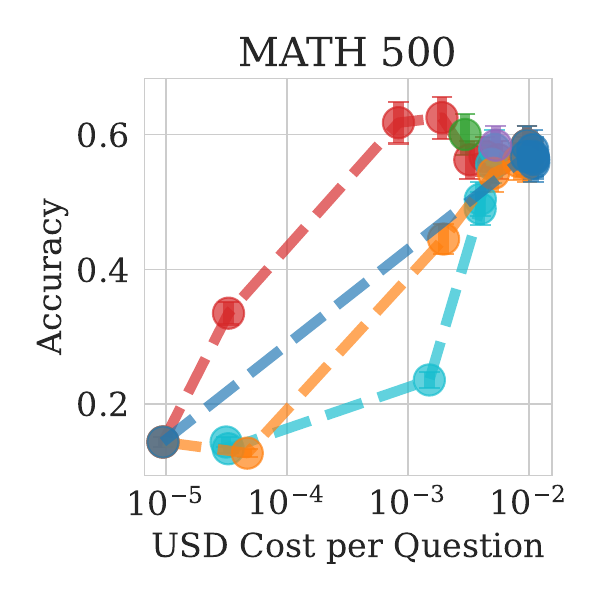}
\includegraphics[width=0.23\textwidth]{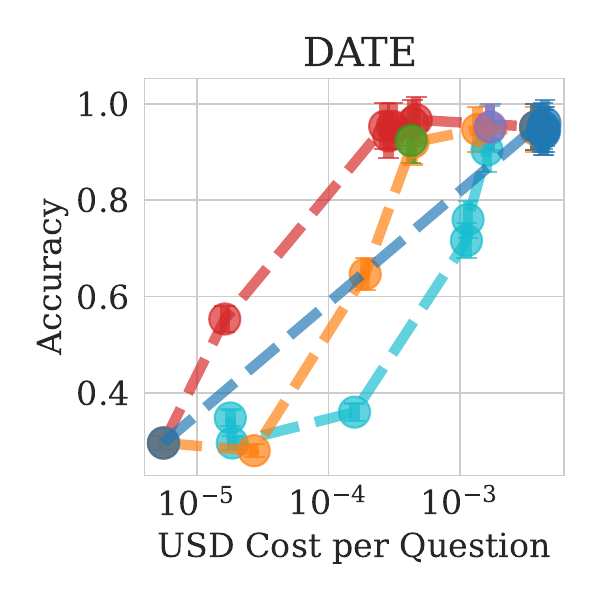}
\end{figure}

\begin{figure}[H]
    \centering
\includegraphics[width=0.23\textwidth]{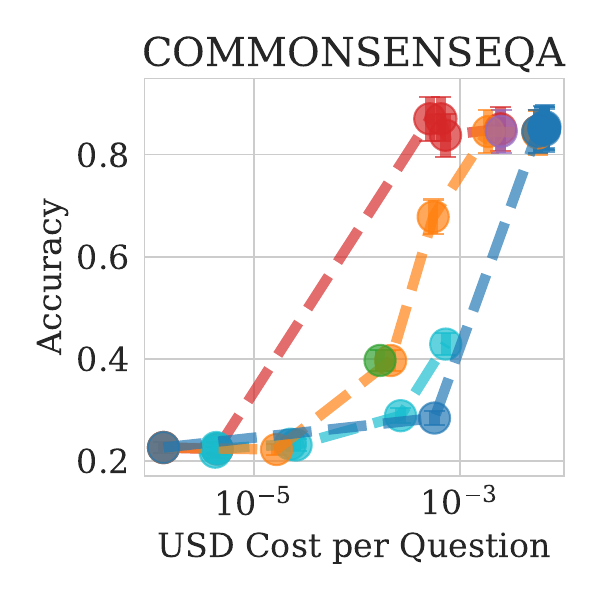}
\includegraphics[width=0.23\textwidth]{figs/aqua_90_CI_LLAMA.pdf}
\includegraphics[width=0.23\textwidth]{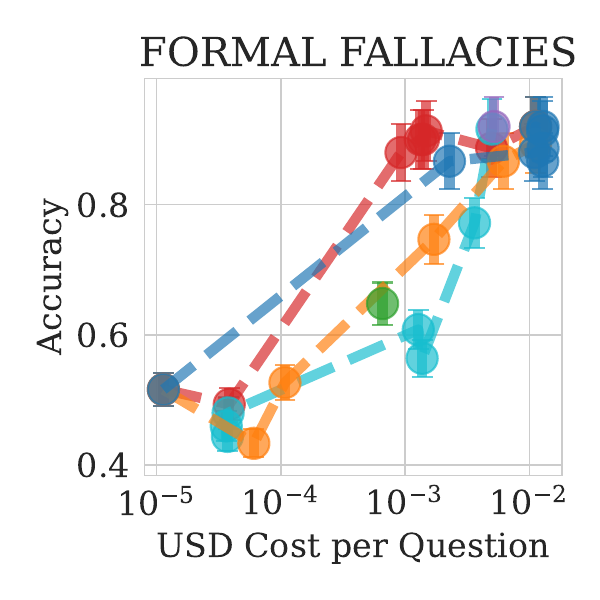}
\includegraphics[width=0.23\textwidth]{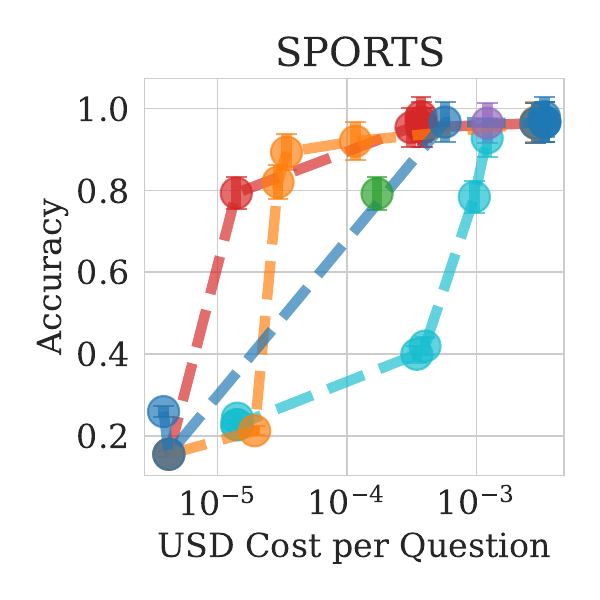}
\end{figure}

\begin{figure}[H]
    \centering
\includegraphics[width=0.23\textwidth]{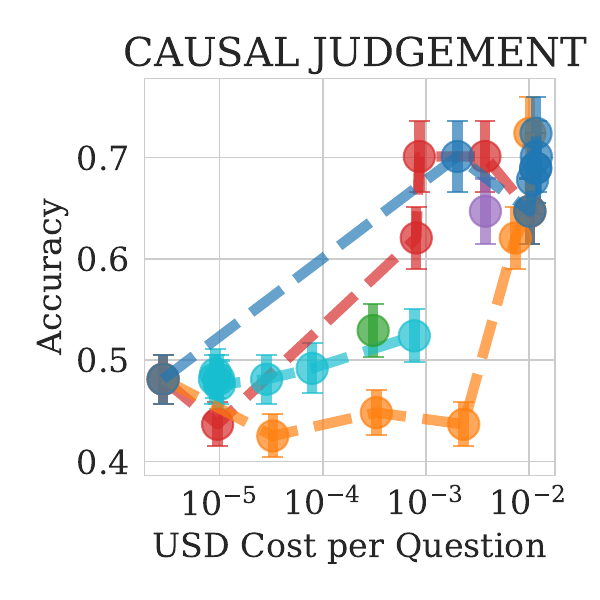}
\includegraphics[width=0.23\textwidth]{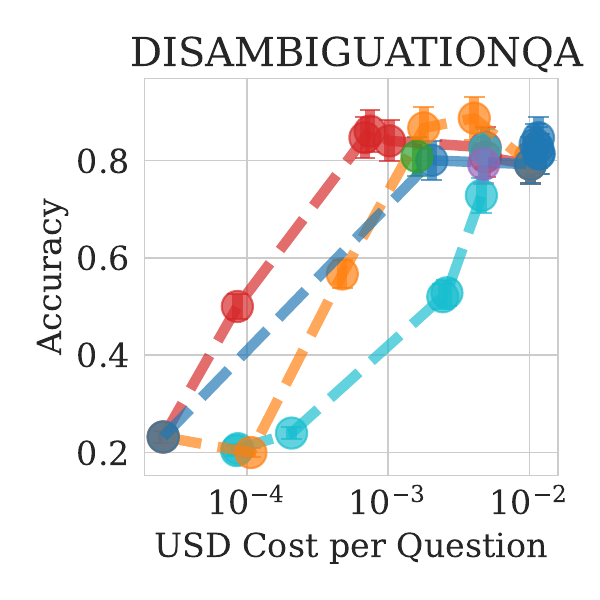}
\includegraphics[width=0.23\textwidth]{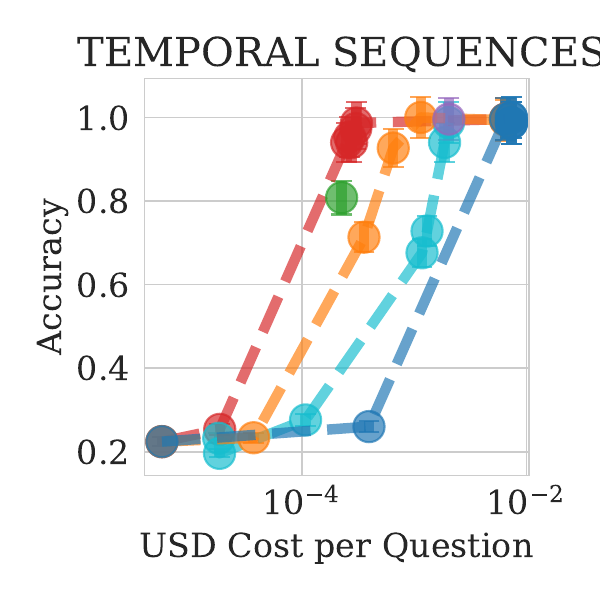}
\includegraphics[width=0.23\textwidth]{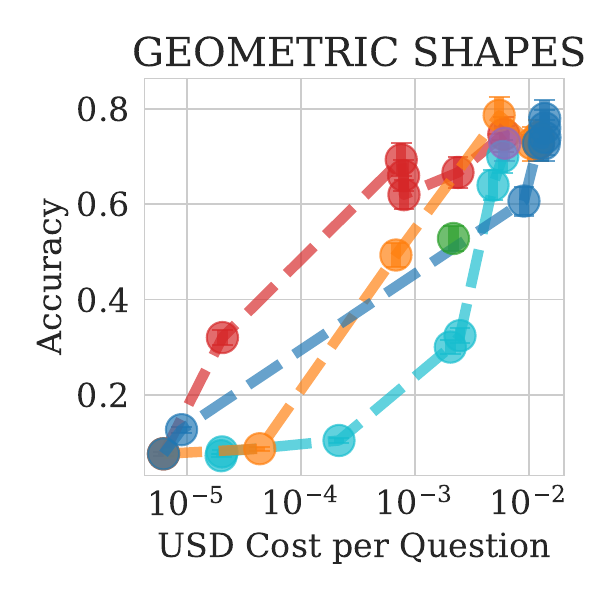}
\end{figure}

\begin{figure}[H]
    \centering
\includegraphics[width=0.23\textwidth]{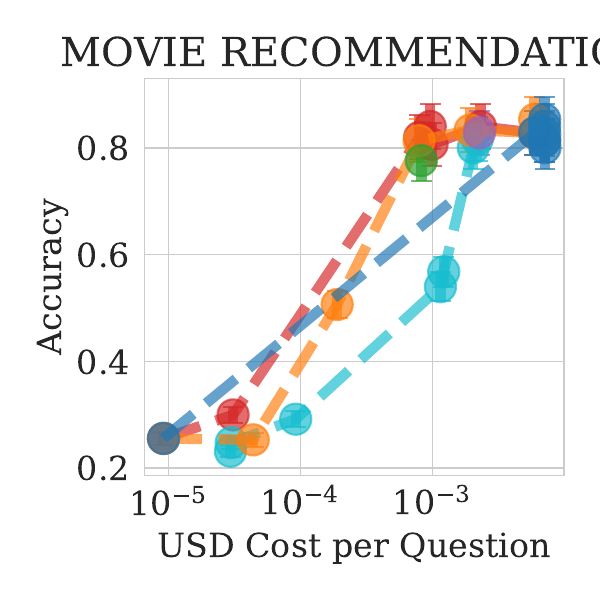} \includegraphics[width=0.23\textwidth]{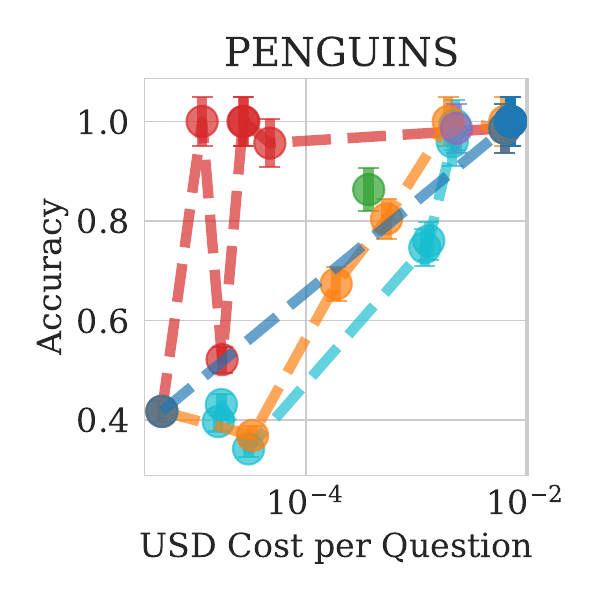}
\includegraphics[width=0.23\textwidth]{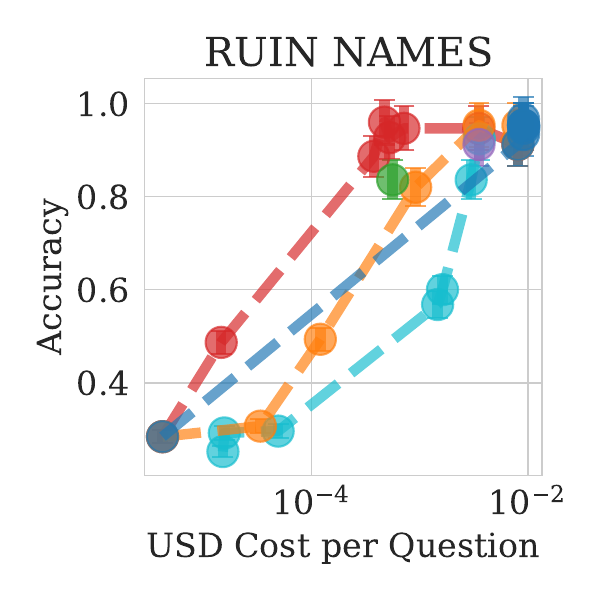}
\includegraphics[width=0.23\textwidth]{figs/bigbench_snarks_90_CI_LLAMA.pdf}
\caption{Accuracy vs. dollar cost of different algorithms for 16 datasets using the LLAMA cascade}
\label{fig:main_llama}
\end{figure}
\begin{figure}[h]
    \centering
\includegraphics[trim={0 1.15cm 0 0},clip, width=0.8\linewidth]{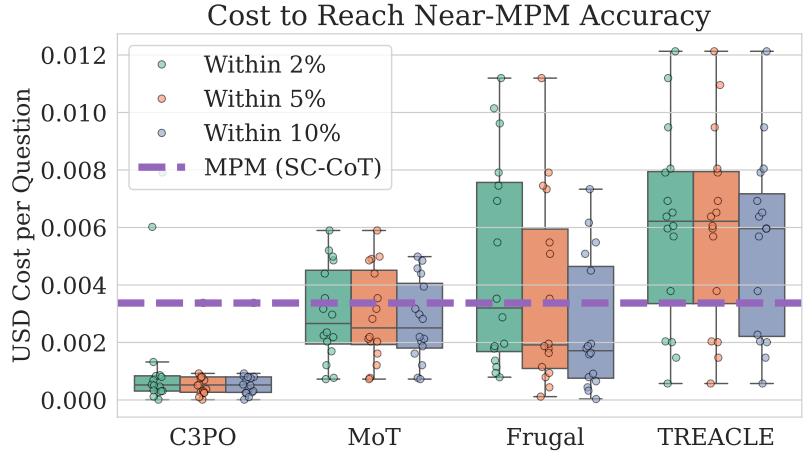}
    \caption{Surpassing existing cascade approaches tremendously, C3PO offers markedly superior cost-effectiveness across 16 benchmarks, requiring \textbf{less than 20\% of the cost of the most powerful model (MPM)} (cost shown in purple) for an accuracy gap of at most 2, 5, and 10\% using a LLAMA cascade. In this boxplot, each dot represents a dataset and the whiskers extend to 90\% coverage.}
    \label{fig:boxplotllama}
\end{figure}

\begin{figure}[H]
    \centering
\includegraphics[width=0.23\textwidth]{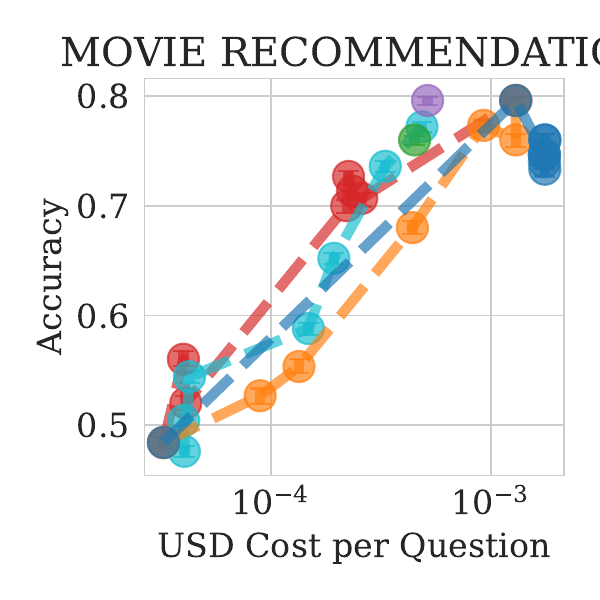} \includegraphics[width=0.23\textwidth]{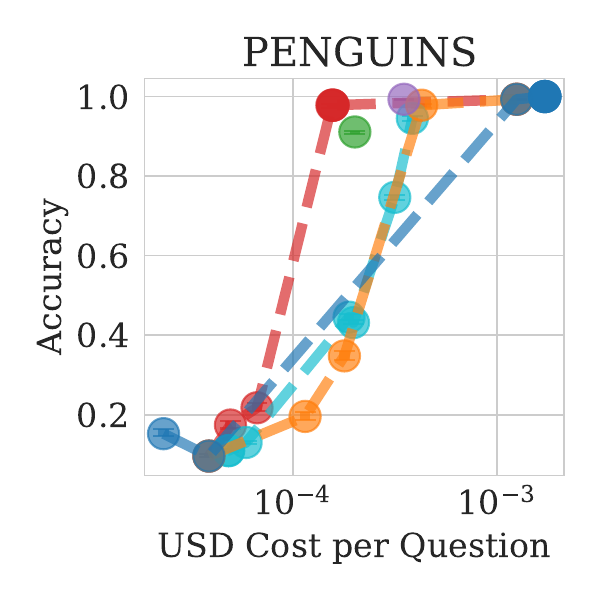}
\includegraphics[width=0.23\textwidth]{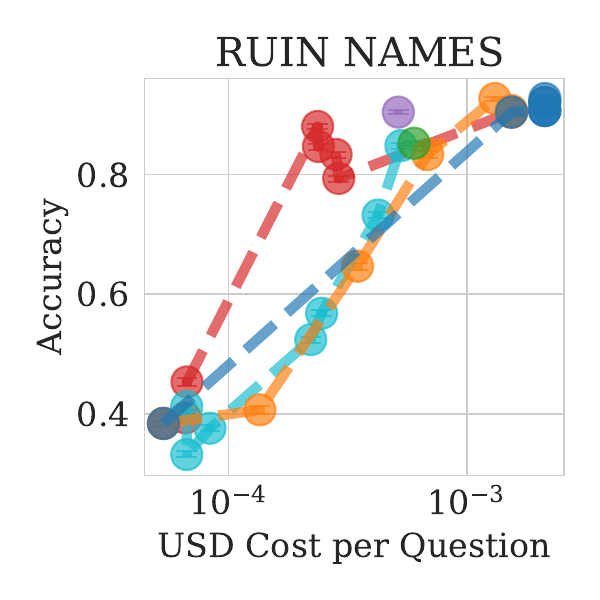}
\includegraphics[width=0.23\textwidth]{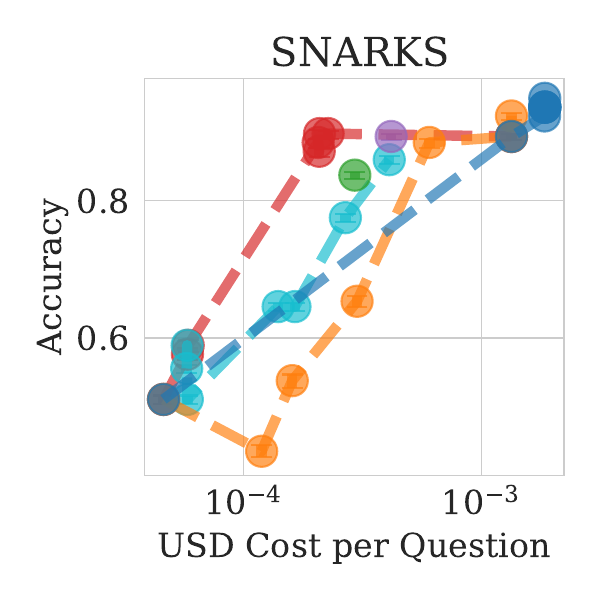}
\caption{Accuracy vs. dollar cost of different algorithms for 16 datasets using the QWEN cascade}
\label{fig:main_qwen}
\end{figure}

\begin{figure}[h]
    \centering
\includegraphics[trim={0 1.15cm 0 0},clip, width=0.8\linewidth]{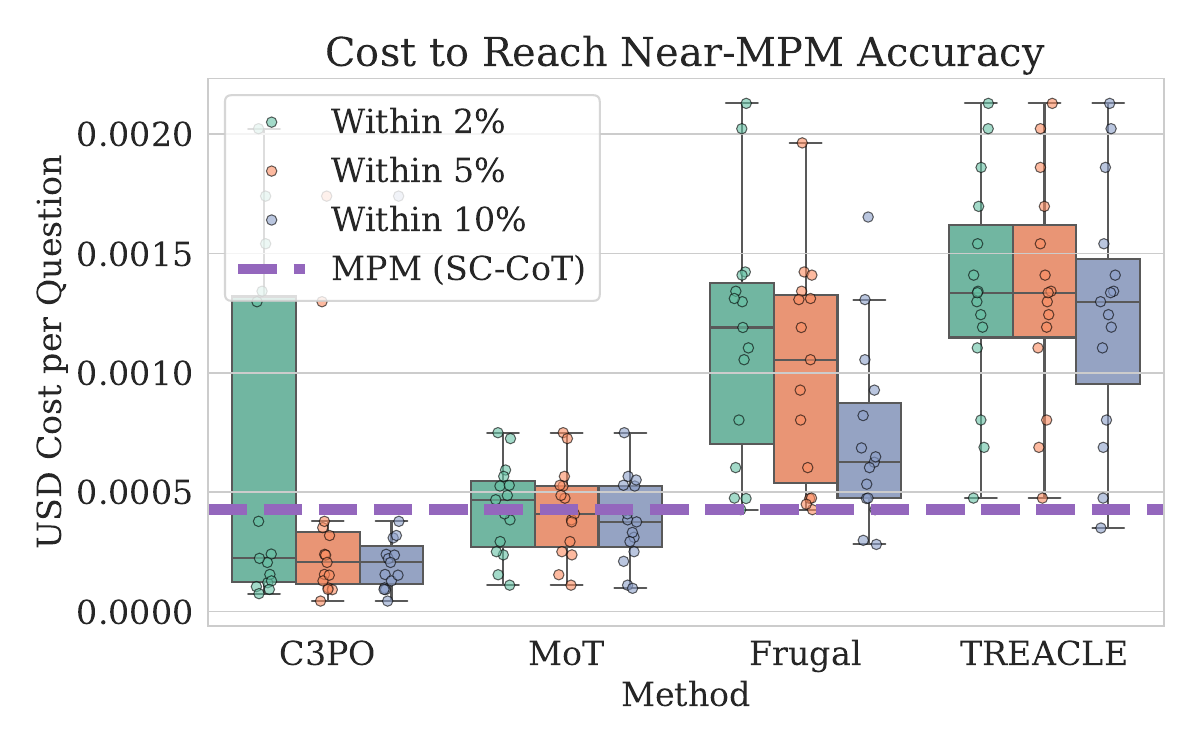}
    \caption{Surpassing existing cascade approaches tremendously, C3PO offers markedly superior cost-effectiveness across 16 benchmarks, requiring \textbf{less than 60\% of the cost of the most powerful model (MPM)} (cost shown in purple) for an accuracy gap of at most 2, 5, and 10\% using a QWEN cascade. In this boxplot, each dot represents a dataset and the whiskers extend to 90\% coverage.}
    \label{fig:boxplotqwen}
\end{figure}

\begin{figure}[h]
    \centering
\includegraphics[trim={0 1.15cm 0 0},clip, width=0.8\linewidth]{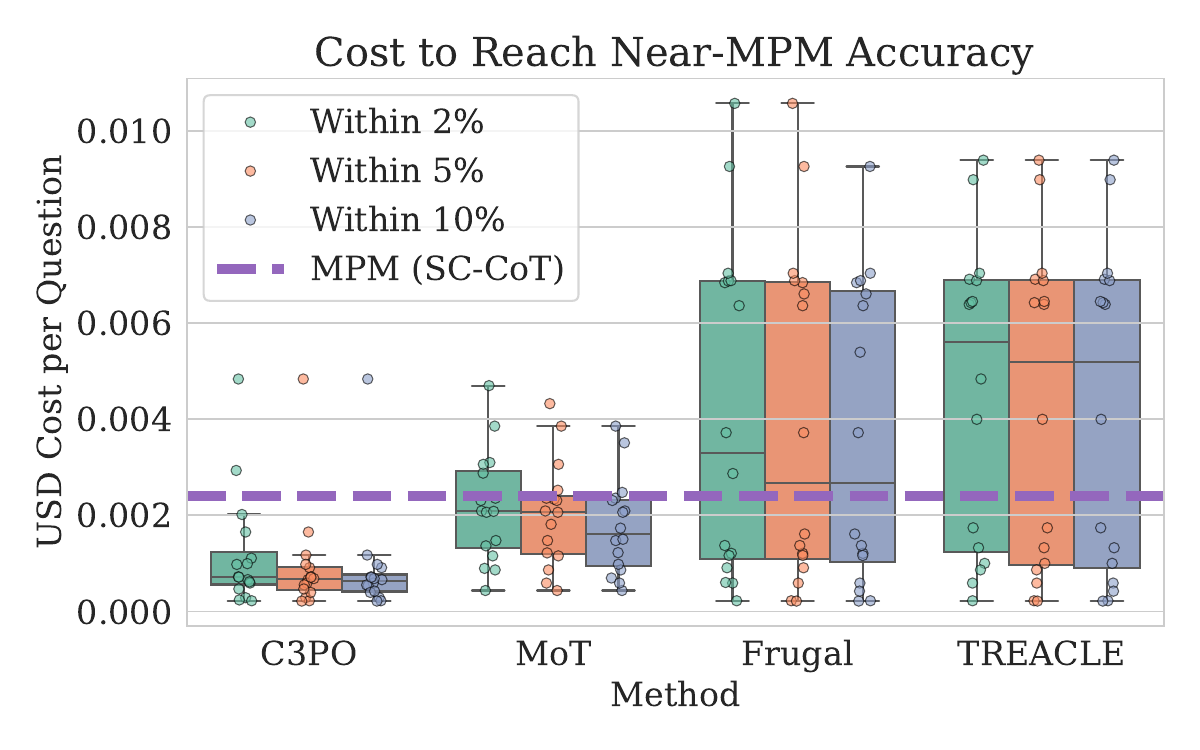}
    \caption{Surpassing existing cascade approaches tremendously, C3PO offers markedly superior cost-effectiveness across 16 benchmarks, requiring \textbf{less than 50\% of the cost of the most powerful model (MPM)} (cost shown in purple) for an accuracy gap of at most 2, 5, and 10\% using a GPT cascade. In this boxplot, each dot represents a dataset and the whiskers extend to 90\% coverage.}
    \label{fig:boxplotgpt}
\end{figure}

From Figure~\ref{fig:main_llama}, we observe that the proposed C3PO outperforms its competitors for most datasets and cost configurations. The accuracy improvement offered by C3PO in comparison to baselines is more pronounced for relatively lower dollar costs, demonstrating C3PO's efficacy in resource-constrained 
settings. This is further supported by Figure~\ref{fig:boxplotllama}, which shows that across different datsets, C3PO requires significantly lower costs compared to the baselines to achieve near-MPM accuracy. 
Similar conclusions can be drawn for QWEN and GPT cascades from Figures~\ref{fig:main_qwen},~\ref{fig:boxplotqwen},~\ref{fig:main_gpt}, and~\ref{fig:boxplotgpt}. 

Note that for the GPT cascade, we use o3-mini as the MPM as it is the most expensive of the three GPT models we use to form the cascade. However, for some datasets, e.g. \textit{sports, disambiguationQA, movie 
recommendation}, o3-mini's accuracy is significantly lower than that of a much cheaper GPT-4o-mini. As a result, accuracies of all algorithms deteriorate with increasing budgets for those datasets.

\clearpage
\subsection{QWEN Results}\label{app:qwen}

\begin{figure}[H]
    \centering
\includegraphics[width=\linewidth]{figs/method_legend.pdf}
\includegraphics[width=0.23\textwidth]{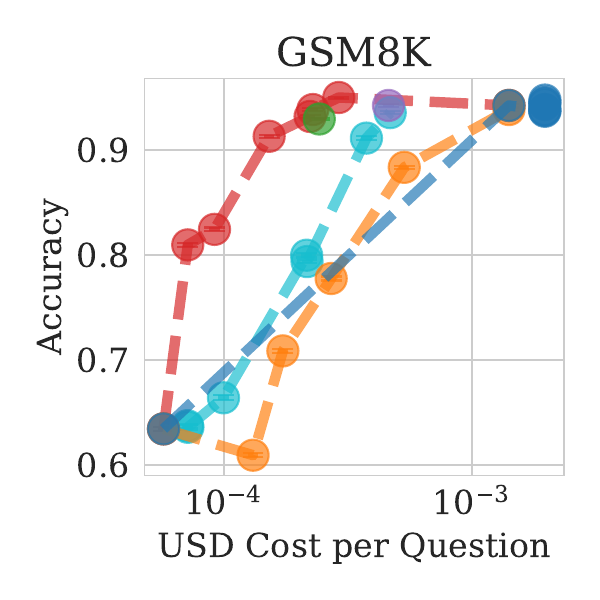}
\includegraphics[width=0.23\textwidth]{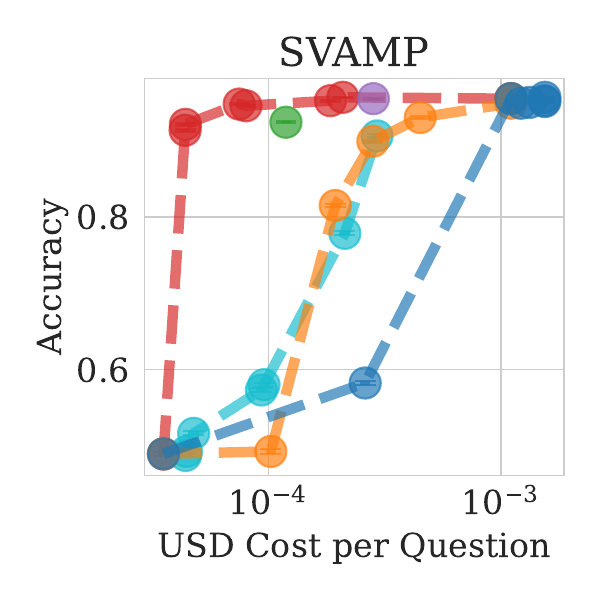}
\includegraphics[width=0.23\textwidth]{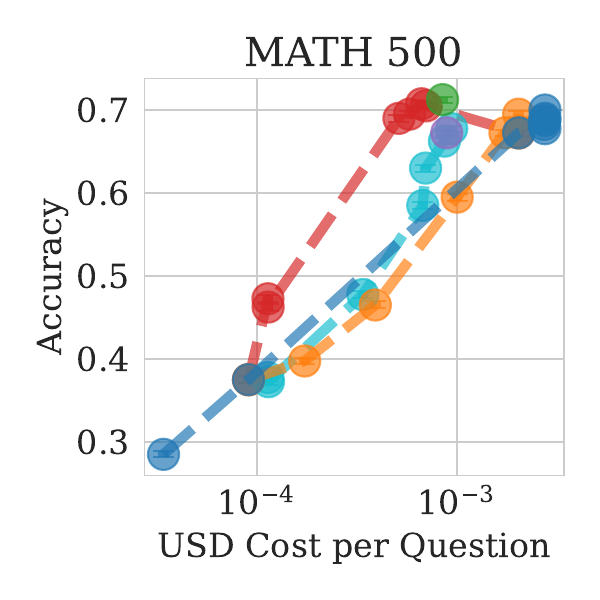}
\includegraphics[width=0.23\textwidth]{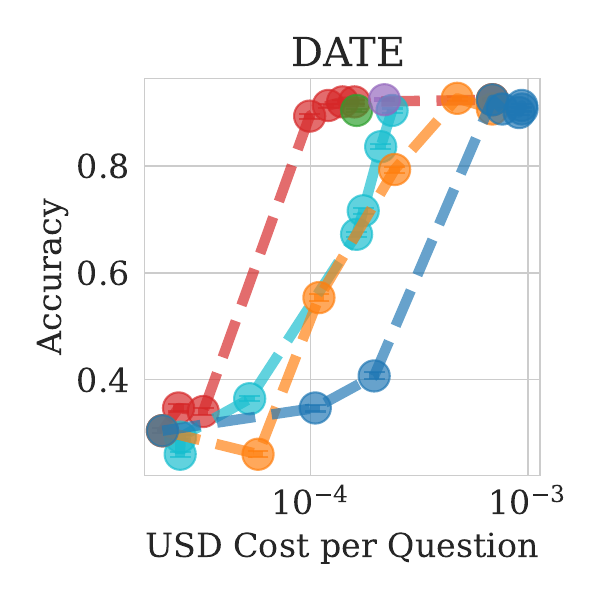}
\end{figure}

\begin{figure}[H]
    \centering
\includegraphics[width=0.23\textwidth]{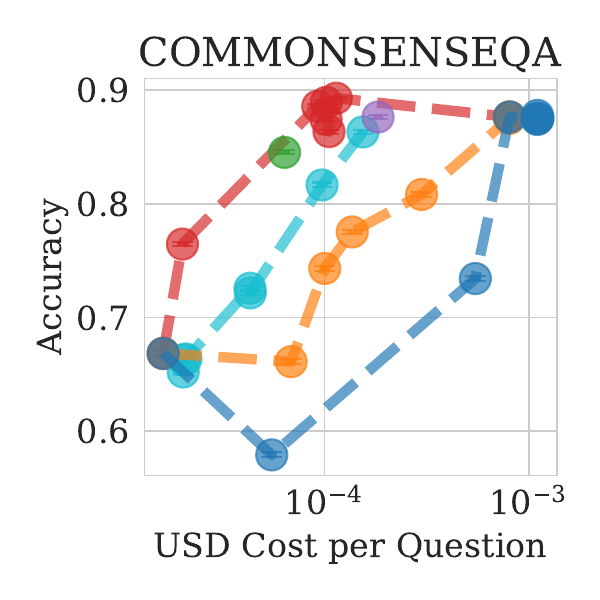}
\includegraphics[width=0.23\textwidth]{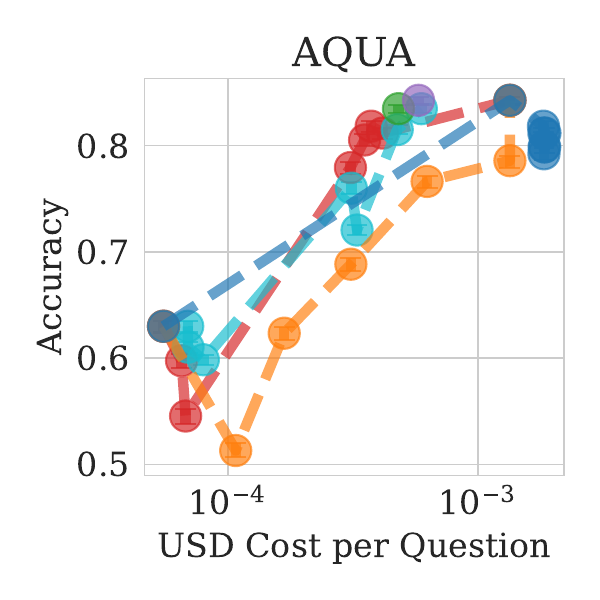}
\includegraphics[width=0.23\textwidth]{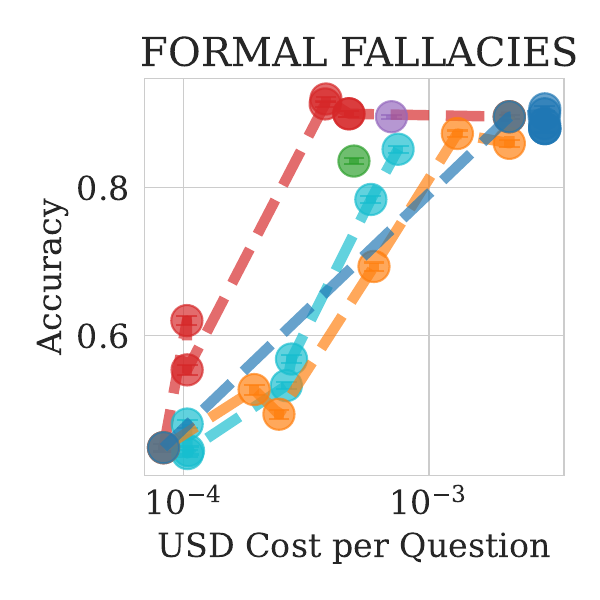}
\includegraphics[width=0.23\textwidth]{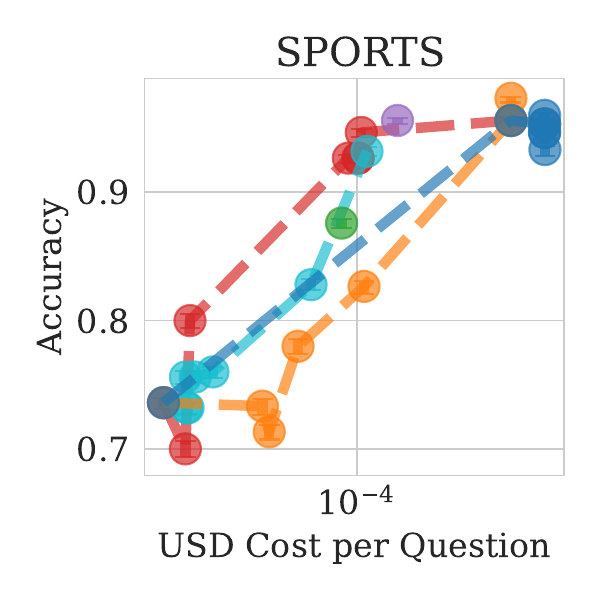}
\end{figure}

\begin{figure}[H]
    \centering
\includegraphics[width=0.23\textwidth]{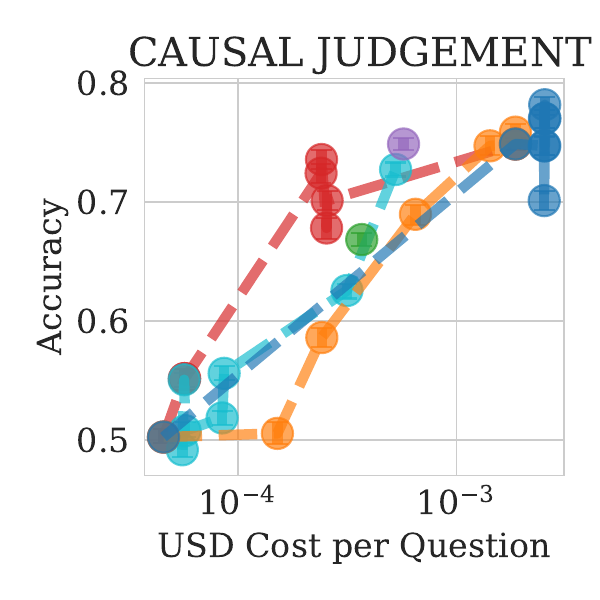}
\includegraphics[width=0.23\textwidth]{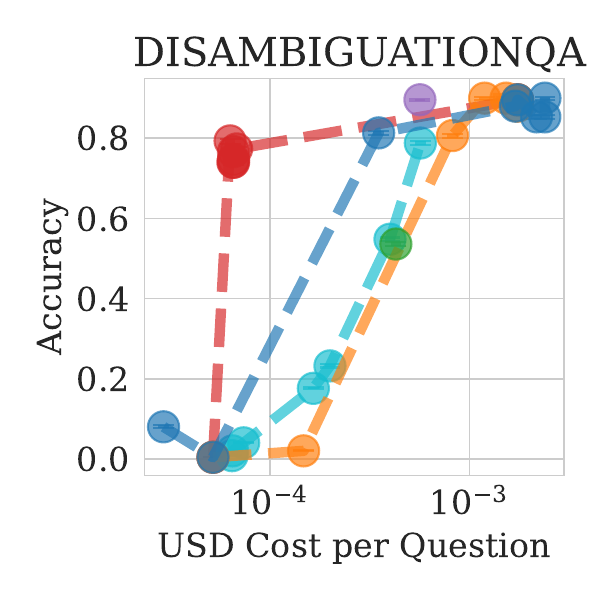}
\includegraphics[width=0.23\textwidth]{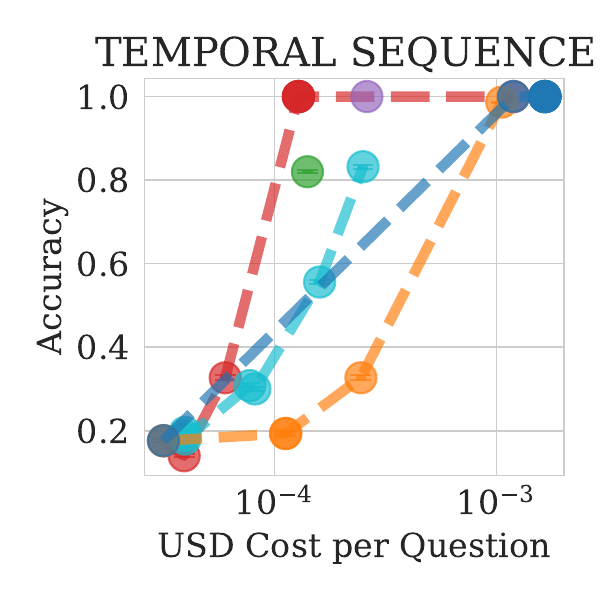}
\includegraphics[width=0.23\textwidth]{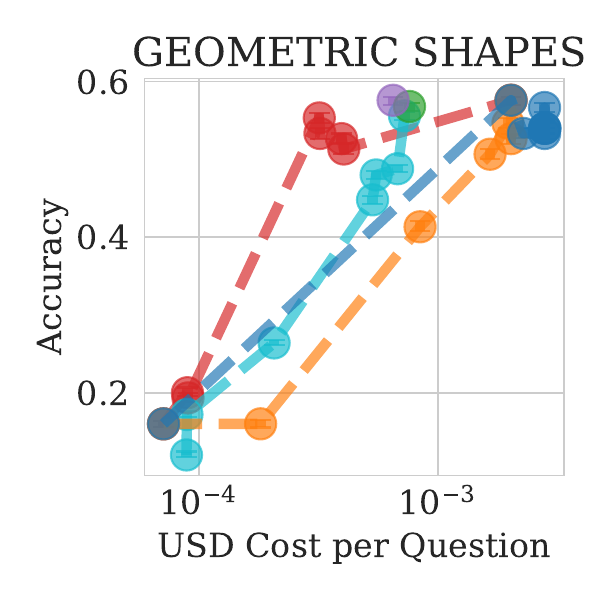}
\end{figure}

\newpage
\newpage
\subsection{GPT Results}\label{app:gpt}

\begin{figure}[H]
    \centering
\includegraphics[width=\linewidth]{figs/method_legend.pdf}
\includegraphics[width=0.23\textwidth]{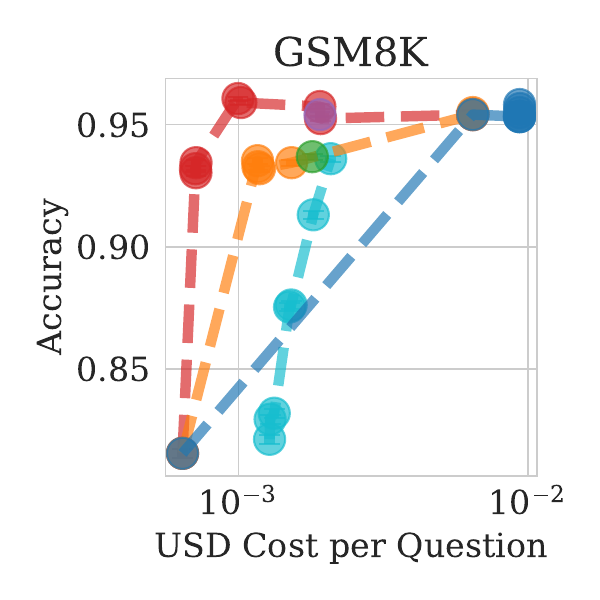}
\includegraphics[width=0.23\textwidth]{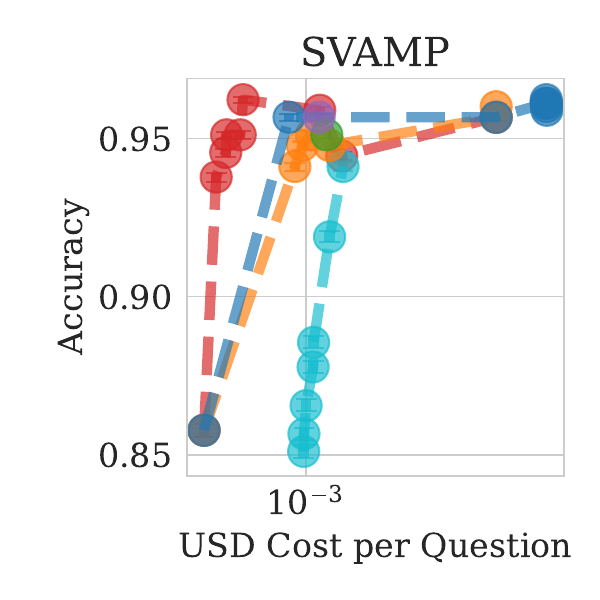}
\includegraphics[width=0.23\textwidth]{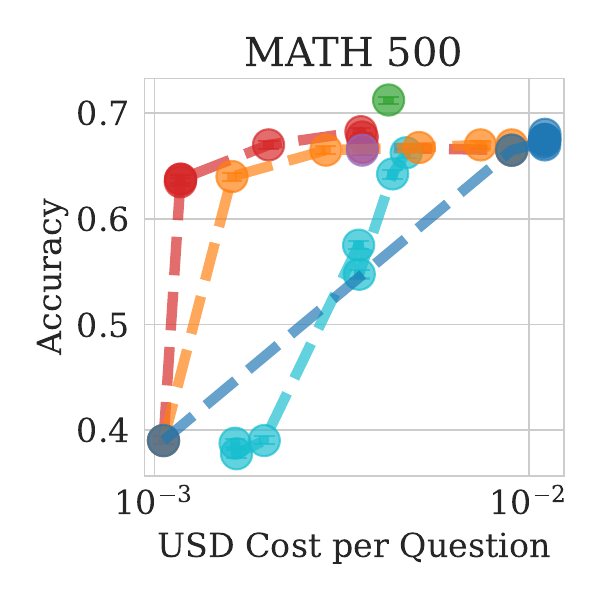}
\includegraphics[width=0.23\textwidth]{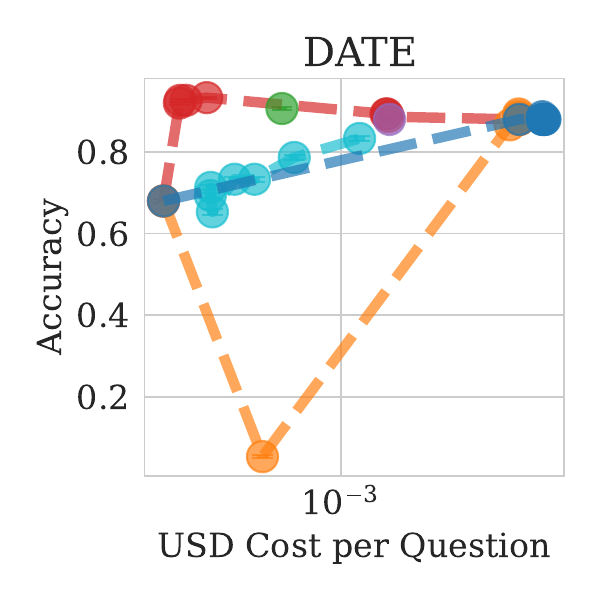}
\end{figure}

\begin{figure}[H]
    \centering
\includegraphics[width=0.23\textwidth]{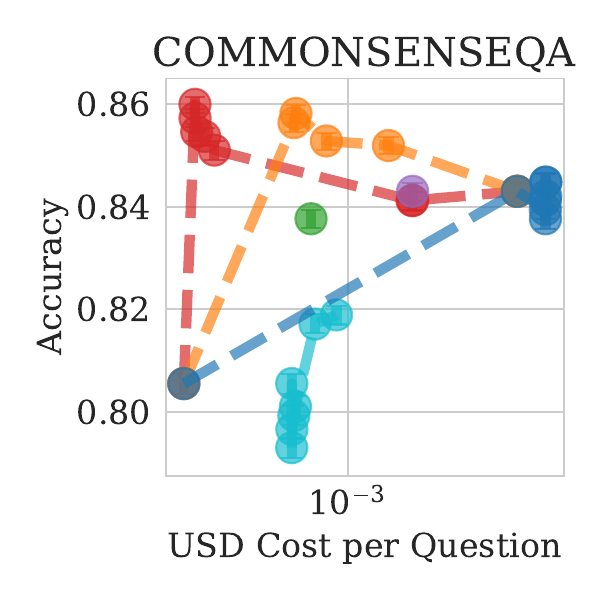}
\includegraphics[width=0.23\textwidth]{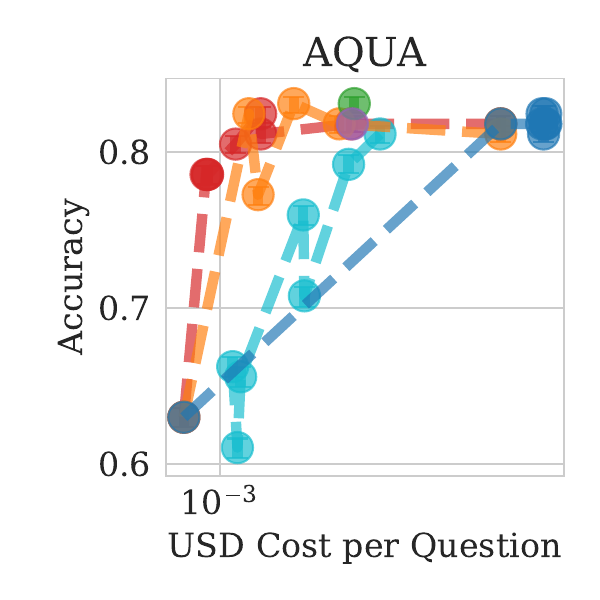}
\includegraphics[width=0.23\textwidth]{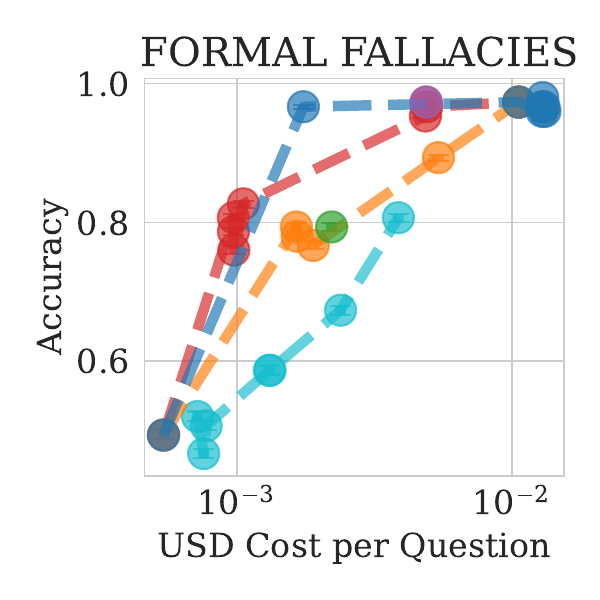}
\includegraphics[width=0.23\textwidth]{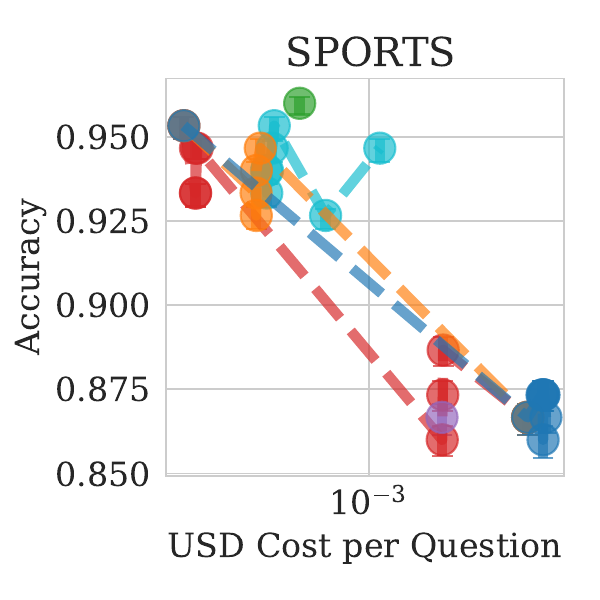}
\end{figure}

\begin{figure}[H]
    \centering
\includegraphics[width=0.23\textwidth]{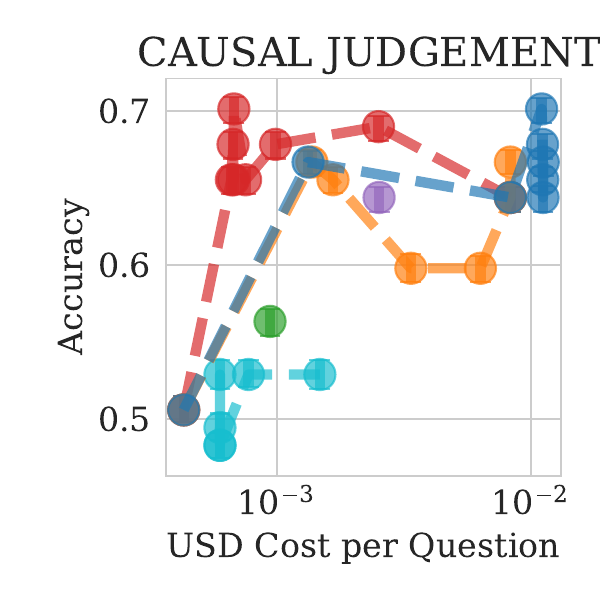}
\includegraphics[width=0.23\textwidth]{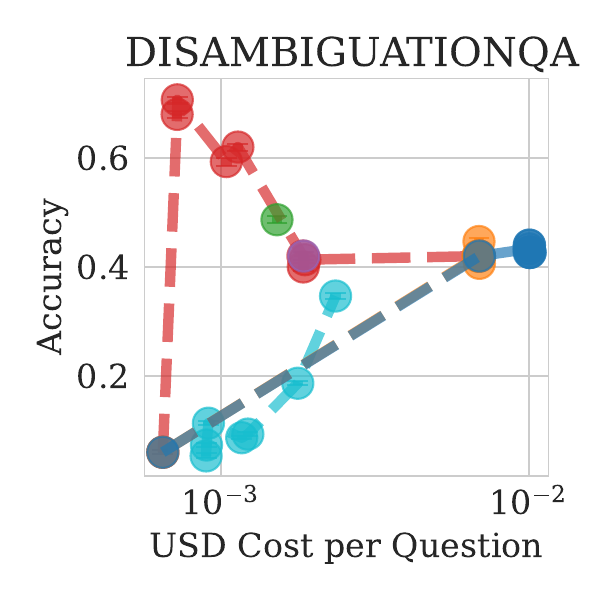}
\includegraphics[width=0.23\textwidth]{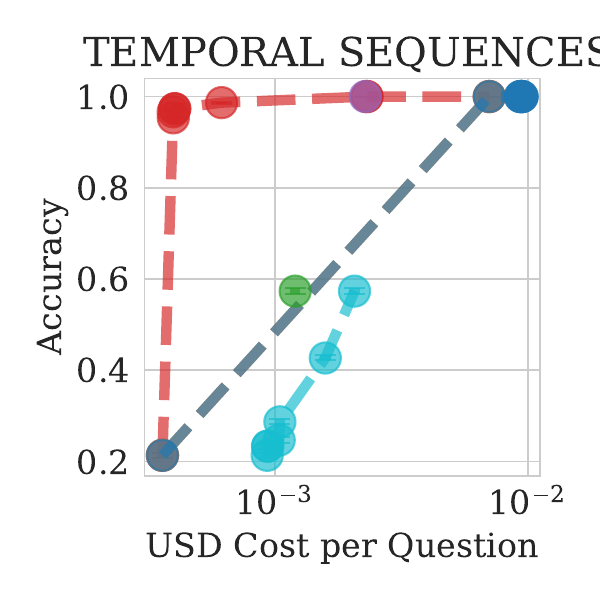}
\includegraphics[width=0.23\textwidth]{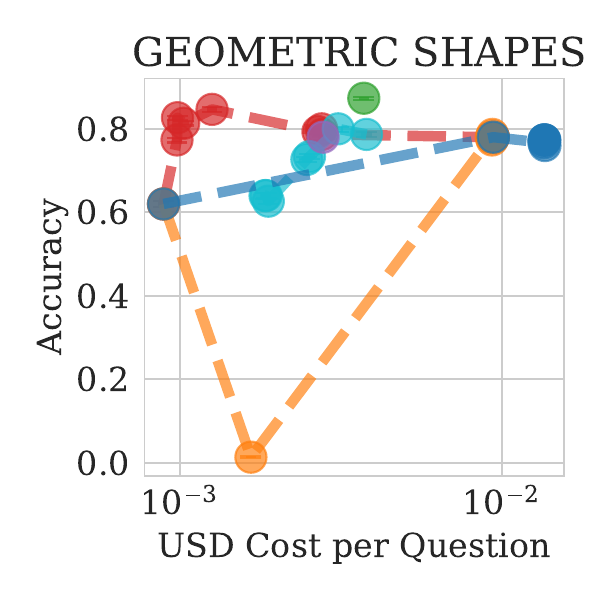}
\end{figure}

\begin{figure}[H]
    \centering
\includegraphics[width=0.23\textwidth]{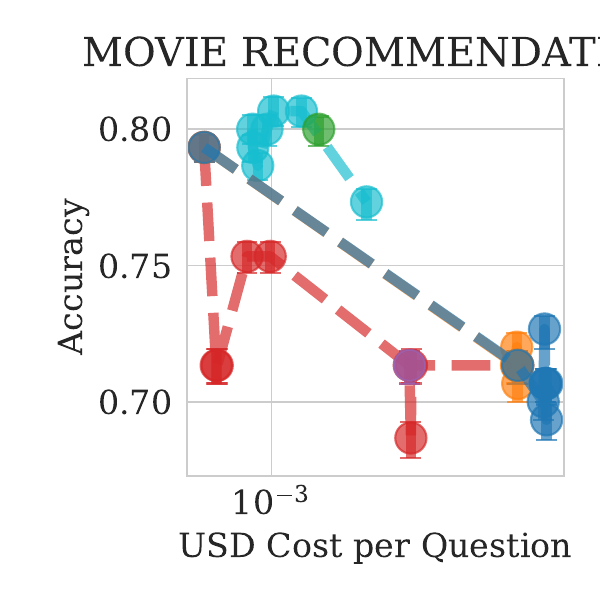} \includegraphics[width=0.23\textwidth]{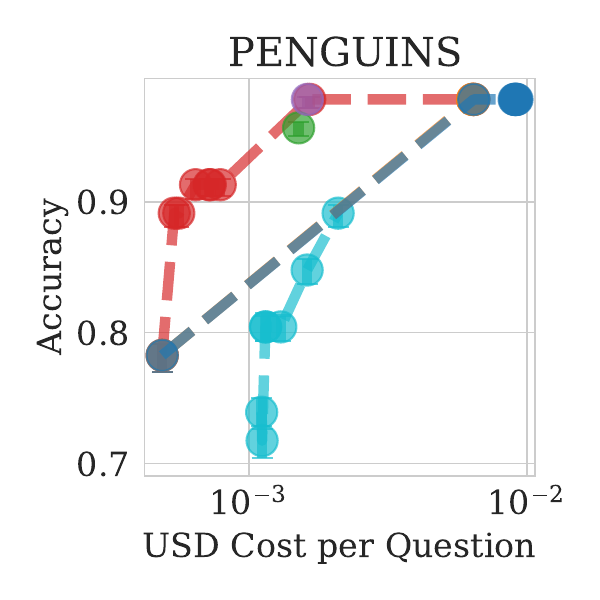}
\includegraphics[width=0.23\textwidth]{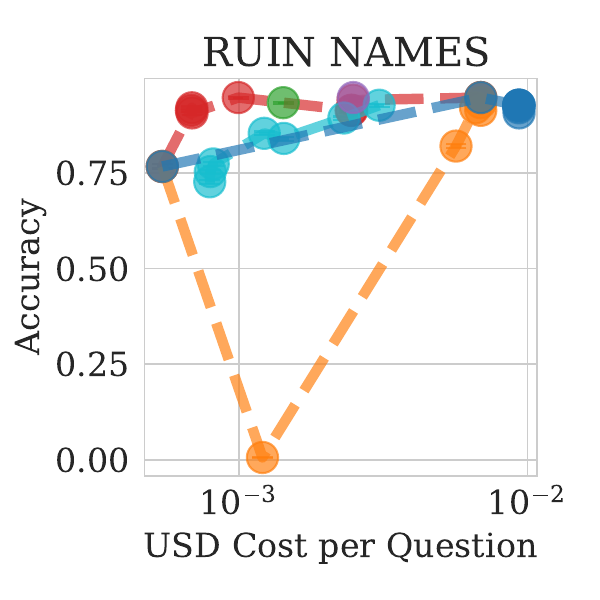}
\includegraphics[width=0.23\textwidth]{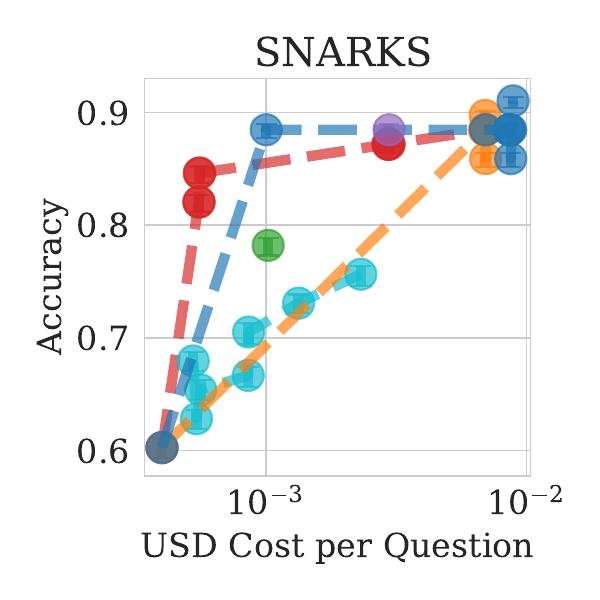}
\caption{Accuracy vs. dollar cost of different algorithms for 16 datasets using the GPT cascade}
\label{fig:main_gpt}
\end{figure}

\newpage
\begin{figure}
    \centering
\includegraphics[width=0.5\linewidth]{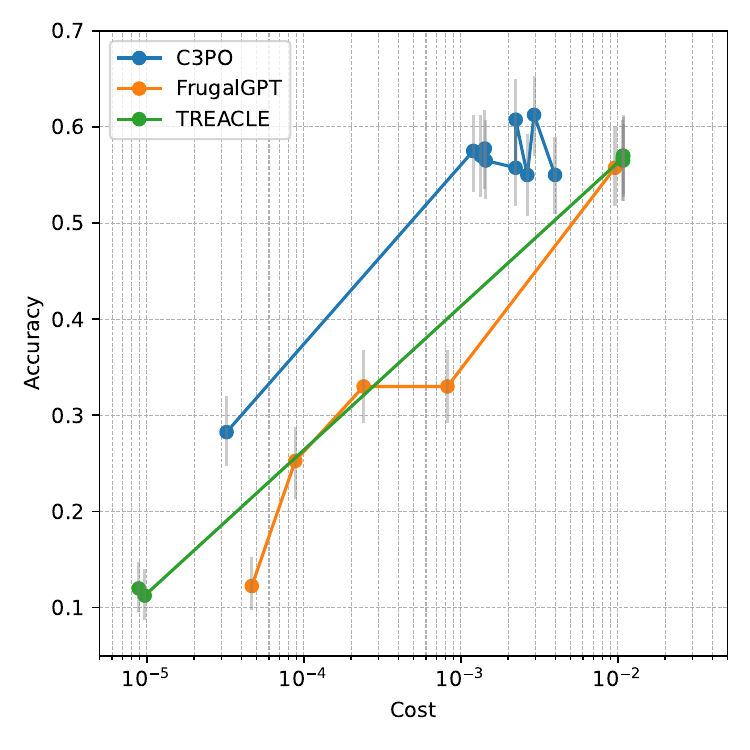}    \caption{Distribution shift experiment. Training on SVAMP, testing on MATH500 shows that C3PO has the best domain adaptation performance.}
    \label{fig:dist_shift_2}
\end{figure}
\clearpage

\subsection{Comparison of C3PO and baselines in terms of performance  and budget allocation for questions with varying difficulties in MATH-500}
\begin{figure}[ht]
    \centering
    \includegraphics[width=0.9\linewidth, trim={0cm 0 5cm 0},clip]{figs/method_legend.pdf}
\includegraphics[width=0.8\linewidth]{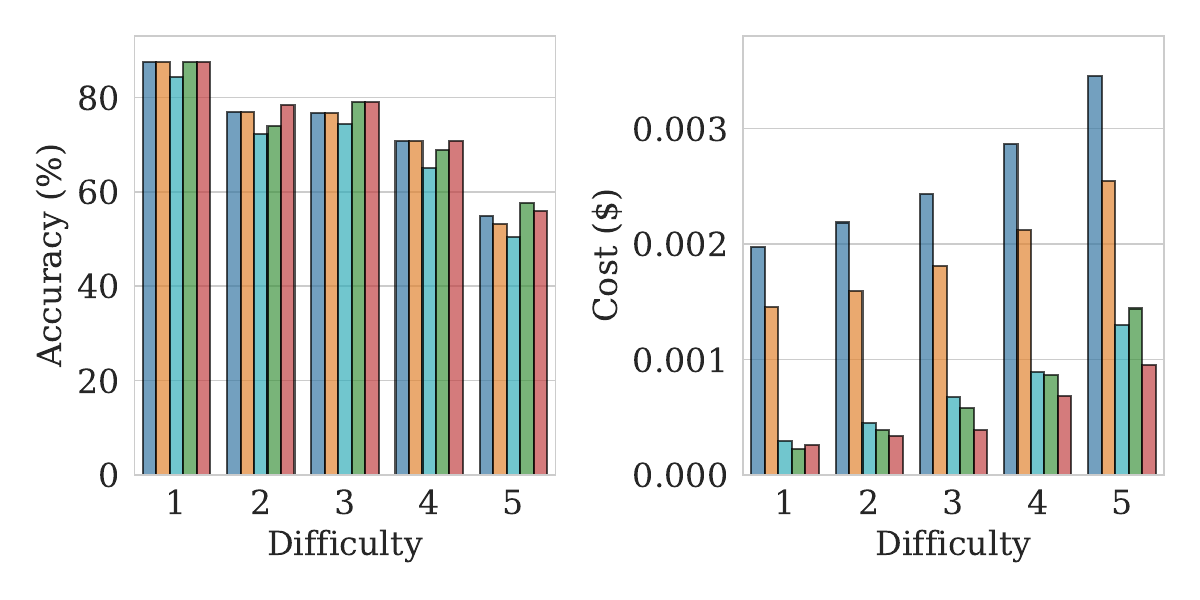}
\includegraphics[width=0.8\linewidth]{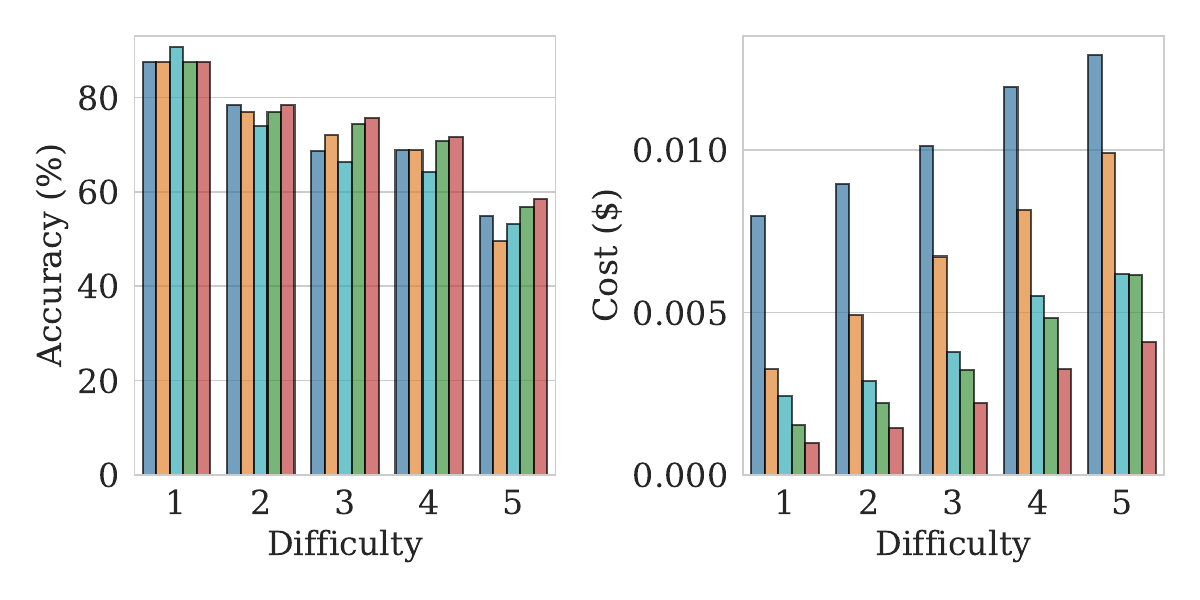}
    \caption{Average accuracies and dollar costs of different algorithms for five different difficulty levels of the MATH-500 dataset using the QWEN (top) and GPT (bottom) cascade.}
    \label{fig:difficulty_MATH 500_qwen_gpt}
\end{figure}

In Figure~\ref{fig:difficulty_MATH 500_llama} in the main paper, we showed that using the LLAMA cascade, C3PO outperforms most of the baselines at almost all difficulty levels with significantly lower inference costs.
Similar results are obtained for the QWEN and GPT cascades and are shown in Figure~\ref{fig:difficulty_MATH 500_qwen_gpt}.

\newpage

\section{Few-shot Prompting Strategy and Prompt Examples}

To ensure a fair comparison across all evaluated methods, we use a fixed few-shot prompting strategy that provides consistent formatting and reasoning guidance. Each model receives the same demonstration examples, with uniform instructions, answer formatting, and step-by-step rationales.

The prompt begins with a general instruction to use the boxed answer format (in LaTeX) and then presents a sequence of fully worked-out examples covering diverse mathematical topics, such as geometric series, completing the square, and factoring. This context is followed by the final question to be solved. 

Below, we show an excerpt of the few-shot prompts used in our evaluations of free form answer datasets:

\begin{tcolorbox}[colback=blue!1!white,
                  colframe=blue!255!black,
                  title=Free Form Answer Mathematical Question Prompt Template (MATH-500),
                  fonttitle=\bfseries,
                  coltitle=white,
                  boxrule=0.8pt,
                  arc=5pt,
                  top=3pt,
                  bottom=3pt,
                  left=3pt,
                  right=3pt,
                  ]

\small\ttfamily
You are given a mathematical question. Solve the following problem and output correct result in a given format. Output answer at the end, use format $\backslash$boxed$\{$answer in LaTeX$\}$ \boxed{ <answer\quad in \quad LaTeX> }. \\
Think step-by-step.

\vspace{2em}
Question: Kevin Kangaroo begins hopping on a number line at 0. He wants to get to 1, but he can hop only $\frac{1}{3}$ of the distance. Each hop tires him out so that he continues to hop $\frac{1}{3}$ of the remaining distance. How far has he hopped after five hops? Express your answer as a common fraction. \\
\vspace{0.5em}
Let's think step by step. Kevin hops $1/3$ of the remaining distance with every hop. His first hop takes $1/3$ closer. For his second hop, he has $2/3$ left to travel, so he hops forward $(2/3)(1/3)$. For his third hop, he has $(2/3)^2$ left to travel, so he hops forward $(2/3)^2(1/3)$. In general, Kevin hops forward $(2/3)^{k-1}(1/3)$ on his $k$th hop. We want to find how far he has hopped after five hops. This is a finite geometric series with first term $1/3$, common ratio $2/3$, and five terms. Thus, Kevin has hopped $\frac{\frac{1}{3}\left(1-\left(\frac{2}{3}\right)^5\right)}{1-\frac{2}{3}} = \boxed{\frac{211}{243}}$. Thus, the answer is $\boxed{\frac{211}{243}}$.

\vspace{2em}
Question: What is the area of the region defined by the equation $x^2+y^2-7 = 4y-14x+3$? \\
\vspace{0.5em}
Let's think step by step. We rewrite the equation as $x^2 + 14x + y^2 - 4y = 10$ and then complete the square, resulting in $(x+7)^2-49 + (y-2)^2-4=10$, or $(x+7)^2+(y-2)^2=63$. This is the equation of a circle with center $(-7, 2)$ and radius $\sqrt{63}$, so the area of this region is $\pi r^2 = \boxed{63\pi}$.

\vspace{2em}
Question: If $x^2+y^2=1$, what is the largest possible value of $|x|+|y|$? \\
\vspace{0.5em}
Let's think step by step. If $(x,y)$ lies on the circle, so does $(x,-y),$ $(-x,-y),$ and $(-x,y)$, (which all give the same value of $|x| + |y|$), so we can assume that $x \ge 0$ and $y \ge 0$. Then $|x| + |y| = x + y$. Squaring, we get $(x + y)^2 = x^2 + 2xy + y^2 = 1 + 2xy$. Note that $(x - y)^2 \ge 0$. Expanding, we get $x^2 - 2xy + y^2 \ge 0$, so $2xy \le x^2 + y^2 = 1$. Hence, $1 + 2xy \le 2$, which means $x + y \le \sqrt{2}$. Equality occurs when $x = y = \frac{1}{\sqrt{2}}$, so the maximum value of $|x| + |y|$ is $\boxed{\sqrt{2}}$.

\vspace{2em}
Question: \{\{ question \}\} \\
Let's think step by step.
\label{fig:prompt_math_500}
\end{tcolorbox}

We also used datasets with multiple choice questions. For these datasets such as CommonSenseQA we used prompts of the style shown on the next page.

\begin{tcolorbox}[colback=blue!1!white,
                  colframe=blue!255!black,
                  title=Multiple Choice Answer Prompt Template (CommonSenseQA),
                  fonttitle=\bfseries,
                  coltitle=white,
                  boxrule=0.8pt,
                  arc=5pt,
                  top=3pt,
                  bottom=3pt,
                  left=3pt,
                  right=3pt,
                  ]

\small\ttfamily
You are given the following question. Solve it correctly and output a single letter (A, B, C, D, E or F) corresponding to the correct answer to the question. Output answer at the end, use format $\backslash$boxed$\{$single letter (A, B, C, D, E)$\}$ \boxed{ <single letter (A, B, C, D, E)> } \\
Think step-by-step.

\vspace{0.5em}
Question: What do people use to absorb extra ink from a fountain pen? \\
Answer Choices: \\
A) shirt pocket \\
B) calligrapher's hand \\
C) inkwell \\
D) desk drawer \\
E) blotter \\
The answer must be an item that can absorb ink. Of the above choices, only blotters are used to absorb ink. \\
Thus, the answer is \boxed{E}.

\vspace{0.5em}
Question: What home entertainment equipment requires cable? \\
Answer Choices: \\
A) radio shack \\
B) substation \\
C) television \\
D) cabinet \\
The answer must require cable. Of the above choices, only television requires cable. \\
It means the correct answer is \boxed{C}.

$\vdots$

\vspace{0.5em}
Question: Google Maps and other highway and street GPS services have replaced what? \\
Answer Choices: \\
A) united states \\
B) mexico \\
C) countryside \\
D) atlas \\
The answer must be something that used to do what Google Maps and GPS services do, which is to give directions. Of the above choices, only atlases are used to give directions. \\
The correct answer is \boxed{D}.

\vspace{0.5em}
Question: Before getting a divorce, what did the wife feel who was doing all the work? \\
Answer Choices: \\
A) harder \\
B) anguish \\
C) bitterness \\
D) tears \\
E) sadness \\
The answer should be the feeling of someone getting divorced who was doing all the work. Of the above choices, the closest feeling is bitterness. \\
Thus, the answer is \boxed{C}.

\vspace{0.5em}
Question: \{\{ question \}\} \\
The answer

\end{tcolorbox}
Shown above are the prompts for MATH-500 and CommonSenseQA datasets as examples of the prompt templates for free form and multiple choice reasoning, respectively.
All LLMs employed in our study are prompted with the exact same few-shot demonstrations for evaluating all baselines and our proposed method. This strategy ensures a fair evaluation setting, where the differences in empirical performances between any pair of LLM-cascades can be solely attributed to the effectiveness of their decision rules and crucially not to the utilization of unfair prompting advances.

\newpage

\section{Case Study and Analysis}

\subsection{Question Level Case Study} 
This case study highlights a geometric reasoning problem where the goal is to compute the sum of distances from the intersection points of a circle and hyperbola to a fixed external point. While smaller models like LLaMA 1B and 3B produce inconsistent or incorrect answers, and TREACLE escalates unnecessarily to the largest (and most expensive) model, our method (C3PO) successfully identifies the smallest model (LLaMA 70B) that produces the correct answer. This demonstrates C3PO’s ability to balance cost and accuracy by leveraging confidence-based early exits with probabilistic guarantees.
\begin{tcolorbox}[colback=blue!1!white,
                colframe=blue!255!black, 
  title=\textbf{Case Study: Geometric Intersection Problem (MATH-500 Question \#106)}, 
  fonttitle=\bfseries, arc=4pt, boxrule=0.5pt]

\textbf{Question:} The graphs of $x^2 + y^2 + 6x - 24y + 72 = 0$ and $x^2 - y^2 + 6x + 16y - 46 = 0$ intersect at four points. Compute the sum of the distances from these four points to the point $(-3,2)$.

\vspace{0.5em}
\textbf{Model Responses and Sample Rationales:}

\begin{itemize}
  \item \textbf{LLaMA 1B:} 
  ``8'', ``6'', ``18.67'', ``18.24'', ``$\sqrt{7}$''\\
  \textit{Sample rationale:} To solve this problem, we need to find the intersection points and then calculate the distances from these points to the point $(-3,2)$. [...] The intersection points are estimated and plugged into the distance formula. The model computes the distance and returns $\boxed{18.67}$, the answer is wrong due to numerical errors.

  \item \textbf{LLaMA 3B:} 
  ``$75^\circ$'', ``$2\sqrt{41}+18$'', ``17'', ``$10\sqrt{3}+2\sqrt{19}$'', ``40''\\
  \textit{Sample rationale:} The model attempts trigonometric simplifications unrelated to the geometric problem, mistakenly treating it as an angle computation task. The conclusion $\boxed{75^\circ}$ is clearly off-topic, illustrating reasoning drift.

  \item \textbf{LLaMA 70B:} 
  ``40'', ``$16\sqrt{5}$'', ``40'', ``40'', ``24''\\
  \textit{Sample rationale:} The model rewrites both equations into standard circle and hyperbola form, identifies the center $(-3,12)$ and point $(-3,2)$ as vertically aligned, and correctly computes total distance as $4 \times 10 = \boxed{40}$.

  \item \textbf{LLaMA 405B:} 
  ``40'', ``36'', ``40'', ``40'', ``40''\\
  \textit{Sample rationale:} Completing the square, the model identifies the circle centered at $(-3,12)$ and hyperbola centered at $(-3,8)$, recognizes symmetric geometry and uses translation and distance argument to correctly deduce total distance is $\boxed{40}$.
\end{itemize}

\vspace{0.5em}
\textbf{Cascade Behavior Analysis:}
\begin{itemize}
  \item \textbf{FrugalGPT} exits too early at \textbf{LLaMA 3B} and returns an incorrect answer.
  \item \textbf{TREACLE} escalates to the \textbf{LLaMA 405B} model and answers correctly, but at significantly higher cost.
  \item \textbf{C3PO (Ours)} exits at \textbf{LLaMA 70B}, which is the \textit{cheapest} model to consistently return the correct answer (\textbf{``40''}).
\end{itemize}

\textbf{Why C3PO wins:} It optimally balances cost and accuracy. While weaker models return a mix of incorrect and inconsistent answers, and TREACLE incurs maximum cost, C3PO identifies the minimal model in the cascade capable of producing the correct answer.

\end{tcolorbox}

\newpage

\subsection{Detailed study of C3PO's cost efficiency}  
Although Figures~\ref{fig:main_llama},~\ref{fig:main_qwen}, and~\ref{fig:main_gpt} demonstrate C3PO's superior cost efficiency compared to baselines at the dataset level, they do not provide explicit and fine-grained information of C3PO's cost allocation across different questions within each dataset.

Here, we conduct a detailed study to examine how C3PO spends its inference cost budget for questions where it succeeds in providing the correct answer and where it fails to do so. Specifically, for each baseline and LLM-cascade, we divide the test set into the following four non-overlapping categories; {\color{red} `very bad'}, {\color{orange} `bad'}, {\color{cyan} `good'}, and {\color{green} `very good'}. A question falls into the  {\color{red} `very bad'} category if C3PO fails to answer it correctly in spite of spending a higher inference cost compared to a baseline. If C3PO provides an incorrect answer to a question but incurs a lower cost compared to a baseline as well, then that question is {\color{orange} `bad'}. Clearly, having a {\color{orange} `bad'} question is better than having a {\color{red} `very bad'} question in terms of efficient usage of inference cost.
Similarly, {\color{cyan} `good'} questions refer to those test set examples, where C3PO provides  correct responses at the expense of increased cost compared to a baseline. Finally,  {\color{green} `very good'} questions are the ones, for which a correct answer from C3PO coincides with a lower cost compared to a baseline. 

\begin{figure}[h!]
    \centering
    \includegraphics[width=0.9\linewidth, trim={7cm 19cm 5cm 19cm},clip]{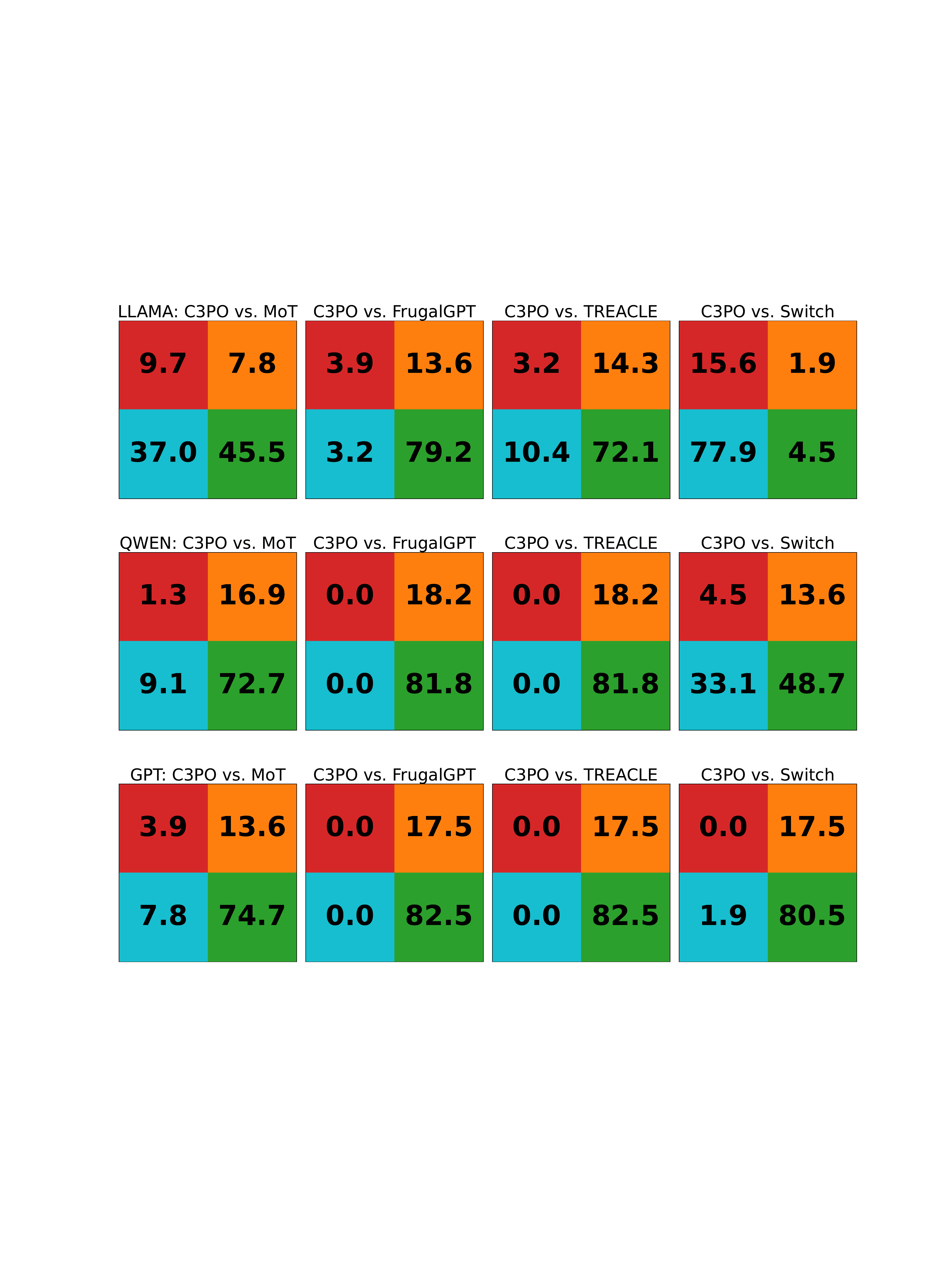}
    \caption{{Percentage of \color{red} `very bad'}, {\color{orange} `bad'}, {\color{cyan} `good'}, and {\color{green} `very good'} questions for each baseline and LLM cascade from the \textbf{AQuA} dataset}
    \label{fig:good_bad_ugly_aqua}
\end{figure}

We report the distribution of these four categories of questions in the AQuA dataset for each LLM cascade and baselines in Figure~\ref{fig:good_bad_ugly_aqua}. We observe that in comparison to most baselines, when C3PO provides an incorrect answer, it is more likely to save inference cost. Additionally, for the majority of the questions, where C3PO's answer matches the true answer, it incurs reduced cost. This demonstrates the universal nature of C3PO's cost effectiveness across different types of questions. 

\begin{figure}[h!]
    \centering
    \includegraphics[width=0.9\linewidth, trim={7cm 19cm 5cm 19cm},clip]{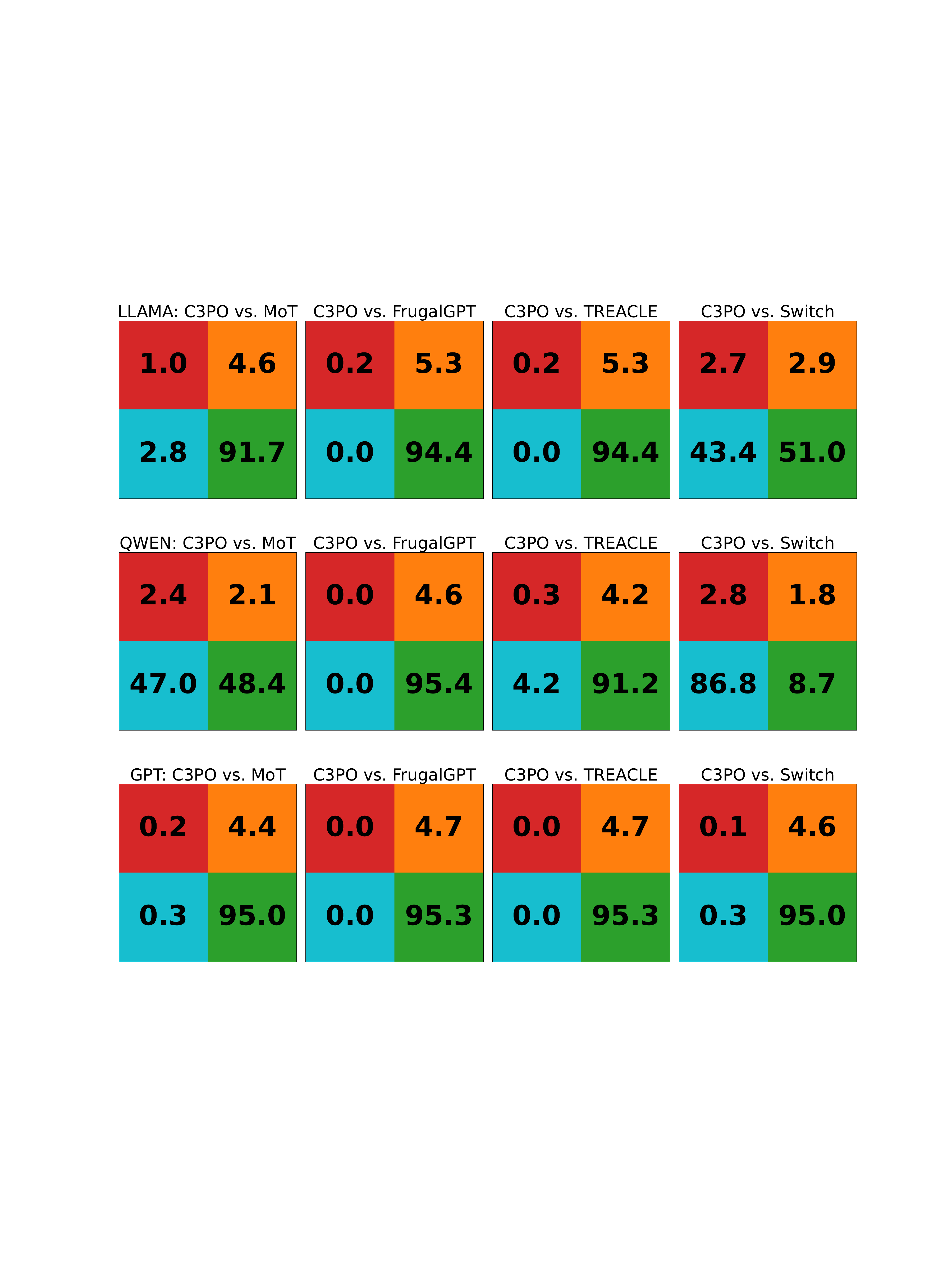}
    \caption{Percentage of {\color{red} `very bad'}, {\color{orange} `bad'}, {\color{cyan} `good'}, and {\color{green} `very good'} questions for each baseline and LLM cascade from the \textbf{SVAMP} dataset. For each baseline and LLM-cascade, we divide the test set into the following four non-overlapping categories; {\color{red} `very bad'}, {\color{orange} `bad'}, {\color{cyan} `good'}, and {\color{green} `very good'}. A question falls into the  {\color{red} `very bad'} category if C3PO fails to answer it correctly in spite of spending a higher inference cost compared to a baseline. If C3PO provides an incorrect answer to a question but incurs a lower cost compared to a baseline as well, then that question is {\color{orange} `bad'}. Clearly, having a {\color{orange} `bad'} question is better than having a {\color{red} `very bad'} question in terms of efficient usage of inference cost.
Similarly, {\color{cyan} `good'} questions refer to those test set examples, where C3PO provides  correct responses at the expense of increased cost compared to a baseline. Finally,  {\color{green} `very good'} questions are the ones, for which a correct answer from C3PO coincides with a lower cost compared to a baseline. We observe that in comparison to most baselines, when C3PO provides an incorrect answer, it is more likely to save inference cost. Additionally, for the majority of the questions, where C3PO's answer matches the true answer, it incurs reduced cost. This demonstrates the universal nature of C3PO's cost effectiveness across different types of questions. }
    \label{fig:good_bad_ugly_svamp}
\end{figure}

\begin{figure}[h!]
    \centering
    \includegraphics[width=0.9\linewidth, trim={7cm 19cm 5cm 19cm},clip]{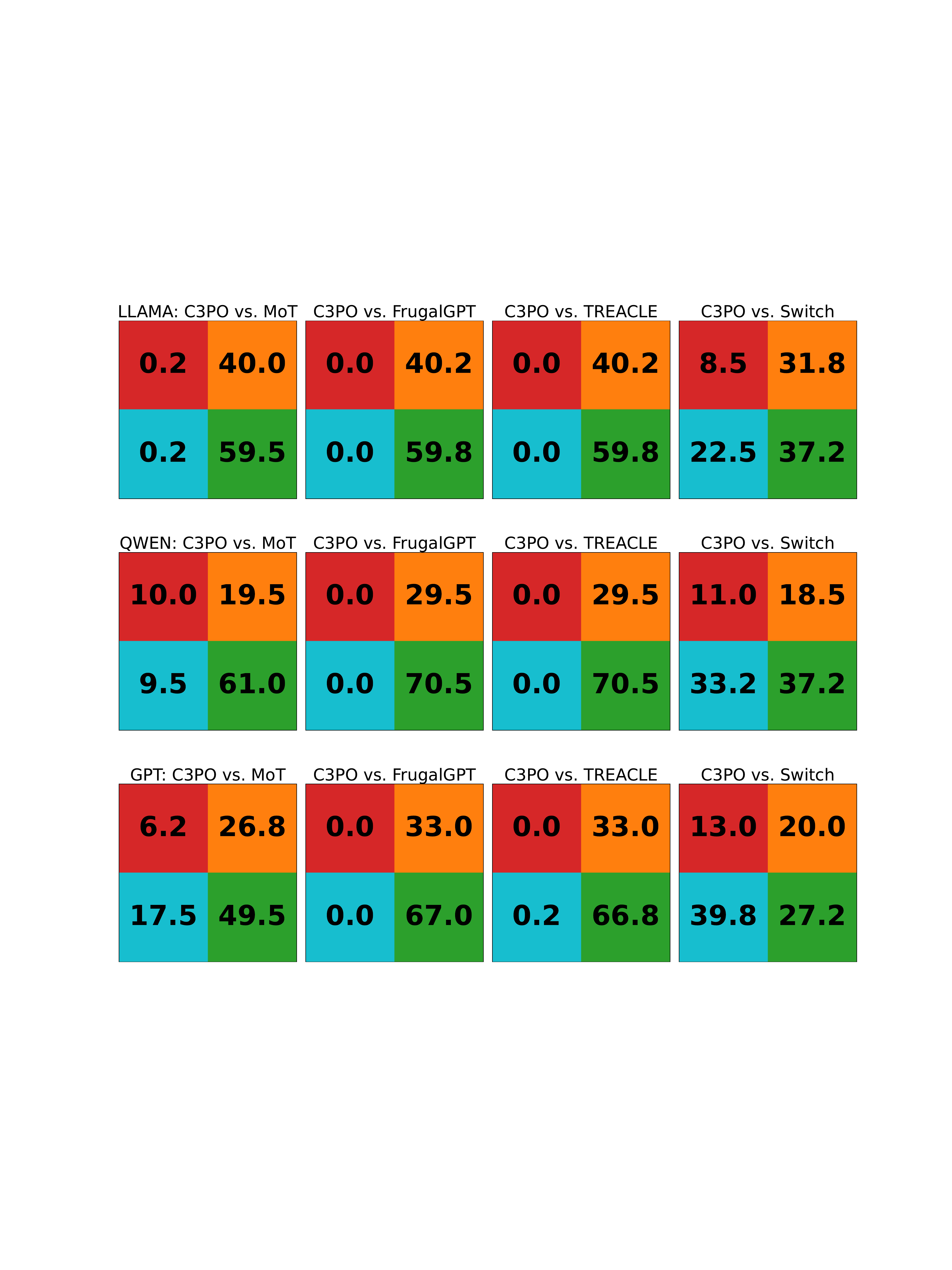}
    \caption{Percentage of {\color{red} `very bad'}, {\color{orange} `bad'}, {\color{cyan} `good'}, and {\color{green} `very good'} questions for each baseline and LLM cascade from the \textbf{MATH-500} dataset. For each baseline and LLM-cascade, we divide the test set into the following four non-overlapping categories; {\color{red} `very bad'}, {\color{orange} `bad'}, {\color{cyan} `good'}, and {\color{green} `very good'}. A question falls into the  {\color{red} `very bad'} category if C3PO fails to answer it correctly in spite of spending a higher inference cost compared to a baseline. If C3PO provides an incorrect answer to a question but incurs a lower cost compared to a baseline as well, then that question is {\color{orange} `bad'}. Clearly, having a {\color{orange} `bad'} question is better than having a {\color{red} `very bad'} question in terms of efficient usage of inference cost.
Similarly, {\color{cyan} `good'} questions refer to those test set examples, where C3PO provides  correct responses at the expense of increased cost compared to a baseline. Finally,  {\color{green} `very good'} questions are the ones, for which a correct answer from C3PO coincides with a lower cost compared to a baseline. We observe that in comparison to most baselines, when C3PO provides an incorrect answer, it is more likely to save inference cost. Additionally, for the majority of the questions, where C3PO's answer matches the true answer, it incurs reduced cost. This demonstrates the universal nature of C3PO's cost effectiveness across different types of questions.}
    \label{fig:good_bad_ugly_math500}
\end{figure}

\begin{figure}[h!]
    \centering
\includegraphics[width=0.9\linewidth, trim={7cm 19cm 5cm 19cm},clip]{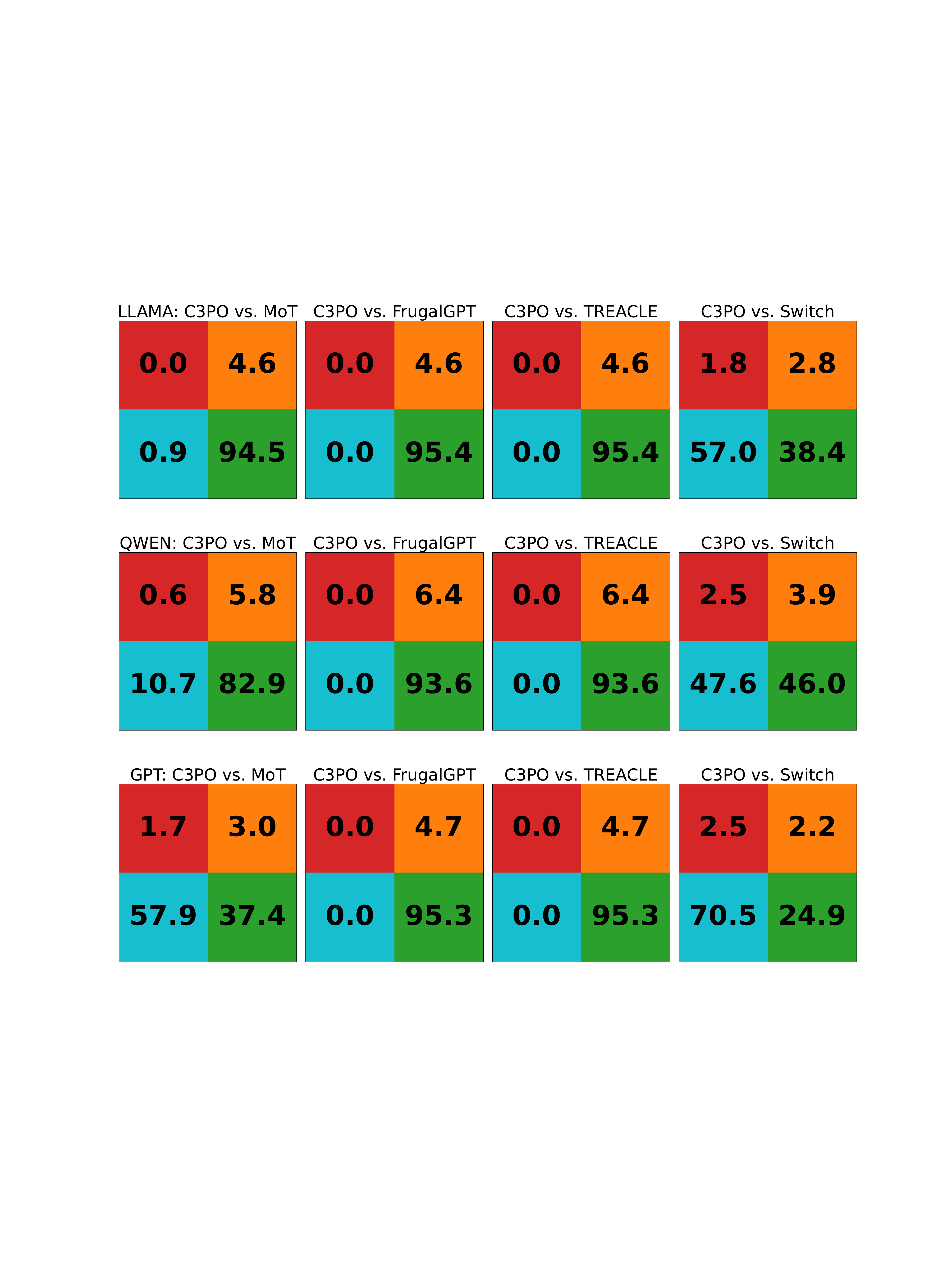}
    \caption{Percentage of {\color{red} `very bad'}, {\color{orange} `bad'}, {\color{cyan} `good'}, and {\color{green} `very good'} questions for each baseline and LLM cascade from the \textbf{GSM8K} dataset. For each baseline and LLM-cascade, we divide the test set into the following four non-overlapping categories; {\color{red} `very bad'}, {\color{orange} `bad'}, {\color{cyan} `good'}, and {\color{green} `very good'}. A question falls into the  {\color{red} `very bad'} category if C3PO fails to answer it correctly in spite of spending a higher inference cost compared to a baseline. If C3PO provides an incorrect answer to a question but incurs a lower cost compared to a baseline as well, then that question is {\color{orange} `bad'}. Clearly, having a {\color{orange} `bad'} question is better than having a {\color{red} `very bad'} question in terms of efficient usage of inference cost.
Similarly, {\color{cyan} `good'} questions refer to those test set examples, where C3PO provides  correct responses at the expense of increased cost compared to a baseline. Finally,  {\color{green} `very good'} questions are the ones, for which a correct answer from C3PO coincides with a lower cost compared to a baseline. We observe that in comparison to most baselines, when C3PO provides an incorrect answer, it is more likely to save inference cost. Additionally, for the majority of the questions, where C3PO's answer matches the true answer, it incurs reduced cost. This demonstrates the universal nature of C3PO's cost effectiveness across different types of questions.}
\label{fig:good_bad_ugly_gsm8k}
\end{figure}

\begin{figure}[h!]
    \centering
    \includegraphics[width=0.9\linewidth, trim={7cm 19cm 5cm 19cm},clip]{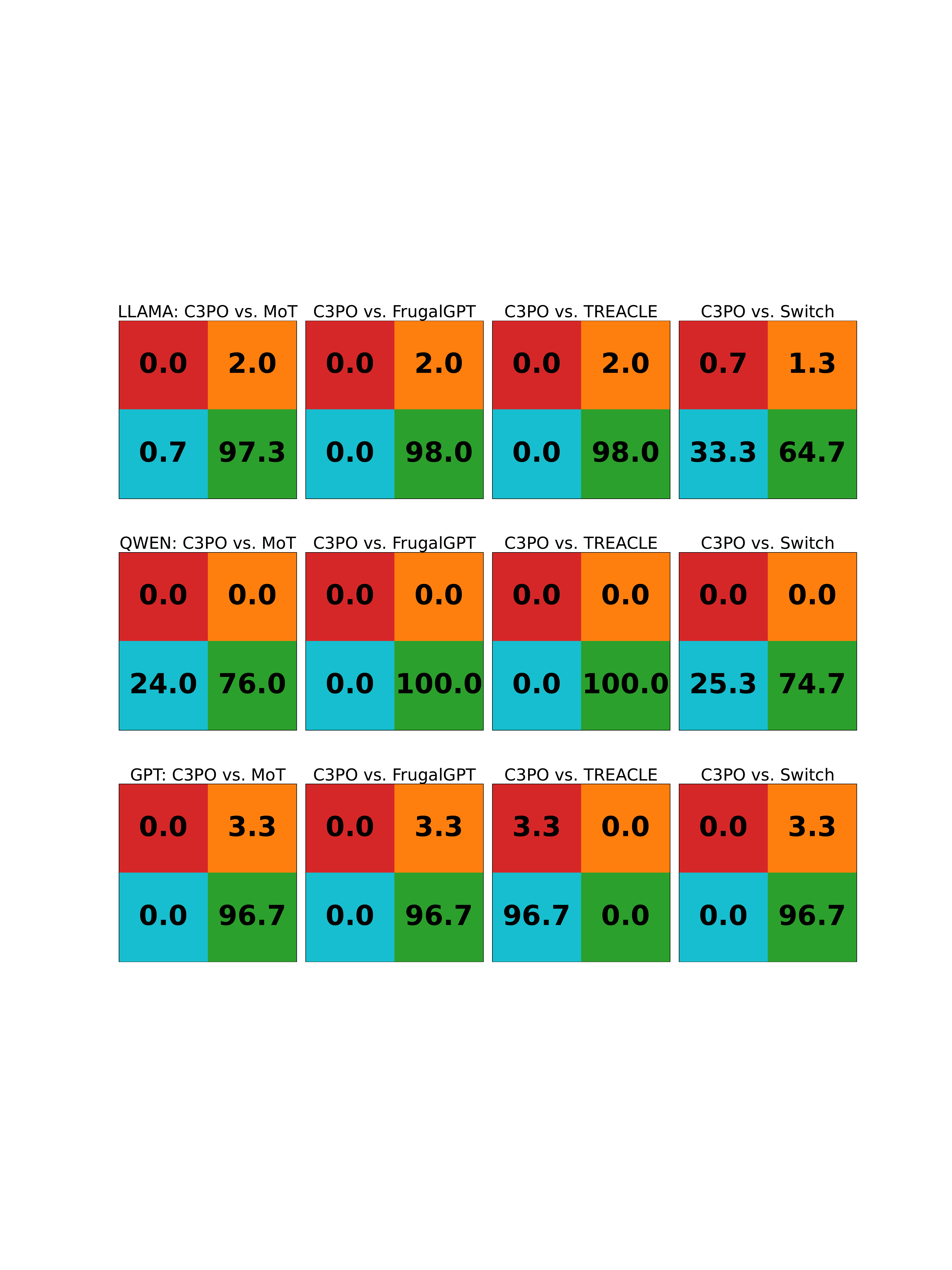}
    \caption{Percentage of {\color{red} `very bad'}, {\color{orange} `bad'}, {\color{cyan} `good'}, and {\color{green} `very good'} questions for each baseline and LLM cascade from the \textbf{BigBench Temporal Sequences} dataset.}
    \label{fig:good_bad_ugly_bb_temporal}
\end{figure}

Similar results are obtained for other datasets and are included in Figures~\ref{fig:good_bad_ugly_svamp},~\ref{fig:good_bad_ugly_math500},~\ref{fig:good_bad_ugly_gsm8k},~\ref{fig:good_bad_ugly_bb_temporal}.


\clearpage

\newpage

\section*{NeurIPS Paper Checklist}
\begin{enumerate}

\item {\bf Claims}
    \item[] Question: Do the main claims made in the abstract and introduction accurately reflect the paper's contributions and scope?
    \item[] Answer: \answerYes{} %
    \item[] Justification: Every claim in the paper is supported with experiments in Section 5 and proofs in \antonios{Appendix} Material.

\item {\bf Limitations}
    \item[] Question: Does the paper discuss the limitations of the work performed by the authors?
    \item[] Answer: \answerYes{} %
    \item[] Justification: There is a discussion of limitations of our work in the Conclusion section.

\item {\bf Theory Assumptions and Proofs}
    \item[] Question: For each theoretical result, does the paper provide the full set of assumptions and a complete (and correct) proof?
    \item[] Answer: \answerYes{} %
    \item[] Justification: We state the assumptions clearly in the proofs.

    \item {\bf Experimental Result Reproducibility}
    \item[] Question: Does the paper fully disclose all the information needed to reproduce the main experimental results of the paper to the extent that it affects the main claims and/or conclusions of the paper (regardless of whether the code and data are provided or not)?
    \item[] Answer: \answerYes{} %
    \item[] Justification: We provide all details required to reproduce our results: we describe our method in detail and provide a full pseudocode description of our approach. . In addition, the LLMs used in our work are all publicly accessible. We include our code for running our experiments in a zip file in the \antonios{Appendix} materials.

\item {\bf Open access to data and code}
    \item[] Question: Does the paper provide open access to the data and code, with sufficient instructions to faithfully reproduce the main experimental results, as described in supplemental material?
    \item[] Answer: \answerYes{} %
    \item[] Justification: We make all experimental data and code available and commit to making it publicly available upon acceptance.

\item {\bf Experimental Setting/Details}
    \item[] Question: Does the paper specify all the training and test details (e.g., data splits, hyperparameters, how they were chosen, type of optimizer, etc.) necessary to understand the results?
    \item[] Answer: \answerYes{} %
    \item[] Justification: We follow standard splits and implementations from previous works. Where we make any changes we clearly state them.

\item {\bf Experiment Statistical Significance}
    \item[] Question: Does the paper report error bars suitably and correctly defined or other appropriate information about the statistical significance of the experiments?
    \item[] Answer: \answerYes{} %
    \item[] Justification: We conduct repeated trials and provide 90\% confidence interval error bars in our figures.

\item {\bf Experiments Compute Resources}
    \item[] Question: For each experiment, does the paper provide sufficient information on the computer resources (type of compute workers, memory, time of execution) needed to reproduce the experiments?
    \item[] Answer: \answerYes{} %
    \item[] Justification: We detail our computer hardware in the paper. Our method requires very little compute. All LLM queries are performed by API calls.

\item {\bf Code Of Ethics}
    \item[] Question: Does the research conducted in the paper conform, in every respect, with the NeurIPS Code of Ethics \url{https://neurips.cc/public/EthicsGuidelines}?
     \item[] Answer: \answerYes{}
    \item[] Justification: We have read and confirm that our work fully complies with the Code of Ethics.

\item {\bf Broader Impacts}
    \item[] Question: Does the paper discuss both potential positive societal impacts and negative societal impacts of the work performed?
    \item[] Answer: \answerYes{} %
    \item[] Justification: We have an Appendix section dedicated to broader impacts.

\item {\bf Safeguards}
    \item[] Question: Does the paper describe safeguards that have been put in place for responsible release of data or models that have a high risk for misuse (e.g., pretrained language models, image generators, or scraped datasets)?
    \item[] Answer: \answerNA{}
    \item[] Justification: We do not release new datasets or models. 

\item {\bf Licenses for existing assets}
    \item[] Question: Are the creators or original owners of assets (e.g., code, data, models), used in the paper, properly credited and are the license and terms of use explicitly mentioned and properly respected?
    \item[] Answer: \answerYes{}
    \item[] Justification: We include references to each of the benchmark datasets used in the paper, as well as the LLMs used for the experiments. The licenses are listed for each dataset.

\item {\bf New Assets}
    \item[] Question: Are new assets introduced in the paper well documented and is the documentation provided alongside the assets?
    \item[] Answer: \answerYes{}
    \item[] Justification: We do not introduce a new dataset, but our source code includes comments and will be released under a MIT license upon acceptance.

\item {\bf Crowdsourcing and Research with Human Subjects}
    \item[] Question: For crowdsourcing experiments and research with human subjects, does the paper include the full text of instructions given to participants and screenshots, if applicable, as well as details about compensation (if any)? 
   \item[] Answer: \answerNA{}
    \item[] Justification: This work does not involve crowdsourcing or research with human subjects.

\item {\bf Institutional Review Board (IRB) Approvals or Equivalent for Research with Human Subjects}
    \item[] Question: Does the paper describe potential risks incurred by study participants, whether such risks were disclosed to the subjects, and whether Institutional Review Board (IRB) approvals (or an equivalent approval/review based on the requirements of your country or institution) were obtained?
    \item[] Answer: \answerNA{}
    \item[] Justification: This work does not require research with human subjects.

\item {\bf Declaration of LLM usage}
    \item[] Question: Does the paper describe the usage of LLMs if it is an important, original, or non-standard component of the core methods in this research? Note that if the LLM is used only for writing, editing, or formatting purposes and does not impact the core methodology, scientific rigorousness, or originality of the research, declaration is not required.
    \item[] Answer: \answerNA{}
    \item[] Justification: The core method development in this research does not involve LLMs as any important, original, or non-standard components.

\end{enumerate}

\end{document}